\documentclass[twoside,11pt]{article}
\usepackage{jmlr2e}
\usepackage{footmisc}
\usepackage{upgreek}

\usepackage[utf8]{inputenc}
\usepackage{subcaption}
\usepackage{alltt}
\usepackage{xhfill}
\usepackage{algorithm}
\usepackage{algpseudocode}
\usepackage{dsfont}

\usepackage{verbatim}
\newcommand{\E}{\mathbb{E}}

\newcommand{\cmt}[1]{{\textcolor{red}{#1}}}
\definecolor{ForestGreen}{rgb}{0.0, 0.27, 0.13}
\newcommand{\resp}[1]{{\color{ForestGreen}{#1}}}

\newcommand{\plam}{p_{\lammat}}

\renewcommand{\cmt}[1]{}
\renewcommand{\resp}[1]{}

\newcommand{\pdist}{\mathbb{P}}
\newcommand{\pden}{p}
\newcommand{\ptrue}{\pdist_{\tstar}}
\newcommand{\qdist}{\mathbb{Q}}
\newcommand{\qden}{q}
\newcommand{\qset}{\mc{Q}}

\newcommand{\snr}{\mrm{snr}}

\newcommand{\eval}{\rho}
\newcommand{\evali}[1][i]{\eval_{#1}}
\newcommand{\tagridge}{\mrm{Ridge}}

\usepackage{adjustbox}
\usepackage{hyperref}
\hypersetup{
    colorlinks=true,
    linkcolor=blue,
    filecolor=magenta,      
    urlcolor=cyan,
    citecolor=blue,
}

\definecolor{mygray}{gray}{0.5}
\definecolor{cblue}{RGB}{8, 85, 153}
\definecolor{darkblue}{RGB}{1, 43, 112}

\definecolor{cgreen}{RGB}{8, 153, 83}

\newcommand{\fref}[1]{Fig~\ref{#1}}

\usepackage{placeins}

\usepackage{fontawesome}
\PassOptionsToPackage{numbers,sort&compress}{natbib}

\usepackage{microtype}
\usepackage{graphicx}
\usepackage{booktabs} %

\usepackage{xspace}
\usepackage{array}
\usepackage{xr}
\usepackage{mathtools}
\usepackage{url}
\usepackage{comment}
\usepackage{graphicx}
\usepackage{enumitem}
\usepackage{wrapfig}
\usepackage{framed}
\usepackage{subcaption}

\newtheorem{mycorollary}{Corollary}

\newtheorem{mylemma}{Lemma}

\newtheorem{mydefinition}{Definition}

\newtheorem{myremark}{Remark}

\usepackage[capitalize]{cleveref}  
\usepackage{datetime}
\crefname{nlem}{Lemma}{Lemmas}
\crefname{nprop}{Proposition}{Propositions}
\crefname{nthm}{Theorem}{Theorems}
\crefname{exa}{Example}{Examples}
\crefname{assumption}{Assumption}{Assumptions}
\crefname{remark}{Remark}{Remarks}
\crefname{myremark}{Remark}{Remarks}
\crefname{equation}{equation}{equations}
\crefname{nclm}{Claim}{Claims}
\Crefname{section}{Sec.}{Sections}
\crefname{ncor}{Corollary}{Corollaries}
\crefname{nrem}{Remark}{Remarks}
\newcommand{\lam}{\lambda}

\def\mc#1{\mathcal{#1}}
\def\mb#1{\mathbb{#1}}
\def\mbf#1{\mathbf{#1}}
\def\mrm#1{\mathrm{#1}}
\def\v#1{\mbf{#1}} %
\def\mat#1{\mbf{#1}}
\newcommand{\R}{\mathbb{R}}
\newcommand{\Rd}{\R^{\dims}}

\newcommand{\sumn}[1][n]{\sum_{i=1}^{#1}}
\newcommand{\sless}[1]{\stackrel{#1}{\leq}}

\newcommand{\noisevar}{\sigma^2}
\newcommand{\noise}[1][i]{\xi_{#1}}

\newcommand{\ax}{x}
\newcommand{\y}{y}
\newcommand{\vy}{\v{y}}
\newcommand{\yi}[1][i]{\y_{#1}}
\newcommand{\axi}[1][i]{\ax_{#1}}

\newcommand{\tvar}{\theta}
\newcommand{\tstar}{\tvar_\star}
\newcommand{\that}{\widehat{\tvar}}

\newcommand{\xmat}{\mat{X}}

\newcommand{\lammat}{\mat{\Lambda}}
\newcommand{\xtx}{\xmat\tp\xmat}

\newcommand{\xridge}[1][\lam \identity_d]{\xmat\tp\xmat+#1}

\newcommand{\xridgenew}{\xridge[\lammat]}

\newcommand{\minnd}{\min\braces{n, d}}

\newcommand{\qtext}[1]{\ensuremath{{\quad\text{#1}\quad}}}

\newcommand{\dims}{\ensuremath{d}}
\newcommand{\samples}{\ensuremath{n}}
\newcommand{\obs}{\samples}
\newcommand{\real}{\ensuremath{\mathbb{R}}}

\newcommand{\order}[1]{\ensuremath{\mathcal{O}\parenth{#1}}}

\newcommand{\brackets}[1]{\left[ #1 \right]}
\newcommand{\parenth}[1]{\left( #1 \right)}

\newcommand{\braces}[1]{\left\{ #1 \right \}}

\newcommand{\tp}{^\top}
\newcommand{\inv}{^{-1}}
\newcommand{\ceil}[1]{\left\lceil #1 \right\rceil}

\newcommand{\myvec}{w} %
\newcommand{\defn}{:=}

\newcommand{\norm}[1]{\ensuremath{\| #1 \|}}
\newcommand{\enorm}[1]{\ensuremath{\| #1 \|}} %

\newcommand{\kull}[2]{\ensuremath{\mc{D}_{\mrm{KL}}(#1\; \| \; #2)}}

\newcommand{\Exs}{\ensuremath{{\mathbb{E}}}}

\DeclareMathOperator{\diag}{diag}

\DeclareMathOperator{\trace}{trace}

\providecommand{\argmax}{\mathop\mathrm{arg max}} %
\providecommand{\argmin}{\mathop\mathrm{arg min}}

\newlength{\widebarargwidth}

\makeatletter
\long\def\@makecaption#1#2{
        \vskip 0.8ex
        \setbox\@tempboxa\hbox{\small {\bf #1:} #2}
        \parindent 1.5em  %
        \dimen0=\hsize
        \advance\dimen0 by -3em
        \ifdim \wd\@tempboxa >\dimen0
                \hbox to \hsize{
                        \parindent 0em
                        \hfil
                        \parbox{\dimen0}{\def\baselinestretch{0.96}\small
                                {\bf #1.} #2
                                }
                        \hfil}
        \else \hbox to \hsize{\hfil \box\@tempboxa \hfil}
        \fi
        }
\makeatother

\long\def\comment#1{}
\definecolor{carnelian}{rgb}{0.7, 0.11, 0.11}
\definecolor{battleshipgrey}{rgb}{0.52, 0.52, 0.51}
\definecolor{darkgray}{rgb}{0.66, 0.66, 0.66}
\definecolor{indiagreen}{rgb}{0.07, 0.53, 0.03}
\definecolor{darkgreen}{rgb}{0.0, 0.2, 0.13}
\definecolor{darkspringgreen}{rgb}{0.09, 0.45, 0.27}
\definecolor{dukeblue}{rgb}{0.0, 0.0, 0.61}
\definecolor{olivedrab7}{rgb}{0.24, 0.2, 0.12}
\definecolor{darkblue}{rgb}{0.0, 0.0, 0.55}
\definecolor{darkscarlet}{rgb}{0.34, 0.01, 0.1}
\definecolor{candyapplered}{rgb}{1.0, 0.03, 0.0}
\definecolor{ao(english)}{rgb}{0.0, 0.5, 0.0}
\definecolor{applegreen}{rgb}{0.55, 0.71, 0.0}

\def\balign#1\ealign{\begin{align}#1\end{align}}
\def\baligns#1\ealigns{\begin{align*}#1\end{align*}}
\def\balignat#1\ealign{\begin{alignat}#1\end{alignat}}
\def\balignats#1\ealigns{\begin{alignat*}#1\end{alignat*}}
\def\bitemize#1\eitemize{\begin{itemize}#1\end{itemize}}
\def\benumerate#1\eenumerate{\begin{enumerate}#1\end{enumerate}}

\newenvironment{talign*}
 {\let\displaystyle\textstyle\csname align*\endcsname}
 {\endalign}
\newenvironment{talign}
 {\let\displaystyle\textstyle\csname align\endcsname}
 {\endalign}

\def\balignst#1\ealignst{\begin{talign*}#1\end{talign*}}
\def\balignt#1\ealignt{\begin{talign}#1\end{talign}}

\graphicspath{{../figs/},{figs/}}

\newcommand{\mdltag}{\ensuremath{\mathrm{MDL}\!\!-\!\!\mathrm{COMP}}}
\newcommand{\normconst}{\mrm{C}_{\lammat}}
 
\newcommand{\luck}{p_\textrm{luck}}
\newcommand{\pracmdltag}{\textrm{Prac-MDL-COMP}}
\newcommand{\lammatopt}{\lammat_{\textrm{opt}}}

\newcommand{\ropt}{\ensuremath{\mc R_{\textrm {opt}}}}

\newcommand{\fstar}{f^\star}
\newcommand{\rkhs}{\mb{H}}
\newcommand{\kmat}{\mat{K}}
\newcommand{\kridge}{\mat{K} + \lam \mat{I}}
\newcommand{\kernel}{\mc{K}}
\newcommand{\Rn}{\real^n}
\newcommand{\kconst}{\mrm{C}_{\lam, \kernel}}
\newcommand{\doth}[1]{\langle #1 \rangle_{\rkhs}}
\renewcommand{\P}{\mathbb{P}}
\newcommand{\plin}{\mc{P}_{\textrm{bndvar}}}

\newcommand{\YobsSpace}{\ensuremath{\mathcal{Y}^n}}
\newcommand{\pdata}{\pden_{\mrm{data}}}
\newcommand{\betahat}{\widehat{\beta}}
\newcommand{\taglin}{\xmat}
\newcommand{\qsetlin}{\qset_{\tagridge}^{\taglin}}
\newcommand{\tagker}{\kmat}
\newcommand{\qsetker}{\qset_{\tagridge}^{\tagker}}

\newcommand{\SNR}{\textrm{SNR}}
\newcommand{\ntest}{n_{\textrm{test}}}
\newcommand{\dmax}{\overline{d}}
\newcommand{\templam}{\overline{\lammat}}
\newcommand{\ystar}{\v{y^\star}}
\newcommand{\normconstk}{\mrm{C}_{\lam}}

\usepackage{lastpage}
\jmlrheading{24}{2023}{1-\pageref{LastPage}}{9/21}{10/23}{21-1133}{Raaz Dwivedi, Chandan Singh, Bin Yu, Martin Wainwright}
\ShortHeadings{Revisiting MDL complexity in overparameterized models}{Dwivedi, Singh, Yu, Wainwright}

\begin{document}

\title{Revisiting minimum description length complexity in overparameterized models}

 \author{\name Raaz Dwivedi$^\star$ \email dwivedi@cornell.edu \\
       \addr Department of Operations Research \& Information Engineering\\
       Cornell Tech, Cornell University\\
       New York City, NY
       \AND
       \name Chandan Singh$^\star$ \email chansingh@microsoft.com \\
       \addr Microsoft Research \\ 
       Seattle, WA
       \AND 
       \name Bin Yu \email binyu@berkeley.edu \\
       \addr Department of Statistics, and Electrical Engineering and Computer Sciences\\ 
       University of California, Berkeley \\
       Berkeley, CA
       \AND
       \name Martin Wainwright \email wainwrigwork@gmail.com \\ 
       \addr Department of Electrical Engineering and Computer Sciences, and Mathematics \\
       Massachusetts Institute of Technology \\
       Cambridge, MA\\ 
       $\star=$ equal contribution.
       }
\editor{Jean-Philippe Vert}
\vspace{-2mm}
\maketitle
\begin{abstract}
Complexity is a fundamental concept underlying statistical learning theory that aims to inform generalization performance. Parameter count, while successful in low-dimensional settings, is not well-justified for overparameterized settings when the number of parameters is more than the number of training samples.
We revisit complexity measures based on Rissanen's principle of minimum description length (MDL) and define a novel MDL-based complexity (MDL-COMP) that remains valid for overparameterized models. MDL-COMP is defined via an optimality criterion over the encodings induced by a good Ridge estimator class. We provide an extensive theoretical characterization of MDL-COMP for linear models and kernel methods and show that it is \emph{not} just a function of parameter count, but rather a function of the singular values of the design or the kernel matrix and the signal-to-noise ratio. For a linear model with $n$ observations, $d$ parameters, and i.i.d. Gaussian predictors, MDL-COMP scales linearly with $d$ when $d<n$, but the scaling is exponentially smaller---$\log d$ for $d>n$. For kernel methods, we show that MDL-COMP informs minimax in-sample error, and can decrease as the dimensionality of the input increases. We also prove that MDL-COMP upper bounds the in-sample mean squared error (MSE). Via an array of simulations and real-data experiments, we show that a data-driven Prac-MDL-COMP informs hyper-parameter tuning for optimizing test MSE with ridge regression in limited data settings, sometimes improving upon cross-validation and (always) saving computational costs. Finally, our findings also suggest that the recently observed double decent phenomenons in overparameterized models might be a consequence of the choice of non-ideal estimators.
\end{abstract}

\section{Introduction}
\label{sec:introduction}
Occam's razor and the bias-variance tradeoff have long provided
guidance for model selection in statistics and machine learning.
Given a dataset, these principles recommend selecting a model that
balances between (a) fitting the data well,
and (b) having relatively low complexity.
Roughly speaking, in the low-dimensional regime, one typically observed the following tradeoff:
On the one hand, a model whose complexity is too low incurs \emph{high bias}, i.e., it does not fit the training data well (under-fitting);
on the other hand, an overly complex model memorizes the training data, suffering from \emph{high variance}, i.e. poor performance on unforeseen data (over-fitting).

There are numerous characterizations of complexity in the statistical machine learning literature that are commonly used to perform model selection, and to establish bounds on prediction error.
These include Akaike information criterion~\citep{akaike1974new}, Mallow's $C_p$~\citep{mallows1973some}, Bayesian information
criterion~\citep{schwarz1978estimating},
Vapnik-Chervonenkis dimension~\citep{vapnik2015uniform}, and Rademacher complexity~\citep{bartlett2002rademacher,anthony2009neural,van2006empirical}. Intuitively, such measures reflect the effective number of parameters used to fit the model.
In the classical regime for linear regression with $d$ features, and $n$ training data points, when $d < n$, and the design matrices are well-conditioned, all of these measures reduce to simple parameter counting for linear models. Another common proxy for model complexity is the degrees of
freedom~\citep{efron1986biased,buja1989linear,efron2004estimation}, 
along with extensions such as effective or generalized degrees of freedom for more complex models~\citep{meyer2000degrees,shen2002adaptive,efron2004estimation,hastie2005elements,zhang2012generalized}, and high-dimensional sparse regression~\citep{zou2007degrees,tibshirani2012degrees,hastie2017generalized}.
As a special case, in an unstructured linear
regression problem based on $n$ samples with $d$ features (with $d <
n$), the degree of freedom is equal to $d$.
However, recent work~\citep{kaufman2014does,janson2015effective,tibshirani2015degrees}
has shown that when moving to the over-parameterized setting ($d >
n$), effective degrees of freedom may be a poor measure of model
complexity. In some past work, the variance of the estimator itself has sometimes been used as a measure of complexity (e.g., in $L^2$-boosting~\citep{buhlmann2003boosting}).  However, such a choice may be misinformed when the bias term is dominant.

Overall, the choice of parameter count as a complexity measure for linear
models is rigorously justified when the data is low-dimensional ($d <
n$), and the design matrix is well-conditioned (so that all $d$
directions contribute in roughly equal measure).
Indeed, under these conditions, the OLS
estimate has good performance when $d \ll n$, and its test error
increases proportionally with $d$ in this regime.  However, using $d$ as the complexity measure when $d>n$ remains unjustified even for linear models, and for low-dimensional linear models when the design matrix is not well-conditioned; and in high-dimensional models, the design matrix by definition is ill-conditioned since it does not even have full rank.

More recently, a line of work has derived generalization error bounds for deep neural networks~\citep{neyshabur2015norm,bartlett2017spectrally,neyshabur2017pac,golowich2017size,li2018tighter,neyshabur2018role}
based on Rademacher-like complexity notions.
However, at least thus far, such bounds remain too loose to inform practical performance~\citep{arora2018stronger}. Moreover, there is growing evidence that heavily overparameterized models once trained are often not complex, due to the implicit regularization induced by model architecture, optimization methods including initialization, and training datasets~\citep{Nakkiran2019DeepDD,neyshabur2014search, arora2019implicit, neyshabur2018role}. 
There has been a series of recent work investigating generalization performance of overparameterized linear models and kernel methods as they can be seen as as tractable settings for providing theoretical insight into the behavior of
(overparameterized) deep neural networks. See, e.g., \cite{belkin2018understand,jacot2018neural,du2018gradient, allen2018convergence,hastie2019surprises,bartlett2020benign,tsigler2020benign} and the references therein.

The starting point for this work is to seek a valid complexity measure for regression tasks with overparameterized linear models, that can be easily extended to kernel methods, and then to use this complexity measure for tuning regularization parameter.
To do so, we build on the optimality principle put forth in the algorithmic complexity of Kolmogorov, Chaitin, and Solomonoff~\citep{kolmogorov1963tables,kolmogorov1968three,li2008introduction} and the principle of minimum description length (MDL) of
Rissanen~\citep{rissanen1986stochastic,barron1998minimum,hansen2001model,grunwald2007minimum}. For linear models, the known MDL complexity measures also scale roughly linearly with dimension $d$, or they can be infinite (see \cref{sub:nml})

\paragraph{Our contributions:} 
We define a new complexity measure MDL-COMP that corresponds to the minimum excess bits required to transmit the data when constructing an optimal code for a given dataset via a family of models based on the ridge estimators. Within this framework, we undertake a detailed study of high-dimensional linear models and kernel methods.

We show that \mdltag\ has a wide range of behavior depending on the design matrix (or the kernel matrix). For linear models with $d$ features, and $n$ samples, it usually scales like $d/n$ for $d<n$, and grows very slowly---logarithmic in $d$---for $d>n$ (see \cref{fig:mdl_gaussian,fig:mdl_spike_design}). We establish that \mdltag\ provides an upper bound for in-sample MSE
(\cref{thm:lam_opt}), and that it satisfies certain minimax optimality
criterion (\cref{thm:minimax_codelength}). For kernel methods, we show that for kernels in Gaussian and Sobolev spaces, MDL-COMP can inform the minimax in-sample generalization (\cref{thm:mdl_kernel} and Corollary~\ref{cor:mdl_kernel_smooth}). Interestingly, for neural tangent kernels (NTKs), we find that MDL-COMP itself can sometimes reduce when the dimensionality of the input increases.

Next, to evaluate the practical usefulness of \mdltag, we consider a data-driven form of
\mdltag-inspired hyperparameter selection, which provides competitive performance with cross-validation for ridge regression (in terms of test MSE), especially in limited data settings in several simulations and real-data experiments. Moreover, this criterion can provide computational savings especially while training overparameterized models in contrast to the vanilla K-fold cross-validation (since computation is only required for a single fold). Finally, we also highlight some insights that our findings provide for the recently observed \emph{double descent} phenomenon on test error of overparameterized models.

\paragraph{Organization:} %
\label{par:organization_}
We start with a background on MDL and setting the notation in \cref{sub:background}, followed by the definitions of complexity central to this work in \cref{sec:complexity}. We then provide our main results and their consequences in \cref{sec:main_results}, and present several numerical experiments in \cref{sec:experiments}.
We conclude with a discussion and directions for future work in
\cref{sec:discussion}, where we also discuss the consequences of our results for double-descent. Proofs of all results and additional experiments are provided in the appendix.

\section{Background on the principle of minimum description length}
\label{sub:background}
In order to define a complexity measure in a principled manner, we
build upon Rissanen's principle of minimum description length (MDL).
It has its intellectual roots in Kolmogorov's theory of complexity.
Both approaches are based on an optimality requirement: namely, the
shortest length program that outputs a given sequence on a universal
Turing machine for Kolmogorov complexity, and the shortest
(probability distribution-based) codelength for the given data for MDL.
Indeed, Rissannen generalized Shannon's coding theory to universal
coding and used probability distributions to define a universal encoding
for a dataset and then used the codelength as an approximate measure
of Kolmogorov's complexity.

\subsection{Basic principle} %
\label{sub:introduction}
From the perspective of minimum description length, a model or a
probability distribution for data is equivalent to a (prefix) code;
one prefers a code that exploits redundancy in the data to compress it
into as few bits as possible;
see~\citep{rissanen1986stochastic,barron1998minimum,grunwald2007minimum}.
Since its introduction, MDL has been used in many tasks including
density estimation~\citep{zhang2006}, time-series
analysis~\citep{tanaka2005discovery}, model
selection~\citep{barron1998minimum, hansen2001model,
  miyaguchi2018high}, and DNN
training~\citep{hinton1993keeping,schmidhuber1997discovering,li2018measuring,blier2018description}.
For readers unfamiliar with MDL, Sections 1 and 2 of the
papers~\citep{barron1998minimum,hansen2001model} provide background on
MDL.

In the MDL
framework, any probability model is viewed as a type of coding, so
that for example, fitting a Gaussian linear model is equivalent to
using a Gaussian code for compressing the data. The goal is to select
the code that provides the shortest description of data; in most
settings, this translates into picking the model with the best fit to
the data. More formally, suppose we are given a set of $\obs$
observations $\vy = \braces{\yi[1], \yi[2], \ldots, \yi[n]}$.  Let
$\qdist$ refer to a probability distribution on the space $\YobsSpace$
(e.g., a subset of $\real^n$) where $\vy$ takes values, and let
$\qset$ denote a set of such distributions.  Given a probability
distribution $\qdist$, we can associate an
\emph{information-theoretic} prefix code for the space $\YobsSpace$,
wherein for any observation $\vy$ we need to use $\log(1/\qdist(\vy))$
number of bits to \emph{describe} $\vy$. The MDL principle dictates
choosing the code $\qdist$ associated with the shortest possible
description length---namely, a code achieving the minimum
$\min_{\qdist \in \qset } \log(1/\qdist(\vy))$. Note that when $\qset$
is simply a parametric family, the direct MDL principle reduces to the
maximum likelihood principle. But the advantage of MDL comes from the
fact that the set $\qset$ can be more complex, e.g. a nested
union of parametric families, or a set of codes not indexed by the
canonical parameter of interest. Furthermore, even without a
generative modeling assumption, the notion of shortest description
over an arbitrary set of codes $\qset$ continues to be
well-defined. (For further discussion about MDL vs maximum likelihood,
we refer the reader to \cref{sub:further_background}.)

\subsection{Two-stage MDL}
\label{sub:two_stage}
One of the earliest notions of MDL is two-stage MDL, often considered for doing model selection over a nested family of parametric model classes, where the dimensionality of the parameter varies across different parametric classes~\citep{hansen2001model}.
This version of MDL turns out to be equivalent to the Bayesian
information criterion (BIC)---that is, it performs model selection
based on a regularized maximum likelihood, where the regularization
term is simply $\frac d2\log n$.  Consequently, apart from the
additional logarithmic term, the MDL complexity in this set-up simply
reduces to parameter counting.

\subsection{Normalized maximum likelihood}
\label{sub:nml}
Many modern approaches to MDL are based on a form of universal coding
known as \emph{normalized maximum likelihood}, or NML for short~\citep{shtar1987universal}.
In
this approach, the distribution $\qdist$ is defined directly on the
space $\YobsSpace$; at least in general, it is \emph{not} explicitly
parametrized by the parameter of interest. More concretely, given a
family of codes $\mc P_{\Theta} = \braces{\pden(\cdot; \tvar),
  \theta\in \Theta}$, the NML code is defined as
\begin{align}
\label{eq:nml_defn}
    q_{\mrm{NML}}(\vy) := \frac{\max_{\tvar} \pden(\vy; \tvar)}{\int_{\YobsSpace} \max_{\tvar'} \pden(\v{y'}; \tvar') d\v{y'}},
\end{align}
assuming that the integral in the denominator is finite.
\cite{shtar1987universal} established that this NML distribution (when
defined) provides the best encoding for the family $\mc P_{\Theta}$ in
a minimax sense. The log normalization constant
\begin{align}
\label{eq:shtarkov_complexity}
  \log \int_{\YobsSpace} \max_{\tvar'} \pden(\v{y'}; \tvar') d\v{y'}
\end{align}
associated with this code is referred to as the \emph{NML or Shtarkov
complexity.} Note that the normalization~\eqref{eq:nml_defn} ensures
that $\qdist_{\mrm{NML}}$ is a valid code by making $q_{\mrm{NML}}$ a
valid density.\footnote{Thus, Kraft's inequality guarantees the
existence of a code corresponding to it.} Such codes are called
\emph{universal codes}, since the codelength
$\log(1/q_{\mrm{NML}}(\vy))$ is universally valid for any $\vy
\in \YobsSpace$.

\paragraph{Known results:} 
Suppose that $\mc P_{\Theta}$ is a parametric class of dimension $d$.
In this special case, under suitable regularity conditions, the NML complexity scales asymptotically as
$\tfrac{1}{2} d \log n$, for fixed $d$ with $n \to \infty$;
consequently, it is asymptotically equivalent to the BIC complexity
measure~\citep{barron1998minimum,foster2004contribution,hansen2001model}
to the first order $\mc O(\log n)$ term.  In recent work,
\cite{grunwald2017tight} further developed a framework to unify
several complexities including Rademacher and NML, and derived excess
risk bounds in terms of this unifying complexity measure, for several
low-dimensional settings.

\paragraph{Challenges with NML complexity:}
In overparametrized settings (and even in several settings otherwise), the NML code suffers from the \emph{infinity} problem, namely the normalization constant in \cref{eq:nml_defn} is infinite, and the NML distribution is not defined. A canonical solution is to truncate the observation space so as to make the integral finite. But with simple models like linear regression, such schemes provide volume of the truncated space as the complexity measure. For example, consider the codes corresponding to Gaussian linear model, where 
\begin{align}
\label{eq:linear_code}
  p(\v{y}; \theta) \propto \exp(-\frac{1}{2\sigma^2}\norm{\xmat\theta-\vy}_2^2),
\end{align}
 and we treat the design matrix $\xmat \in \real^{n\times d}$ and the scalar $\sigma^2$ known and fixed, and $\vy\in\mc Y \subset \real^{n}$ denotes the observation. When $d>n$, and $\xmat$ has full row rank ($n$), it is straightforward to see that, $\max_{\theta}p(\v{y}; \theta) = \text{constant}$, which in turn implies that the NML complexity in \cref{eq:shtarkov_complexity} is infinite; and the NML code~\eqref{eq:nml_defn} is not defined (also, see
\citet[Example 11.1]{grunwald2007minimum}). A canonical solution of this problem is truncation of the response space~\cite{barron1998minimum}, but in the setting above such a restriction leads to a trivial NML complexity measure that depends merely on the volume of the truncated space (and is independent of $\xmat, d$). 

\subsection{Luckiness normalized maximum likelihood}
To deal with the infinity problem discussed above, another solution was proposed in recent works, namely, the luckiness normalized maximum likelihood (LNML) code (see Chapter
11~\citep{grunwald2007minimum}). Given a class of codes $\mc P_{\Theta}$, and a luckiness function $\luck:\Theta \to \real_+$ (not necessarily a code), one way to define the LNML code is
as follows:
\begin{align}
\label{eq:lnml}
    \qden_{\mrm{LNML}}(\vy) := \frac{\max_{\tvar} \parenth{\pden(\v{y}; \tvar) \cdot \luck(\tvar)}}{
    \displaystyle\int_{\v{z}\in\real^n}\max_{\tvar'}\parenth{\pden(\v{z};  \tvar') \cdot \luck(\tvar')}\; d\v{z}}.
\end{align}
Once again, the normalization in \cref{eq:lnml} ensures that $\qdist_{\mrm{LNML}}$ whenever well-defined is a universal code. One can now treat the log normalization constant of this code as a complexity measure, but such a definition would vary with the luckiness function chosen by the user. In the sequel, we provide a principled way to consider a family of such LNML codes and then derive a complexity measure using an optimality criterion over these codes. Theoretical investigations with LNML codes have not been extensively done in the prior work, and are central to the current work. It is worth noting that while NML can be seen as a generalization of the maximum likelihood principle, LNML can be interpreted as a generalization of regularized maximum likelihood principle (see \cref{sub:further_background} for a detailed discussion about relationship betwee MDL and maximum likelihood).

It is useful to note the recent unified interpretation for the LNML codes~\cref{eq:lnml} by \citet{grunwald2019minimum}. Given a luckiness function $\luck$, \citep{grunwald2019minimum} define an \emph{MDL estimator based on $\luck$} as follows:
\begin{align}
\label{eq:mdl_estimator}
    \widehat{\theta}_{\mathrm{luck}}(\v{y}) \defn \arg\max_{\tvar} \pden(\v{y}; \tvar) \cdot \luck(\tvar).
\end{align}
Note that the term inside the $\arg$ on the RHS is the same term appearing in the numerator of \cref{eq:lnml}.
The $\luck$-MDL estimator~\cref{eq:mdl_estimator} is effectively a penalized maximum likelihood estimator, and coincides with the Bayes Maximum A Posteriori estimate (MAP) whenever $\luck$ is a probability density. Indeed, as we later discuss, here we investigate a family of MDL-estimators parameterized by positive semidefinite matrices $\mathbf{\Lambda}$ where for each estimator $\luck = p_{\mathbf{\Lambda}}$ is defined using a particular ridge estimator associated with regularization term based on $\lammat$. In such a case, $\widehat{\theta}_{\mathrm{luck}} = \widehat{\theta}_{\lammat}$ is a ridge estimator (see \cref{eq:ridge_new_closed,eq:lnml_ridge}).

\subsection{Optimal redundancy as a complexity measure}
One metric to measure the effectiveness of any code $\qdist$ when the observations follow a generative model $\vy\sim\pdist_{\star}$,  is the \emph{redundancy} or the expected excess code-length for $\qdist$ compared to the true (but unknown) distribution $\pdist_{\star}$,
given by:
 \begin{align}
 \label{eq:redundancy}
   \frac{1}{n} \Exs_{\vy \sim \pdist_{\star}} \brackets{\log
     \parenth{\frac{1}{\qdist(\vy)}} - \log
     \parenth{\frac{1}{\pdist_{\star}(\vy)}}} =
   \Exs_{\vy\sim\pdist_{\star}}\brackets{ \frac{1}{n} \log \parenth{
       \frac{\pdist_{\star}(\vy)}{\qdist(\vy)}}} = \frac{1}{n}
   \kull{\pdist_{\star}}{\qdist},
\end{align}
where we normalized by $n$, to obtain bits(/nats) per sample point,
since in our notation the code $\qdist$ is defined jointly over the
$n$ observations.  Here $\kull{\pdist_{\star}}{\qdist}$ denotes the
Kullback-Leibler divergence between the two distributions and is
non-negative unless $\qdist = \pdist_\star$. Or, in other words, the
unknown $\pdist_\star$ is also the best possible encoding of the
data.  Thus, the complexity of the data with respect to a given set of
codes $\qset$ can be defined via the best possible excess codelength
also known as \emph{optimal redundancy}:
\begin{align}
\label{eq:ropt_defn}
  \ropt(\pdist_\star, \qset) \defn \min_
  {\qdist\in\qset} 
\kull{\pdist_{\star}}{\qdist}.
\end{align} 
\paragraph{Remark:} For the complexity measure~\eqref{eq:ropt_defn} to be effective, the set of codes $\qset$ should be rich enough; otherwise, the calculated complexity can provide trivial or loose estimates (e.g., as discussed above, in overparameterized setting when $\mc Q$ contains only the NML code and $\mc Y$ is unbounded). As noted earlier, the advantage of the MDL framework comes from the fact that the set $\qset$ does not have to be the usual parametric class used for computing maximum likelihood (and as noted above, the NML and LNML codes are strict global generalizations of the maximum likelihood principle). Furthermore, even without a generative model, one can consider a minimax notion of redundancy as a complexity measure~\cite{barron1998minimum}.
In the sequel, however, we restrict our derivations to settings with a known generative model over the observed data and briefly discuss a minimax set-up in \cref{sub:a_minimax_optimality_via_mdltag}.

\section{Ridge-based minimum description length complexity}
\label{sec:complexity}

In this work, we define a new complexity measure, MDL-COMP, using the optimality principle~\eqref{eq:ropt_defn} and a family of LNML codes, that are induced by ridge estimators. In a nutshell, the luckiness function~\eqref{eq:lnml} is inspired by the penalty used in ridge regression. Our choice of ridge-based LNML codes is informed by multiple reasons: First, we are interested in an operational definition of complexity, and hence we consider encodings that are associated with a computationally feasible set of estimators.  Second, for the complexity to be informative, we prefer the set of codes to be rich enough, and thus the estimators that we consider should achieve good predictive performance. Ridge estimators are known to achieve the minimax performance with linear and kernel methods~\citep{raskutti2011minimax,zhang2015divide,dicker2016ridge}, and often provide competitive predictive performance in applied
machine-learning tasks~\citep{bernau2014cross}.\footnote{We do not simply use the ordinary-least-squares-estimator based
code---which in the NML framework would reduce the problem to a single
NML code, and the Shtarkov complexity---for the following reasons: (i)
it has poor performance in the over-parameterized regime with linear
models, and for any setting of kernel regression (for
infinite-dimensional kernels), and (ii) as shown earlier, the Shtarkov
complexity would be infinite when $\mc{Y}$ is unbounded, in the overparameterized settings
considered in the sequel.}

We now define the details of these codes in
\cref{sub:ridge_lnml_codes} (besides providing a brief recap of ridge
regression), and then formally define the MDL-COMP in
\cref{sub:mdl_comp_defn}.

\subsection{Ridge-based LNML codes}
\label{sub:ridge_lnml_codes}
As in the MDL literature, we design codes over response vectors $\vy
\in \real^n$ conditional on a fixed matrix $\xmat \in \real^{n \times
  d}$ of covariates.  We briefly recap ridge regression estimators for
linear models and kernel methods in \cref{sub:ridge_background},
followed by the definitions of the corresponding LNML codes in
\cref{sub:lnml_code_defn} that underlie our definition of MDL-COMP in
\cref{sub:mdl_comp_defn}.

\subsubsection{Background on ridge regression}
\label{sub:ridge_background}
For linear models, the generalized $\ell_2$-regularized least square estimators, with the penalty~$\tvar\tp\lammat\tvar$ for a positive definite matrix $\lammat$, is given by
\begin{align}
  \label{eq:ridge_new_closed}
  \that_\lammat(\vy) \defn \argmin_{\theta\in \Rd}
  \parenth{\frac{1}{2}\enorm{\v{y}-\xmat\tvar}_2^2 + \frac12\tvar\tp
    \lammat \tvar} = (\xridgenew)\inv \xmat\tp\vy.
\end{align}
A common choice for the regularization matrix is $\lammat = \lam
\mat{I}$. We also define some notation to be useful later on:
\begin{align}
\label{eq:xmatrix}
  \xtx = \mat{U} \mat{D} \mat{U}\tp \qtext{where} \mat{D} =
  \diag(\evali[1], \ldots, \evali[d]),
\end{align}
where for $d>n$, we use the convention that $\evali =0 $ if $i>n$.
Here the matrix $\mat U \in \real^{\dims \times \dims}$ denotes the
(complete set of) eigenvectors of the matrix $\xtx$.

Next, we briefly describe kernel ridge regression. Consider a
reproducing kernel $\kernel$ and the corresponding reproducing kernel
Hilbert space (RKHS), i.e., for all $x\in \real^d$, $\kernel(x, \cdot)
\in \rkhs$ and for any $f \in \rkhs$, we have $\doth{f, \kernel(x,
  \cdot)} = f(x)$. In kernel ridge regression, given the data
$\braces{(x_i, y_i)}_{i=1}^n$, we need find an estimate $\widehat{f}
\in \rkhs$ such that $\widehat{f}(x_i) \approx y_i$. The corresponding
estimate is given by
\begin{align}
  \label{eq:kernel_ridge}
  \widehat{f} \defn \argmin_{f\in \rkhs}
  \parenth{\frac{1}{2}\enorm{\v{y} -\v{f}_1^n}_2^2 + \frac{\lam}{2}
    \enorm{f}_{\rkhs}^2 } \qtext{where} \v{f}_1^n \defn (f(x_1),
  \ldots, f(x_n)) \tp,
\end{align}
and $\lam > 0$ denotes a regularization parameter. The representer theorem
for reproducing kernels implies that to solve~\eqref{eq:kernel_ridge}
it suffices to consider functions of the form $f(\cdot) =
\sum_{i=1}^n\beta_i \kernel(x_i, \cdot)$ for $\beta \in
\real^n$. Substituting this functional form back in
\cref{eq:kernel_ridge} yields a regularized least-squares problem from
$\tvar$ which admits a closed-form:
\begin{align}
  \label{eq:kernel_ridge2}
  \betahat_{\lam}(\vy) \defn \argmin_{\beta\in \Rn}
        \parenth{\frac{1}{2}\enorm{\v{y}-\kmat\beta}_2^2 +
          \frac{\lam}{2}\beta\tp\kmat\beta} = (\kridge)\inv \vy.
\end{align}
where $\kmat$ is the $n\times n$ kernel matrix with
$\kmat_{ij}=\kernel(x_i, x_j)$. With this fit, one can recover the
estimates for in sample observations as $\widehat{\vy} =
\mat{K}\betahat_{\lam}$, and for any new point $x$, the estimate is
given by $\widehat{f}(x) = \sum_{i=1}^n[\betahat_{\lam}]_i
\kernel(x_i, x)$.

\subsubsection{Defining ridge-based LNML codes}
\label{sub:lnml_code_defn}
We start with the linear model setting and then discuss the kernel
setting. In accordance with MDL convention, we assume that the design
matrix $\xmat \in \real^{\obs \times d}$ is known, and only the
information in the response vector $\vy \in \real^\obs$ is to be
encoded.

\paragraph{LNML codes for linear methods:}
We use the notation:
\begin{align}
\label{eq:pdist}
g(\vy; \xmat, \lammat, \tvar) = \frac{1}{(2 \pi \noisevar)^{n/2}}
e^{-\frac{1}{2\noisevar} \enorm{\vy-\xmat\tvar}^2} \cdot e^{
  -\frac{1}{2\noisevar} \tvar\tp\lammat\tvar} , \qtext{for} \vy \in
\real^n, \tvar \in \Rd.
\end{align} 
Next, we define the first term in \cref{eq:pdist} as the code $\pdata$ for
data fit, and the second term $(e^{-\frac{1}{2\noisevar}
  \tvar\tp\lammat\tvar})$ as the luckiness factor, i.e., 
  \begin{align}
  \begin{split}
  \label{eq:data_luck_def}
\pdata(\vy; \xmat, \tvar) &\defn \frac{1}{(2 \pi \noisevar)^{n/2}}
e^{-\frac{1}{2\noisevar} \enorm{\vy-\xmat\tvar}^2}
  \qtext{and} \\
  \luck(\tvar) = \plam(\tvar) &:= e^{-\frac{1}{2\noisevar}
  \tvar\tp\lammat\tvar}
  \end{split}
\end{align}
and comparing with \cref{eq:lnml}, we define the LNML code $\qdist_{\lammat}$ as follows: $\qdist_{\lammat}$ is the a distribution over $\real^n$, that admits the density $\qden_{\mrm{LNML}}$ given by
\begin{align}
\label{eq:ridge_dist}
\qden_{\mrm{LNML}}(\vy) = \qden_{\lammat}(\vy) = \frac{g(\vy; \xmat,
  \lammat, \that_{\lammat}(\vy))}{\normconst} \qtext{where} \normconst
:= \displaystyle\int_{\real^n} g(\v{z}; \xmat, \lammat,
\that_{\lammat}(\v{z}))d\v{z}.
\end{align}
It is a valid density for an LNML code, denoted by $\qdist_{\lammat}$,
since
\begin{align}
\label{eq:lnml_ridge}
\argmax_{\tvar} \parenth{\pden_{\mrm{data}}(\v{y}; \xmat, \tvar) \cdot
  \plam(\tvar)}  \stackrel{\eqref{eq:ridge_new_closed}}{=}
\that_{\lammat}(\v{y}).
\end{align}
In other words, $\qdist_{\lammat}$ is the LNML code induced by the
ridge estimator $\that_{\lammat}$.\footnote{In a similar fashion, one can see that the OLS estimator would correspond to the NML code~\eqref{eq:nml_defn} for the codes given by~\eqref{eq:linear_code}.} We note that the RHS of \cref{eq:pdist} denotes (up to proportionality)  the posterior density on $\theta$ for Gaussian likelihood on data with a Gaussian prior; however, the distribution $\qdist_{\lammat}$ is not central to the Bayesian settings.

Our definition of complexity, as alluded to earlier, makes use of a family of LNML
codes. For the linear setting, we consider the family:
\begin{align}
\label{eq:lnml_family_lin}
\qsetlin \defn \!\braces{\qdist_{\lammat}\!:\!\lammat\! \in\!
  \mc{M}}, \text{ where } \mc{M}\!:=\!\braces{\mat U \diag (\lam_1,
  \ldots, \lam_d)\mat U\tp \vert \lam_j \geq 0, j\!=\!1,\ldots d},
\end{align}
where $\mat U$ denotes the eigenvectors of the matrix
$\xtx$~\eqref{eq:xmatrix}. We note that the discussion to follow depends on the choice of luckiness functions as well as the consequent choice of $\mc M$. While our choice of luckiness function is akin to the choice of conjugate prior in Bayesian settings, namely for analytical tractability, the consequent choice of $\mc M$ makes a particular tradeoff between the codelength needed to encode data and the regularization parameters. See the discussion below, \cref{sub:reg_set}, and \cref{rem:choices}.

Note that the family $\qsetlin$ is not parametric in the classical sense (as it is not
indexed by the canonical parameter $\theta$).  However, since the LNML
codes are induced by the ridge estimators, selecting a particular
$\qdist_{\lammat}$ code corresponds to choosing a particular $\lammat$
based-ridge estimator for all $\vy$. Moreover, this family can be seen as a family of LNML
codes induced due to different choices of the luckiness functions in \cref{eq:lnml}.

A key difference from the NML setting is that the LNML codes are
indexed by hyper-parameter $\lammat$, and thus it remains to add the
codelength corresponding to the index. To this end, we use a simple
quantization-based encoding for the hyper-parameters
$\braces{\lambda_j}$, in the spirit of the two-stage encoding for the
hyper-parameters as in prior
works~\citep{barron1998minimum,hansen2001model,grunwald2007minimum}. In particular, using $\mc L$ to denote the codelength corresponding the regularization parameters, we use
\begin{align}
   \mc L(\lammat) &:=\frac12\sum_{\lambda_i>0} \log\lambda_i,
  \label{eq:lamopt_claim}
\end{align} 
where the factor $\frac12$ is chosen for simplifying our analytical expressions in \cref{thm:complexity_expressions}
and can replaced by $1$ without changing the qualitative conclusions (see \cref{sub:mult_factor_1}). 

Note that the expression~\eqref{eq:lamopt_claim} does not include any  codelength to share the indices of $\braces{i:\lambda_i > 0}$ for the following reasons. In the proof of \cref{thm:complexity_expressions}, we show that $\lambda_i>0$ if and only if $\evali>0$ (for our analytical derivations for linear model settings). These indices are known both to sender and receiver since $\xmat$ is assumed known to both parties so the knowledge of what indices correspond to positive eigenvalues of $\xmat\tp\xmat$ is also apriori shared. Moreover in practice, we often only use one regularization parameter $\lambda_i=\lambda$ so that no additional coding of any index is needed. If one uses multiple $\lambda_i$'s in practice as in \cref{eq:lnml_family_lin}, we can once again argue that for any practical choice, $\lambda_i=0$ if an only if $\evali=0$, so that no further addition of codelengths for indices is needed. We provide a further discussion of our choice of codelength for $\lammat$, and the set $\mc M$ in \cref{sub:further_our_mdl}.

\paragraph{LNML codes for kernel methods:}

Given the estimators~\eqref{eq:kernel_ridge} and
\eqref{eq:kernel_ridge2}, one can define the LNML codes given a set of
points, the kernel and the corresponding kernel matrix as follows.  We
define the function $h(\cdot; \beta, \kmat)$
\begin{align}
\label{eq:k_pdist}
h(\vy; \beta, \mat{K}) \defn \frac{1}{(2 \pi \noisevar)^{n/2}}
e^{-\frac{1}{2 \noisevar} \enorm{\vy - \kmat \beta}^2} \cdot
e^{-\frac{\lam}{2 \noisevar} \beta \tp \kmat \beta},
\end{align} 
Letting the first term in \cref{eq:k_pdist} denote the code $\pdata$
for data fit, and the second term $(e^{-\frac{\lam}{2\noisevar}
  \beta\tp \kmat\beta})$ as the luckiness factor $\luck$, and arguing
similar to \cref{eq:ridge_dist,eq:lnml_ridge}, we conclude that the
LNML code is given by
\begin{align}
\label{eq:k_ridge_dist}
   \widetilde{\qden}_{\lam}(\vy) = \frac{h(\vy; \betahat_{\lam}(\vy),
     \mat{K})}{\kconst} \qtext{where} \kconst \defn
   \displaystyle\int_{\real^n} h(\v{z}; \betahat_{\lam}(\v{z}),
   \mat{K})d\v{z},
\end{align}
and let $\widetilde{\qdist}_{\lam}$ denote the distribution
corresponding to $ \widetilde{\qden}_{\lam}$.  Finally, our definition
of complexity for kernel methods makes use of the following family of
LNML codes:
\begin{align}
\label{eq:lnml_family_ker}
\qsetker = \braces{\widetilde{\qdist}_{\lam} : \lam \geq 0}.
\end{align}
In contrast to the linear setting, here we consider the family of
codes indexed by a single parameter, as using multiple $\lambda$'s
would break the equivalence between
\cref{eq:kernel_ridge,eq:kernel_ridge2}, unless we alter the Hilbert
norm that is being penalized in \cref{eq:kernel_ridge}.

\subsection{Defining MDL-COMP via ridge-LNML codes}
\label{sub:mdl_comp_defn}

We are now ready to define the MDL-based complexity. We define the
complexity as optimal redundancy over the LNML codes with certain
generative assumptions for the data.

\paragraph{MDL-COMP for linear models:}
We consider the generative model
\begin{align}
 \label{eq:linear_model_revisit}  
 \yi = \axi \tp \tstar + \noise, \qtext{for} i = 1, 2, \ldots, n,
 \qtext{or} \vy = \xmat \tstar + \v{\noise[]},
\end{align}
where we assume that $\noise[]\sim \mc N(0, \noisevar \mat I_n)$, so
that $\pdist_\star = \ptrue= \mc{N}(\xmat \tstar, \sigma^2 \mat I_n)
$.\footnote{Although we later discuss a minimax setting while relaxing
this assumption on the noise, see \cref{thm:minimax_codelength}.}
Given this notation, the $\mdltag$ for this setting is defined as the
sum of the optimal redundancy over the codes
$\qsetlin$~\eqref{eq:lnml_family_lin}, and the codelength needed to
encode the optimal hyperparameters:
\begin{subequations}
\begin{align}
    \label{eq:mdl_linear}   
\mdltag(\ptrue, \qsetlin) & \defn \frac{1}{n} (\ropt(\ptrue, \qsetlin)
+ \mc L(\lammatopt)), \qtext{where} \\
\label{eq:ropt_linear}    
\ropt(\ptrue,\!\qsetlin) & \defn \min_{\qdist \in \qsetlin}
\kull{\ptrue}{\qdist} =\min_{\lammat \in
  \mc{M}}\kull{\ptrue}{\qdist_{\lammat}},
\end{align}
\end{subequations}
and $\lammatopt$ denotes the $\arg\min$ in \cref{eq:ropt_linear}. 

\begin{remark}
   We note that for the definitions above, in principle, $\ptrue$ can be replaced by an arbitrary (and not necessarily even linear) generative model, and MDL-COMP would still be a valid complexity measure---as it measures the best possible excess number of bits over the class $\qsetlin$ for encoding data generated by $\ptrue$. However, for establishing analytical results in the following section, we restrict ourselves to a linear generative model $\ptrue$ while analyzing MDL-COMP with linear fitted models.
 \end{remark} 

\paragraph{MDL-COMP for kernel methods:}

We assume throughout this paper that the reproducing kernel $\kernel$
is a Mercer kernel~\citep{mercer1909functions}, which admits the
eigen-expansion:
\begin{align}
\label{eq:mercer_kernel}
  \kernel(x, y) = \sum_{k=1}^{\infty} \mu_k \phi_k(x) \phi_k(y),
  \qtext{for all} x, y \in \real^d,
\end{align}
where $\mu_1 \geq \mu_2 \cdots \geq 0$ denotes the sequence of
(non-negative) eigenvalues of the kernel and
$\braces{\phi_k}_{k=1}^\infty$ denotes the associated eigenfunction
taken to be orthonormal in $\mathbb{L}^2(\nu)$ for a suitably chosen
distribution $\nu$.\footnote{In the model below, we assume $\nu$ is the
marginal distribution of the covariates.} Let $\rkhs$ denote the
reproducing kernel Hilbert space of the kernel $\kernel$.  We consider
the generative model
\begin{align}
\label{eq:kernel_Generative_model}
y_i = \fstar(x_i) + \noise, \qtext{for} i = 1, 2, \ldots, n,
\qtext{or} \vy = \v{(\fstar)}_1^n+ \v{\noise[]}.
\end{align}
where we assume that $x_i$'s are drawn i.i.d. from the distribution
$\nu$, and $\fstar \in \rkhs$, and use the notation $\v{(\fstar)}_1^n
= (\fstar(x_1), \ldots, \fstar(x_n))\tp$. We also assume that
$\noise[]\sim \mc N(0, \noisevar \mat I_n)$, so that $\pdist_\star =
\mb{P}_{\fstar} = \mc{N}((\v{\fstar})_1^n, \sigma^2 \mat
I_n)$.\footnote{We can relax this assumption (without altering the
guarantees derived later) to the noise being zero mean, with variance
$\noisevar$ and being uncorrelated with $\v{(\fstar)}_1^n$.}  With
this notation, we define MDL-COMP for the kernel setting as the
optimal redundancy over the codes $\qsetker$ in the
family~\eqref{eq:lnml_family_ker}:
\begin{align}
\label{eq:mdl_kernel_defn}
     \mdltag(\mb{P}_{\fstar},\!\qsetker) \defn
     \frac{1}{n}\min_{\widetilde{\qdist}\in\qsetker}\kull{\ptrue}{\widetilde{\qdist}}
     \defn \frac{1}{n} \min_{\lam \geq 0}
     \kull{\mb{P}_{\fstar}}{\widetilde{\qdist}_{\lam}}.
\end{align}
where $\kmat=(\kernel(x_i, x_j))_{j=1}^{n}$ denotes the kernel matrix at the observed covariates. Like in the linear setting, one can replace $\mb{P}_{\fstar}$ by an arbitrary generative model and MDP-COMP would continue to be a valid measure of complexity. But for the analytical derivations in the sequel we restrict our attention to generative models of the form~\eqref{eq:kernel_Generative_model} when dealing with kernel methods. Notably, unlike in \cref{eq:mdl_linear}, we do not add a codelength for $\lambda$ in \cref{eq:mdl_kernel_defn}. See

\subsection{Further discussion on \mdltag}
\label{sub:further_our_mdl}

In this section, we provide additional discussion and put the different choices made in our definitions in the context of prior work.
 
\subsubsection{Relation of LNML code~\eqref{eq:lnml_ridge} with prior work}
By definition~\eqref{eq:lnml_ridge}, we have
\begin{align}
\label{eq:log_1_q}
    -\log(\qden_{\lammat}(\vy)) = -\log \parenth{\pden_{\mrm{data}}(\v{y}; \xmat, \that_{\lammat}(\v{y}))}  
    - \log(\plam(\that_{\lammat}(\v{y}))) + \log \normconst.
\end{align}
Notably, the expression on the RHS of \cref{eq:log_1_q} is identical to that in \citet[Eq.~(6)]{grunwald2019minimum}, which is derived as the quantity of interest that characterizes the goodness-of-fit with the complexity of the MDL estimator. Moreover, same expression arises for a more general framework in \citet[Eqn.~(52), arxiv version]{grunwald2017tightarxiv}, while considering regularized ERM estimators under a unified treatment of MDL and several standard complexities.\footnote{\label{footnote:review}We were made aware of these works by the reviewers.} However in both these works, analytical expressions for linear (or kernel) models with squared loss especially for overparameterized models are not considered. Here to make progress, we use the expected value of the quantity $-\log(\qden_{\lammat}(\vy) )$ under the true distribution of the data to compute the redundancy~\eqref{eq:redundancy},
and then use the minimum possible redundancy over $\lammat$ to measure our complexity measure \mdltag. Overall, our choice for the LNML code~\eqref{eq:log_1_q} is consistent with several prior works, and here we further enhance the understanding of such a choice with several analytical and experimental investigations.

\subsubsection{The need for a true generative model}
We highlight that \cref{eq:log_1_q} relies only on a \emph{posited linear model} for the observed data. It is only to define the redundancy expression in \cref{eq:redundancy}, we assume a \emph{generative model} associated with a true parameter $\tstar$\footnote{The subsequent calculations for the linear setting assume a generative linear model but \cref{eq:redundancy} assumes just some generative model.}. The latter assumption is necessary for analytical derivations, and in the sequel we assume a linear generative model. However, for a given dataset our framework does not really require a generative model, as indeed \cref{eq:log_1_q} does not rely on a true linear model. Thus in practice,  in accordance with the MDL principle, we directly minimize the codelength~\eqref{eq:log_1_q} over the choices of $\lammat$, and call it the practical version of MDL (\pracmdltag) that is also used in our experiments to tune the ridge regularization hyper-parameter (see \cref{sec:experiments,eq:mdl_comp_objective}). As noted above such a recommendation is also consistent with the recommendation made in \citet[Eq.~(6)]{grunwald2019minimum} for selecting the best MDL-estimator when tuning over finitely many choices of $\lammat$ with a uniform prior over $\lammat$.

\subsubsection{Regularization set $\mc M$: The tradeoff between expressivity and codelength} 
\label{sub:reg_set}
Our definition~\eqref{eq:lnml_family_lin} makes use of the
eigenvectors of the matrix $\xtx$ to define the set $\mc M$ of all
possible $\lammat$ that we consider for the LNML codes. In simple words, we
assume that the regularization matrix $\lammat$ and the covariance matrix $\xtx$ are simultaneously orthogonally diagonalizable. Such a choice tries to address two concerns:
First, having a rich set of codes for analytical derivations is essential to provide us a better
understanding of how much compression in the data is possible, so that
having richer set than $\braces{\lambda \mat I; \lambda \geq 0}$ (the common choice in practice) is desirable. On the other hand, if we make the set  $\mc M$ too large, e.g., if it is the set of all positive semi-definite matrices, then we would defeat the purpose of measuring the complexity of data, as the bits needed to encode an arbitrary PSD matrix for linear model scale as $\order{\min\braces{d^2, n^2}}$, which
would just overwhelm the bits needed to encode the data itself. In
the prior works with MDL, while deriving analytical expressions, much simpler choices like $\lammat = c\xtx$ have been
made~\citep{hansen2001model}, and the bits needed to transmit the scalar $c$ have been treated as fixed (as we do in \cref{eq:mdl_kernel_defn,sub:mdl_complexity_kernel_methods} for kernel methods when we only have one hyper-parameter; also see \cref{rem:lam_kernel}). 

For our choice of $\mc M$, since matrix $\mat U$ and the indices $\mc I \defn \braces{j:\lambda_j>0}$ can be computed from $\xtx$, we only need to count the bits needed to encode the hyper-parameters
$\braces{\lambda_j, j \in \mc I}$, for which we use \cref{eq:lamopt_claim}. Such a choice allows more flexibility in the fitted model without blowing up the codelength for the hyperparameter too quickly. (We note that other careful choices of $\mc M$ are also possible; also see \cref{rem:choices}.)

\subsubsection{Our LNML encoding vs one-part universal encoding}
For analytical derivations, we allow $\lammat$ to belong to a continuous family in \cref{eq:lnml_family_lin}. In such a setting, as noted in \citet[Sec.~2.3, Eqns.~(17,18)]{grunwald2019minimum}, we can associate a code length for $\lammat$ as well to define a \emph{meta universal} (LNML) code.\footnote{We were made aware of the recent one-part meta-universal distribution framework by the reviewers during the review process.}  where in the density~\eqref{eq:lnml} would be replaced by
\begin{align}
\label{eq:full_lnml}
    \qden(\vy)= \displaystyle\frac{\max_{\tvar, \lammat} \pdata(\vy; \xmat, \tvar) \cdot \plam(\tvar) \cdot \uppi(\lammat)}{\int_{\real^d} \max_{\tvar', \lammat'} (\pdata(\v{z}; \xmat, \tvar') \cdot \pden_{\lammat'}(\tvar') \cdot \uppi(\lammat')) d\v{z}}
\end{align}
for a suitable luckiness function $\uppi$ over the space of $\lammat$ being considered. Such a one-part code was introduced for unifying model selection and estimation in \citet[Sec.~2.3, Eqns.~(17,18)]{grunwald2019minimum} as it jointly tries to identify the best possible $\tvar$ (estimation) as well as the regularization parameter $\lammat$ (model selection). With such a choice, the codelength would be indexed by the choice of the luckiness function $\uppi$ on the matrices $\lammat$, that we have to choose. (Notably, for the LNML code~\eqref{eq:full_lnml} to be well-defined, the $\uppi$ should be such that the denominator in display~\eqref{eq:full_lnml} is finite; ootherwise we still have an infinity problem like for the NML in \cref{eq:nml_defn}.)

Instead of searching for a suitable prefix code on $\lammat$, our treatment for MDL in \cref{eq:mdl_linear} instead involves adding a codelength after optimizing redundancy in \cref{eq:mdl_linear} using a simple quantization scheme~\eqref{eq:lamopt_claim}. While a convenient choice for analytical calculations, this choice is also well-motivated by practical scenarios, when one typically deals with finitely many $\lammat$, in which case a uniform prior over them, i.e., $\uppi(\lammat) \propto 1$ is a reasonable choice (\cite{grunwald2019minimum}).
Nevertheless in the continuous case a uniform prior is not well-defined and thus our choice warrants different interpretations: (A) using an improper luckiness function $\uppi(\lammat) \equiv 1$ so that the expression from \cref{eq:full_lnml} degenerates to \cref{eq:lnml} and posthoc compensation with the codelength~\eqref{eq:lamopt_claim}, or (B) as a direct analog of the treatment of encoding hyperparameters in crude two-stage  MDL~\citep{barron1998minimum,hansen2001model,grunwald2007minimum}. On the other hand, our theoretical results also provide two indirect justifications for our choice: The $\lammatopt$ achieving the minimum in \cref{eq:ropt_linear} is indeed an \emph{optimal} choice of regularization matrix $\lammat$ on two ends: (a) it minimizes the in-sample mean squared error (\cref{thm:lam_opt}), and (b) it achieves the minimax codelength over a class of noise distributions with bounded variance (\cref{thm:minimax_codelength}).

\subsubsection{Codelength for discretized $\lammat$}
\label{sub:codelength}
Our codelength for $\lammat$ in \cref{eq:lamopt_claim} is in fact an approximation itself. In principle, we want to encode $\lambda_i$ as an integer $ \ceil{\lambda_i/\Delta}$ for some small enough resolution $\Delta$; and
$\Delta=\frac{1}{\sqrt{n}}$ is often the default choice. To encode integers, \cite{rissanen1983universal} shows that the best possible universal codelength takes the form $\log^{\star}\ceil{\lambda_i/\Delta} + C$, where  $C \approx 1.52$ is a universal constant and for any integer $k\in\mathbb N$, we have $\log^{\star}(k) := \log_2 k + \log_2 \log_2 k + \cdots$~\citet[Eqn.~(4)]{lee2001intrductionto}. For the linear models, using this exact universal codelength would yield to an additive term of order $\frac{d}{2n}\log n$ in \mdltag\ expressions in \cref{thm:complexity_expressions} for $d<n$, and $\log n$ for $d>n$. As the typical scaling of MDL is indeed of order $d/n$ for $d<n$, and $\log d$ for $d>n$ (see \cref{fig:mdl_gaussian} and \cref{sub:sketch_gaussian}), such an adjustment does not alter the scaling suggested by the current \mdltag\ expressions, and hence we continue to use the approximate $\log \lambda_i$ codelength in our subsequent discussion. 

\begin{remark}
  \label{rem:lam_kernel}
  For the kernel setting, we do not add another $\lam$ dependent codelength in \cref{eq:mdl_kernel_defn}, as there is only one regularization parameter in kernel regression; such a treatment is akin to the assuming fixed number of bits for transmitting the scalar $c$ in prior works with MDL when using simpler choices like $\lammat = c\xtx$ (see \citep{hansen2001model}. Moreover, adding a term $\log(\lambda_\textrm{opt})/n$ on the RHS of \cref{eq:mdl_kernel_defn} does not alter the qualitative conclusions in Theorem~\ref{thm:mdl_kernel} or quantitative conclusions in Corollary~\ref{cor:mdl_kernel_smooth}.
\end{remark}

\begin{remark}
\label{rem:choices}
    Our choices, namely (a) the luckiness function, (b) the class of hyper-parameters in the luckiness function, and (c) how we encode these hyper-parameters directly impact our complexity calculations to follow. For example, a different choice of $p_{\lammat}$, or $\qsetlin$, or  a different codelength (instead of \cref{eq:lamopt_claim}) would lead to a different complexity measure; and the true complexity should be defined as the infimum complexity across all possible choices. Thus, our complexity measure (\mdltag) should be viewed as an upper bound on the true MDL complexity, which can be loose, however, as we demonstrate in the sequel, it is tighter than the naive parameter count when the linear model is overparameterized.
\end{remark}

\section{Main results}
\label{sec:main_results}
We are now ready to state our main results. We start with an explicit characterization of MDL-COMP for linear models, and its consequences in \cref{sub:mdl_complexity_for_linear_models}. We then characterize it for kernel methods and unpack the consequences in \cref{sub:mdl_complexity_kernel_methods}.

\subsection{Characterizing MDL-COMP for linear models} %
\label{sub:mdl_complexity_for_linear_models}
Our first result provides an explicit expression for
MDL-COMP~\eqref{eq:mdl_linear} for the linear models.
\begin{theorem}
\label{thm:complexity_expressions}
For the linear model~\eqref{eq:linear_model_revisit}, let $\mat U$ and
$\braces{\evali}$ denote the eigenvectors and eigenvalues of $\xtx$, 
 define the vector $\v{\myvec} \defn \mat U\tp \tstar$, and recall that $\sigma^2$ denotes the noise variance.  Then the
MDL complexity~\eqref{eq:mdl_linear} and the optimal
redundancy~\eqref{eq:ropt_linear} are given by
\begin{subequations}
  \begin{align}
  \label{eq:mdl_linear_explicit}  
  \mdltag(\ptrue, \qsetlin) & = \frac{1}{2n}\sumn[\minnd]
        \log\parenth{\evali + \frac{\noisevar}{\myvec_i^2}}, \quad
        \mbox{and} \\
    \label{eq:ropt_lin_model}   
    \ropt(\ptrue,\!\qsetlin) & = \frac{1}{2n}\sumn[\minnd] \log
    \parenth{ 1 + \frac{\evali\myvec_i^2}{\noisevar}}.
  \end{align}
\end{subequations}
\end{theorem}
\noindent See \cref{sub:proof_of_thm:complexity_expressions} for the
proof.  
We recall that the NML complexity for over-parameterized settings is typically infinite, or just a function of the volume of the space when one truncates the space of observation for theoretical analysis (\cref{sub:nml}). Even in the latter case, for overparameterized setting, the complexity does not depend on the design matrix. Our notion of MDL-COMP on the other hand, as seen by \cref{thm:complexity_expressions}, is not merely a parameter count or a
simple function of $d$ and $n$. Rather, it depends on the interaction
between the eigenvalues of the covariance matrix $\xtx$, and the
rotated true parameter scaled by noise variance $\v{\myvec}/\sigma=
\mat U\tp \tstar/\sigma$.  The
expression~\eqref{eq:mdl_linear_explicit} is oracle in nature since it
depends on an unknown quantity, namely the true parameter $\tstar$ via
the relation $\v{\myvec} = \mat U \tp \tstar$.\footnote{\label{footnote:defining_w}When model is under-specified in terms of the features, i.e., $\xmat$ includes a subset of features needed to correctly specify the model~\eqref{eq:linear_model_revisit}, $\v{\myvec}$ is defined by considering the restricted version of $\tstar$; and when it is over-specified, i.e., $\xmat$ denotes a superset of features, $\v{\myvec}$ is defined by appending zeros to $\tstar$ as necessary. Refer to footnote~\ref{proof:thm1_under_over_specified}, and \cref{sub:mdl_plots_details} for further discussion.}

In \cref{sec:experiments}, we propose a
data-driven and \mdltag\ inspired hyper-parameter selection criterion
called \pracmdltag\ to tune the ridge hyper-parameter. Later in
\cref{sec:experiments}, we provide a data-driven approximation to
$\mdltag$ so that our proposed complexity can also be computed as a
practical complexity measure without requiring knowledge of $\tstar$.

Next, we discuss some consequences for linear models: We illustrate the scaling of \mdltag\ in various settings in
\cref{ssub:scaling_of_mdlcomp}, and then prove in
\cref{sub:mdl_complexity_versus_the_mmse} that \mdltag\ informs the fixed
design generalization error (see \cref{thm:lam_opt}). Furthermore, in
\cref{sub:a_minimax_optimality_via_mdltag}, we establish a certain minimax optimality property of the code that defines \mdltag\ (see
\cref{thm:minimax_codelength}). We turn to MDL-COMP for kernel methods in \cref{sub:mdl_complexity_kernel_methods}.

\subsubsection{Scaling of \mdltag\ for various covariate designs}
\label{ssub:scaling_of_mdlcomp}
\begin{figure}[ht]
    \centering
    \includegraphics[width=0.9\textwidth]{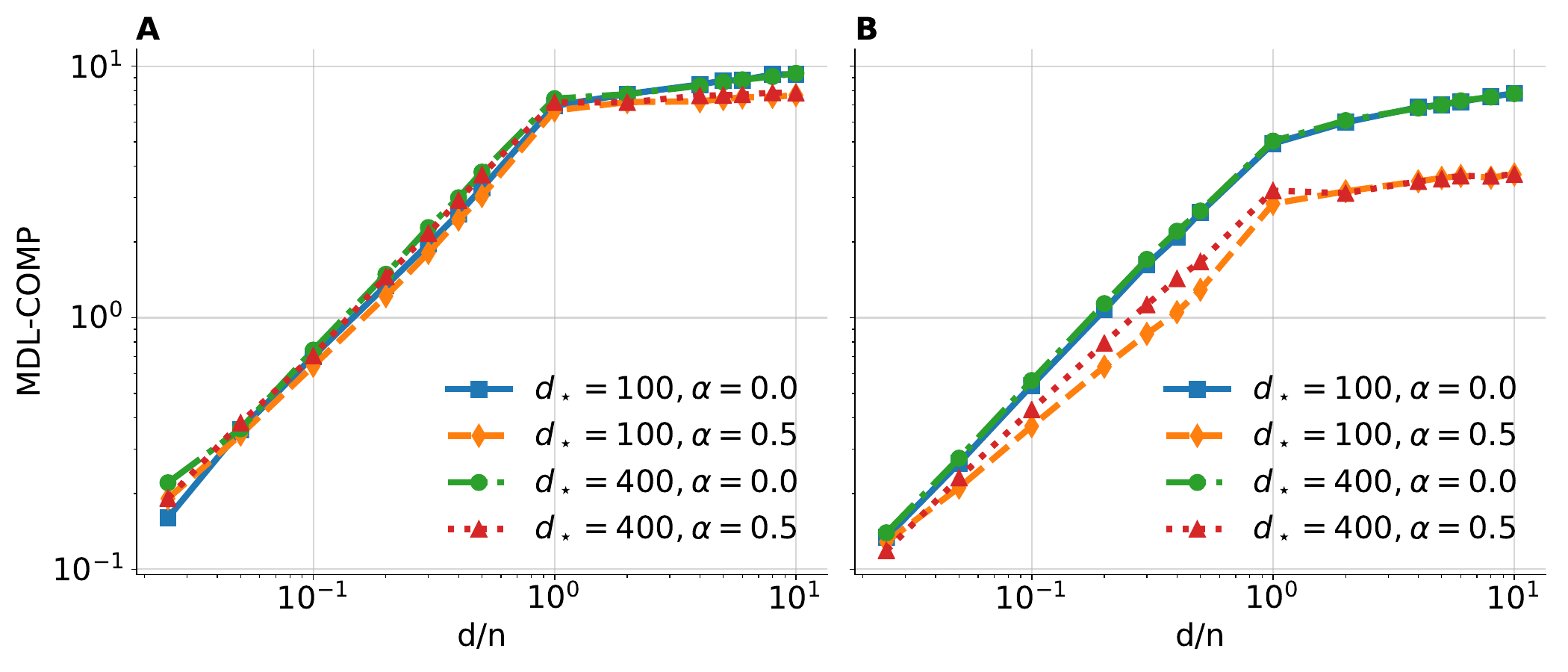}
    \caption{Scaling of MDL-COMP for Gaussian design. \textbf{(A)} $\sigma^2=1$, and \textbf{(B)} $\sigma^2=0.01$. We fix the generated data with $n=200$ samples, and vary the dimensionality $d$ of the covariates used for fitting the data. Here $\alpha$ denotes the decay of the eigenvalues in the covariance matrix for the covariates, and $d_\star$ denotes the true dimensionality of $\tstar$.
    } 
    \label{fig:mdl_gaussian}
\end{figure}

\begin{figure}[ht]
    \centering
    \includegraphics[width=0.9\textwidth]{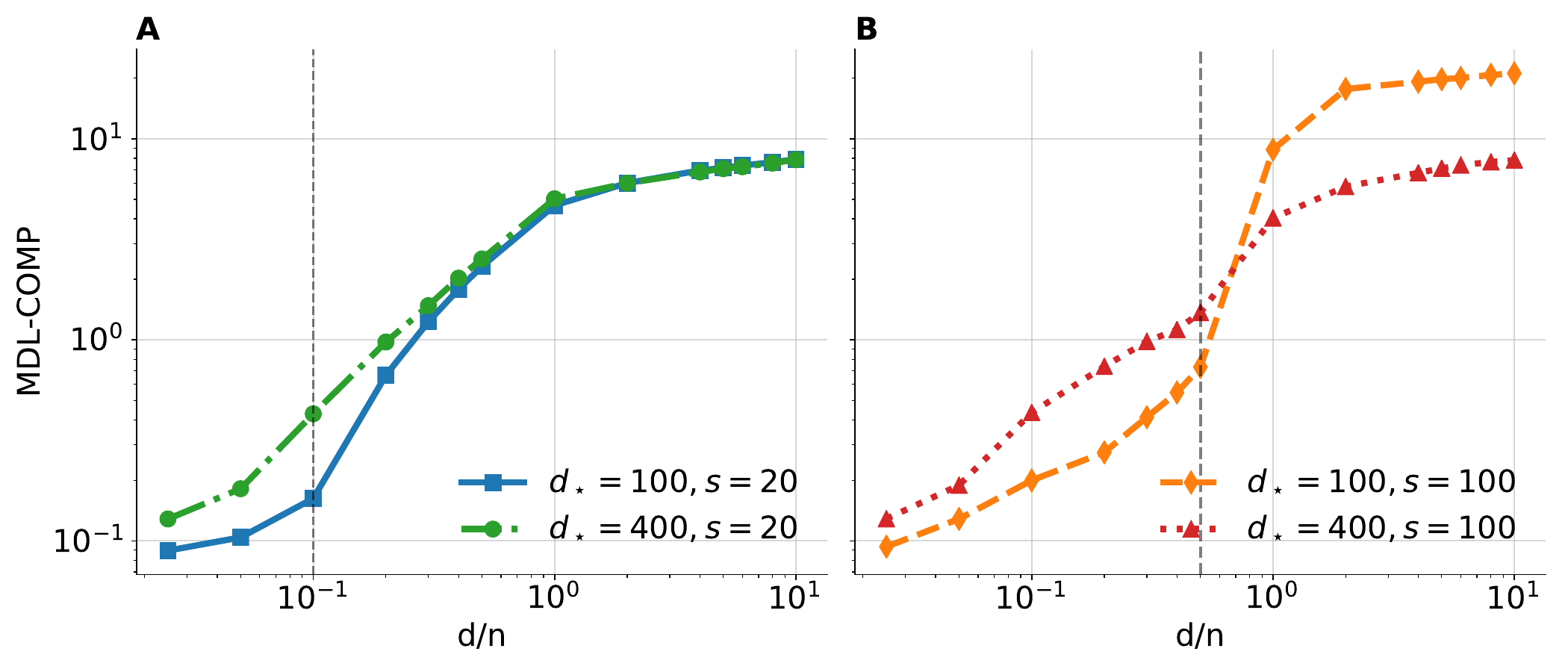}
    \caption{Scaling of MDL-COMP for spike design. We fix the generated data with $n=200$ samples, and $\noisevar=0.01$, and vary the dimensionality $d$ of the covariates used for fitting the data. Here $s$ denotes the spike dimension of the covariance matrix, and $d_\star$ denotes the true dimensionality of $\tstar$.
    } 
    \label{fig:mdl_spike_design}
\end{figure}

We now numerically compute \mdltag\ in several synthetic settings. Below we plot results for three different settings on the covariate design $\xmat$, and two different settings for the true parameter. In all cases, the rows of $\xmat$ are drawn from $\mc{N}(0, \Sigma)$. In \cref{fig:mdl_gaussian}, we consider two cases $\Sigma = \mat I_d$ (labeled as $\alpha=0$), and $\Sigma = \diag(1, 2^{-\alpha}, \ldots, d^{-\alpha})$ for $\alpha=0.5$. 
In \cref{fig:mdl_spike_design}, we choose a \emph{spike design}, where $\Sigma = \diag(16, 16, \ldots, 16, 1, \ldots, 1)$ with the first $s$ (spike dimension) diagonal entries taking value 16, and the rest taking value $1$. In both figures, we evaluate two different settings of $d_\star$ for the true dimensionality of $\tstar$, and select the true parameter $\tstar$ by drawing i.i.d. entries from standard normal, and then normalizing it to have norm $1$. Note that as we move from left to right on the x-axis in these figures, only the covariates used for fitting the model vary, and the generated data remains fixed (so that the model is under-specified for $d<d_\star$, and correctly (over) specified for $d \geq d_\star$). See \cref{sub:mdl_plots_details} for more details on the simulation set-up.

In both the figures, we note the non-linear scaling of MDL-COMP in the overparameterized regime ($d>n$). As we vary the dimension $d$ of the covariates used for computing \mdltag, in \cref{fig:mdl_gaussian}, we observe a linear scaling of \mdltag\ with $d$ for $d<n$, but a slow logarithmic or $\log d$ growth for $d>n$. On the other hand, the growth is clearly not linear even for $d<n$ for some of the spike design settings in \cref{fig:mdl_spike_design}. We provide further discussion on the set-up and scaling of MDL-COMP from these figures in \cref{sub:mdl_plots_details,sub:sketch_gaussian,sub:ropt_large_scale} (also see \cref{thm:rmt_expressions}).

\subsubsection{\mdltag\ informs fixed design mean-squared error}
\label{sub:mdl_complexity_versus_the_mmse}

Next, we show that $\mdltag$ (without the codelength for $\lammat$)
directly bounds the optimal fixed design prediction error (MSE). For
the training data points $\braces{(x_i, y_i)}_{i=1}^n$ generated from
a true model~\eqref{eq:linear_model_revisit} with true parameter $\tstar$, the
fixed design prediction MSE of an estimator $\that$ is given by
\begin{align}
  \label{eq:in_sample_mse}
    \textrm{fixed design pred. MSE} :=
    \frac{1}{n}\sum_{i=1}^{n}(x_i\tp\that-x_i\tp\tstar)^2 =
    \frac{1}{n}\enorm{\xmat \widehat{\theta} -\xmat \tstar}^2.
\end{align}
Note that this fixed design prediction error is very different from
the training MSE \mbox{$(\frac{1}{n}\enorm{\xmat \widehat{\theta}
    -\vy}^2)$.}  The in-sample MSE can be considered as a valid proxy
for the out-of-sample MSE (which in turn is usually estimated by test
MSE~\eqref{eq:test_mse}, when the out-of-sample points have the same
covariates as that of the training data; and it has been often used in
prior works to study the bias-variance trade-off for different
estimators~\citep{raskutti2014early}. Our next result shows that the
optimal redundancy $\ropt$ bounds the optimal in-sample MSE, and that
$\lammatopt$ that achieves $\ropt$ and defined \mdltag\ also minimizes
the in-sample MSE.  The reader should recall the
definition~\eqref{eq:ropt_linear} of $\ropt$ and $\lammatopt$.
\begin{theorem}
\label{thm:lam_opt}
For the ridge estimators~\eqref{eq:ridge_new_closed}, we have
\begin{subequations}
\begin{align}
  \Exs_{\noise[]}
  \brackets{\frac{1}{n}\enorm{\xmat\that_{\lammatopt}-\xmat\tstar}^2}
  \label{eq:opt_lam_insample_mse}
  & = \min_{\lammat \in \mc{M}}\Exs_{\noise[]}
  \brackets{\frac{1}{n}\enorm{\xmat\that_{\lammat}-\xmat\tstar}^2}
  \\
  & \leq \frac{\noisevar}{n}\ropt(\ptrue,\!\qset_\tagridge),
  \label{eq:ropt_bounds_insample_mse}
\end{align}
where $\Exs_{\noise[]}$ denotes the expectation over the noise variables $(\noise[1], \ldots, \noise[n])$ from \eqref{eq:linear_model_revisit}.
\end{subequations}
\end{theorem}
\noindent See \cref{sub:proof_of_thm:lam_opt} for the proof
of this claim. \\

Later, we show that in experiments, tuning the ridge model via a
practical (data driven) variant of $\mdltag$ can often minimize
out-of-sample MSE, and provide predictive performance that is
competitive with CV-tuned ridge estimator.

\subsection{Characterizing MDL-COMP for kernel methods}
\label{sub:mdl_complexity_kernel_methods}
We now turn to the non-linear settings, namely kernel methods. Our next result provides a bound on \mdltag\ for kernel methods. 
\begin{theorem}
\label{thm:mdl_kernel}
For the kernel setting~\eqref{eq:kernel_Generative_model}, the MDL complexity~\eqref{eq:mdl_kernel_defn} is bounded as
\begin{align}
    \label{eq:mdl_kernel}
    \mdltag(\mb{P}_{\fstar},\!\qsetker) \leq \inf_{\lam}\parenth{
    \frac{\lam}
    {2n} \frac{\enorm{\fstar}^2_{\rkhs}}{\sigma^2} + \frac{1}{2n}\sumn \log\parenth{\frac{\evali}{\lam}+1}}
\end{align}
where $\braces{\evali}_{i=1}^n$ denote the eigenvalues of the kernel
matrix $\kmat=(\kernel(x_i, x_j))_{i, j=1}^n$.
\end{theorem}
See \cref{sub:proof_of_theorem_thm:mdl_kernel} for the proof. Note that unlike \cref{thm:complexity_expressions}, here we establish an upper bound on MDL-COMP since closed-form expression does not exist. However, as we remark in the proof, we expect this bound to be tight for carefully constructed $\fstar$ for a wide range of behavior on $\braces{\evali}$.

The bound~\eqref{eq:mdl_kernel} in \cref{thm:mdl_kernel} is generic, and can be applied to any kernel setting including the neural tangent kernels that have recently been investigated in the theoretical literature on deep neural networks. Like \cref{eq:mdl_linear_explicit}, this expression also depends on the various problem parameters, including the signal-to-noise ratio $\frac{\enorm{\fstar}_{\rkhs}}{\sigma} $, and the eigenvalues of the kernel matrix. When additional information on the decay of the eigenvalues $\braces{\evali}$ is available, we can characterize a more refined bound on \mdltag\ as in the next corollary. For simplicity in exposition, we use $a_i \precsim b_i$ to denote that $a_i \leq c b_i$ for all $i$ with some universal constant $c$ independent of $i$.

\begin{corollary}
\label{cor:mdl_kernel_smooth}
Suppose $\kernel(x, x)=1$, $\evali$ denote the eigenvalues of the kernel matrix $\kmat$, and $\mathrm{SNR}\defn\frac{\enorm{\fstar}_{\rkhs}}{\sigma}>C$ for some universal constant $C$. Then, we have
\begin{align}
\label{eq:mdl_kernel_smooth}
    \mdltag(\mb{P}_{\fstar},\!\qsetker)
    \leq 
    \begin{cases}
      \displaystyle  \frac{d\log^2 (nd \cdot \mathrm{SNR}^2)}{n}
    &\text{ if }\evali \precsim n\exp(-i^{1/d}), \\
    \displaystyle C_{d, \omega, \mathrm{SNR}} \cdot \parenth{\frac{\log (n\cdot \mathrm{SNR}^2)}{n}}^{\frac{2\omega}{2\omega+d}}
    &\text{ if }\evali \precsim n i^{-2\omega/d}, \omega\!>\!d/2, \\[4mm]
    \displaystyle C_{d, a, \mathrm{SNR}} \cdot \parenth{\frac{\log (n\cdot \mathrm{SNR}^2)}{n}}^{\frac{d+a}{d+a+1}}
    &\text{ if }\evali \precsim ni^{-d-a}, d\!+\!a\!>\!1,
    \end{cases}
\end{align}
where the constants $ C_{d, \omega, \mathrm{SNR}},  C_{d, a, \mathrm{SNR}}$ are independent of $n$, and are defined in \cref{eq:constant_ker_smooth}.
\end{corollary}
See \cref{sub:proof_of_cref_cor_mdl_kernel_smooth} for the proof, where we also provide expressions for the constants appearing on the RHS above. We note that the $n$ appearing in the scaling of the eigenvalues is not an arbitrary assumption, but an immediate consequence of the fact that the trace of the kernel matrix is equal to $n$ since $\kernel(x, x) = 1$. We now contextualize the consequences of Corollary~\ref{cor:mdl_kernel_smooth} compared to prior work on kernel methods, and some recent work on neural networks. Contrary to the linear model setting, here our discussion focuses primarily on the classical non-parameteric setting in low dimensions, in particular assuming $n\gg d$. 

\paragraph{\mdltag\ informs minimax in-sample MSE for kernel methods when $n \gg d$}
The eigenvalue decay rate of order $\exp(-i^{1/d})$ and,
$i^{-2\omega/d}$ are known to be exhibited by Gaussian kernels and
reproducing kernels (like Mat\'ern kernels) for Sobolev spaces of smoothness $\omega$ in
$\real^d$ respectively (see
\citet[Thm.~15,16]{santin2016approximation}).  Up to logarithmic
factors, the scaling of \mdltag\ in \cref{eq:mdl_kernel_smooth} with
the sample size $n \gg d$ for these settings matches with the
minimax-optimal scaling of the in-sample risk
$\frac{1}{n}\sumn(\fstar(x_i)-\widehat{f}(x_i))^2$~\citep{stone1982optimal,raskutti2014early,wainwright2019high}.
Consequently, in such cases, the scaling of \mdltag\ is directly
informative of the minimax fixed design prediction error. This
matching of rates between \mdltag\ and the minimax error provides an
indirect justification for our choice of LNML
codes~\eqref{eq:lnml_family_ker} based on ridge estimators.

\paragraph{Applying MDL-COMP for neural tangent kernels}
We now briefly illustrate an example of how our theory can be used to approximately characterize the MDL-complexity for neural networks in certain settings.
A suitable class of kernels, namely, neural tangent kernels (NTK) have been used in recent years to understand the theoretical properties of deep neural networks~\citep{jacot2018neural,hayou2019meanfield,bietti2019inductive}. In particular, it has been established that for certain scaling of parameters of a deep neural network, as the width of the network goes to infinity, the function represented by DNNs converges to that of a function in the RKHS of a suitable kernel. As expected, the nature of these kernels are governed primarily by the number of layers, the non-linear activation function of DNN, and the input data distribution. More recently, several works have established a spectral characterization of these kernels, e.g., the eigenvalue decays at the rate $i^{-d-a}$ with $a=0$ for
deep neural tangent kernel (NTK) with ReLU activation functions, and
$a=\frac{1}{2^{L}}$ for an $L$-layer deep NTK with step activation
function (see~\citet[Cor.~2,3]{bietti2021deep}) when the covariates
are drawn uniformly from the $d$-dimensional unit sphere.
Combing these results with Corollary~\ref{cor:mdl_kernel_smooth} readily yields the MDL complexity bounds for associated kernels.
For instance, for NTK with ReLU activations, the bound~\eqref{eq:mdl_kernel_smooth} and 
the constant~\eqref{eq:constant_ker_smooth} from the proof show
that the MDL-COMP for NTK with ReLU activations scales as $\SNR^{2/(1+d)} \cdot \left((d\log n)/{n}\right)^{d/(d+1)}$, up to logarithmic factors.
Thus we note that for a fixed but large sample size $n \gg d$, this complexity can decrease as the dimensionality $d$ of the data increases, under the assumption that the change in $\SNR$ with dimension $d$ does not alter the scaling with respect to $n$.

\section{Experiments with data-driven MDL-COMP}
\label{sec:experiments}

This section proposes a practical version of \mdltag. Simulations and
real-data experiments show that this data-driven \mdltag\ is useful
for informing generalization. In the experiments to follow, this
data-driven \mdltag\ as a hyperparameter tuning criteria. While
\cref{thm:lam_opt} guarantees that the optimal regularization defining
(the oracle) $\mdltag$ also obtains the minimum possible in-sample
MSE, in this section we numerically illustrate the usefulness of the
\pracmdltag\ for achieving good test MSE which is computed on a fresh
set of samples $(x_i', y_i')_{i=1}^{\ntest}$ as follows:
\begin{align}
  \label{eq:test_mse}
\textrm{test-MSE} & \defn \frac{1}{\ntest} \sum_{i=1}^{\ntest} (y_i'-
\that \tp x_i')^2.
\end{align}
We start by defining \mdltag\ inspired hyper-parameter tuning  (model selection for regularization parameter) in \cref{subsec:mld_comp_practical} that we call \pracmdltag; see \cref{eq:mdl_comp_objective}. Then, in \cref{sub:test_mse} we demonstrate that
solving~\eqref{eq:mdl_comp_objective} correlates well with minimizing
test MSE (\cref{fig:vary_lambda}) for linear models. Next, in \cref{sub:real_experiments} we find that, for real datasets, \mdltag\ is competitive with cross-validation
(CV) for tuning regularization hyperparameter, and actually outperforming CV in the low-sample regime (\cref{fig:pmlb}). We note that CV is computationaly costlier to implement than since our method requires only training one model per choice (see end of \cref{subsec:mld_comp_practical}). In \cref{sub:fmri}, we evaluate our method on fMRI data and compare with two other baselines, Bayesian ARD and  BIC, besides CV and find from 
\cref{fig:fmri_results,fig:bic_fmri} that \pracmdltag\ outperforms the ARD and BIC, and remains competitive with CV (note that BIC and \pracmdltag\ have same order of computational cost). We also provide comparison of CV and our method using neural tangent kernels in \cref{fig:fmri_ntk_results} of \cref{sub:fmri_ntk}. Overall, we find that \pracmdltag\ provides a computationally efficient competitive alternative to CV for hyperpoarameter tuning for ridge regression (without losing on test error) across a range of real-world datasets.

In this section, the linear models (ridge) and kernel methods are fit using
scikit-learn~\citep{pedregosa2011scikit} and optimization for
hyper-parameter tuning (see~\eqref{eq:mdl_comp_objective}) is
performed using SciPy~\citep{2020SciPy-NMeth}.  Code and documentation for easily reproducing the results are provided  at
\href{https://github.com/csinva/mdl-complexity}{\faGithub~github.com/csinva/mdl-complexity}.

\subsection{\mdltag\ inspired hyper-parameter tuning}
\label{subsec:mld_comp_practical}

As defined, the complexity $\mdltag$ can not be computed in practice,
since it assumes knowledge of true parameter. Moreover, ridge
estimators are typically fit with only one regularization parameter,
shared across all features.  As an alternative that circumvents these
issues, we propose the following data-driven \emph{practical
MDL-COMP}:
\begin{align}
 \label{eq:mdl_comp_objective}
 \pracmdltag = \min_{\lam} \frac{1}{n} \parenth{\frac{\enorm{\xmat
       \that_{\lam}-\vy}^2}{2 \noisevar} + \frac{\lam \enorm{\that_{\lam}}^2}{2
     \noisevar} + \frac{1}{2} \sum_{1=i}^{\min\braces{n,d}} \log
   \parenth{1 + \frac{\evali}{\lam}}},
\end{align}
where $\that_{\lam} \defn \parenth{\xtx+\lam \mat I}\inv \xmat\tp \vy$ is the
ridge estimator~\eqref{eq:ridge_new_closed} for $\lammat = \lambda
\mat I_d$, and we use $\evali$ to denote the non-zero eigenvalues of
the matrix $\xtx$. This expression can be derived in two different ways: (i) The expression inside the minimizer is the $n$-sample ``plug-in'' estimate of the objective~\eqref{eq:ropt_linear} as shown in our proof in Theorem~\ref{thm:complexity_expressions} in
expression~\eqref{eq:exp_1} (up to a constant offset $1/2$), when we enforce the choice $\lammat = \lambda \mat I$. (ii) As discussed in \cref{sub:further_our_mdl}, minimizing the LNML code length~\eqref{eq:log_1_q} to tune the hyper-parameter and find a good linear model fit is justified under a uniform prior over finitely many choices of hyper-parameter, a setting common in practice. Using the definitions~\eqref{eq:data_luck_def}, \cref{eq:log_1_q}, and the expression for the normalization constant ($\normconst$) in \cref{eq:t3} from the proof of Theorem~\ref{thm:complexity_expressions} with $\lammat = \lambda \mat I$, we find that the objective~\eqref{eq:log_1_q} scaled by $n$ can be simplified as follows:
\begin{align}
 -\frac1n\log(\qden_{\lammat}(\vy)) &= -\frac1n\log \parenth{\pden_{\mrm{data}}(\v{y}; \xmat, \that_{\lammat}(\v{y}))}  
    - \frac1n\log(\plam(\that_{\lammat}(\v{y}))) + \frac1n\log \normconst  \\
    &=  \frac1n\parenth{\frac{\enorm{\xmat
       \that_{\lam}-\vy}^2}{2 \noisevar} + \frac{\lam \enorm{\that_{\lam}}^2}{2
     \noisevar} + \frac{1}{2} \sum_{1=i}^{\min\braces{n,d}} \log
   \parenth{1 + \frac{\evali}{\lam}}},
\end{align}
which indeed is same as the objective in \cref{eq:mdl_comp_objective} used to define \pracmdltag.

Moreover, using \cref{eq:ropt_expression} from the
proof of \cref{thm:complexity_expressions}, we can also obtain the plug-in estimate for optimal redundancy ($\ropt$) as follows:
\begin{align}
\label{eq:approx_ropt}
    \widehat{\mc{R}}_{\textrm{opt}} := \frac1{2n} \sum_{i=1}^{\minnd}\log\parenth{1+\frac{\evali}{\widehat{\lam}_{\textrm{opt}}}},
\end{align}
where $\widehat{\lam}_{\textrm{opt}}$ is the optimal hyperparameter for the objective~\eqref{eq:mdl_comp_objective}. Since we only have one hyper-parameter in defining \pracmdltag, we can also treat the approximate $\widehat{\mc{R}}_{\textrm{opt}}$ as a proxy for \mdltag\ over the class $\braces{\qdist_{\lam}: \lam \in (0, \infty)}$.

For the same reasons as linear methods, a reasonable data-driven MDL criterion for tuning $\lambda$ with kernel methods can be given by
\begin{align}
\label{eq:kernel_prac_mdl_comp}
    \pracmdltag_{\kmat} = \min_{\lam} 
    \frac{1}{n}\parenth{\frac{\enorm{\kmat \that_{\lam}-\vy}^2}{2\noisevar} + \frac{\lam \that_{\lam}\tp \kmat \that_{\lam}}{2\noisevar} 
    + \frac12 \sum_{i=1}^n \log \parenth{1+\frac{\evali}{\lam}}}
\end{align}
where for the kernel case, we use the notation $\that_{\lam} :=\! (\kridge)\inv \vy $~\eqref{eq:kernel_ridge}, and
$\{ \evali \}_{i=1}^n$ denote the eigenvalues of the kernel matrix~$\kmat$.

Finally, while our theoretical results assume that that $\sigma$ is known, in practice, often $\sigma$ is unknown. In such a setting, we can estimate $\sigma$ in the underparameterized as in least squares, namely, $\widehat{\sigma}^2 := \norm{\xmat\widehat{\theta}_{\textrm{OLS}}-\vy}^2/(n-d)$, where $\widehat{\theta}_{\textrm{OLS}}:= (\xtx)\inv\xmat\tp\vy$. Estimating the noise variance in the overparameterized settings is an active area of interest, and one possibility is to use a variance estimate from ridge regression with a suitable choice of hyperparameter~\citep[Eqn.~(2)]{liu2020estimation}. \cite{liu2020estimation} establish a consistency and central limit theorem for this estimate under a linear generative model with high-dimensional asymptotics as $d/n \to \tau$ for $\tau \in [0, \infty)$.\footnote{In all our simulations, we set $\sigma=1$, same as ground truth. For the FMRI experiments, the results with $\sigma=1$ were better than those obtained by estimating the variance in observations across 10 repetitions.}

\paragraph{Computational benefits of \pracmdltag\ over cross-validation:} %
\label{sub:run_time_comparisons_between_cross_validation_and_pra}
We note that for a given hyperparameter $\lambda$, implementing the \pracmdltag\ criterion requires (1) solving a regularized least squares problem of size $n \times d$, and (2) computing the eigenvalues of the matrix $\xtx$, both of which take $\order{dn^2}$ time when $d>n$, and $\order{nd^2}$ time when $n>d$ when using practical and stable numerical solvers. Thus the overall computational complexity is $\order{\min(n, d)^2 \max(n, d)}$. On the other hand, implementing $k$-fold cross-validation for a given $\lambda$ requires us to solve $k$ regularized least squares problem of size $\order{n} \times d$, and thereby the overall computational complexity is $\order{k \cdot \min(n, d)^2 \max(n, d)}$. In simple words, \pracmdltag\ is $\order{k}$ computationally more efficient than $k$-fold cross-validation.

\subsection{\pracmdltag\ informs test MSE in Gaussian simulations}
\label{sub:test_mse}

\begin{figure}[ht]
    \centering \includegraphics[width=\textwidth]{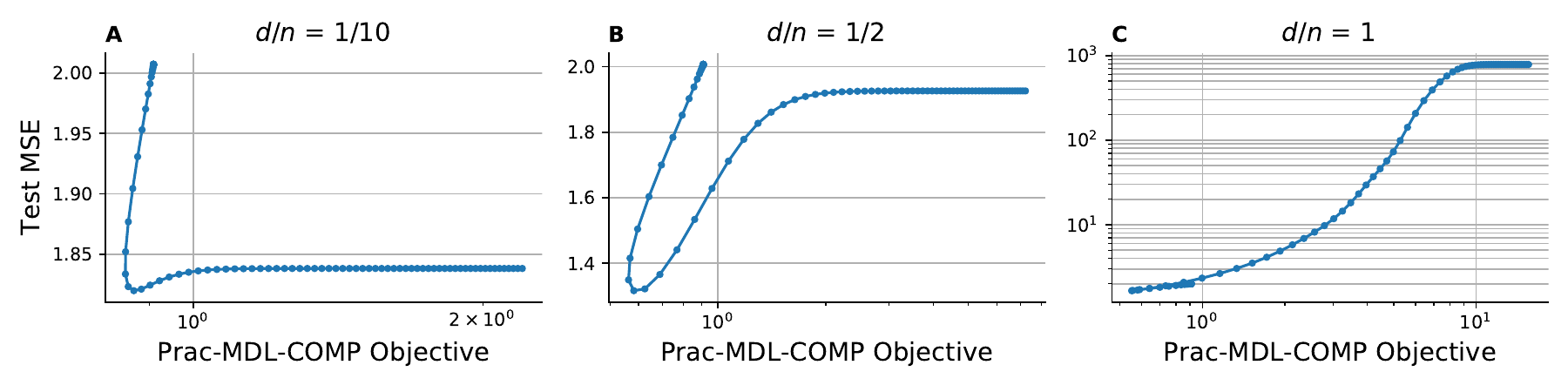}
    \makebox[\textwidth][c]
            {\includegraphics[height=1.37in]{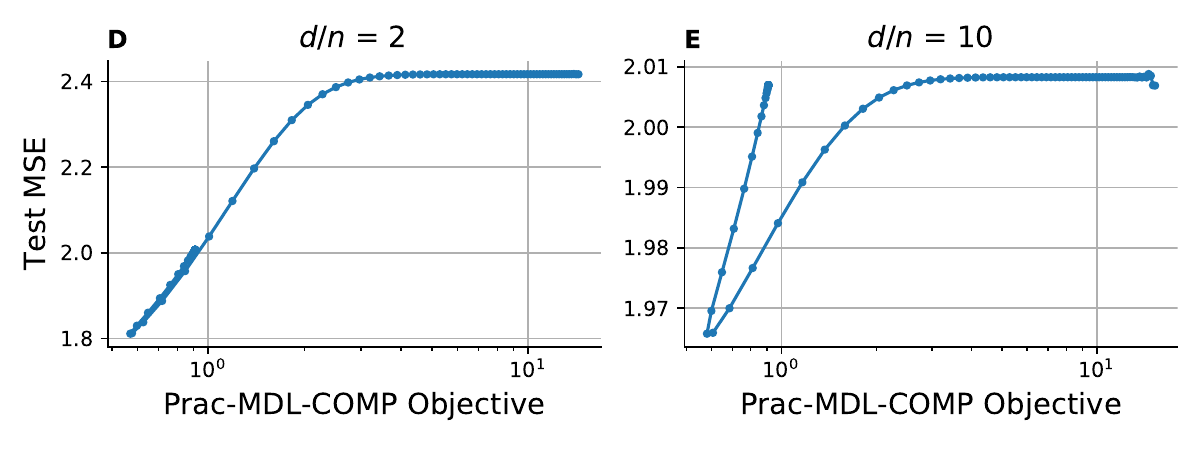}
              \includegraphics[height=1.32in]{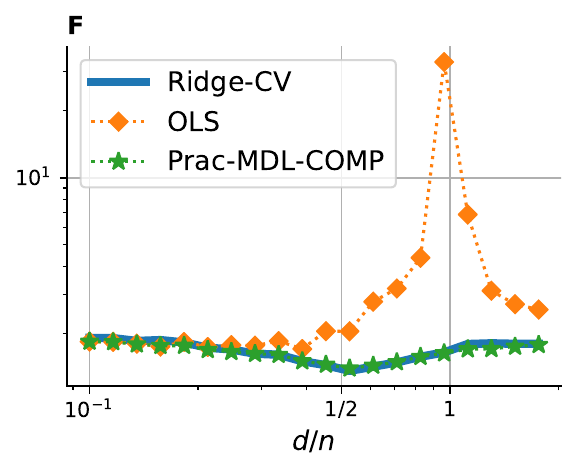} }
    \caption{Minimizing the objective~\eqref{eq:mdl_comp_objective}
      which defines \pracmdltag\ selects models with low test
      error. \textbf{A-E.}  Different panels show that this holds even
      as $d/n$ is varied. \textbf{F}.  \pracmdltag\ selects a model
      with Test MSE very close to Ridge cross-validation, avoiding the
      peak exhibited by OLS.}
    \label{fig:vary_lambda}
\end{figure}

\cref{fig:vary_lambda} shows that the model corresponding to the optimal
$\lambda$ achieving the minimum in \cref{eq:mdl_comp_objective}, has a low
test MSE (panels A-E), and comparable to the one obtained by leave-one-out
cross validation (LOOCV, panel F).
The setup follows the Gaussian model as in \cref{eq:linear_model_revisit} with the noise variance $\noisevar$ set to 1, with sample size $n=100$ fixed.
Here the covariates are drawn i.i.d. from $\mc N(0, 1)$, and then fixed. The true parameter $\tstar$ is set to be in dimension $50$ (extended to larger dimensions by appending zeros); its entries are first drawn i.i.d. from $\mc N(0, 1)$ and then scaled so that $\Vert\tstar\Vert=1$. We tune the parameter $\lambda$ over 20 values equally spaced on a log-scale from $10^{-3}$ to $10^6$. 

We vary the number of covariates ($d$) used for fitting the model and report the results for $d/n \in \braces{1/10, 1/2, 1, 2, 10}$ (noting that we have a misspecified model when fitting with $d<50$ features). Across all panels (A-E), we observe that the minima of the test MSE and the objective~\eqref{eq:mdl_comp_objective} for defining \pracmdltag\  often occur close to each other (points towards the bottom left of these panels).
\cref{fig:vary_lambda}F shows the generalization performance of the models
selected by \pracmdltag\ in the same setting. Selection via \pracmdltag\
generalizes well, very close to the best ridge estimator selected by leave-one-out
cross-validation. While the tuned ridge estimators (via CV, or MDL-COMP) exhibit the usual U-shaped curve for the test error in \cref{fig:vary_lambda}F, the OLS estimator exhibits a peak, a phenomenon termed as double-descent (that has been seriously investigated in recent works; see \cref{sec:discussion} for further discussion)
.

\cref{sec:supp_experiments} shows more results suggesting that \pracmdltag\ can select models well even under different forms of misspecification.

\subsection{Experiments with PMLB datasets}
\label{sub:real_experiments}

In this section, we report results on the behavior
\pracmdltag~\eqref{eq:mdl_comp_objective} when used to perform model
selection on real datasets.  Datasets are taken from PMLB
\citep{olson2017pmlb, OpenML2013}, a repository of diverse tabular
datasets for benchmarking machine-learning algorithms. We omit
datasets that are simply transformations of one another, or that have
too few features; doing so yields a total of 19 datasets spanning a
variety of tasks, such as predicting breast cancer from image
features, predicting automobile prices, and election results from
previous elections~\citep{simonoff2013analyzing}.  The mean number of
data points for these datasets is 122,259 and the mean number of
features is 19.1. For a given dataset, we fix $d$ to be the number of
features, and we vary $n$ downwards from its maximum value (by
subsampling the dataset) to construct instances with different values
of the ratio $d/n$.  The hyperparameter $\lambda$ takes on 10 values
equally spaced on a log scale between $10^{-3}$ and $10^3$.  The test
set consists of 25\% of the entire dataset.

\begin{figure}[ht]
    \centering
    \makebox[\textwidth][c]{\includegraphics[width=1\textwidth]{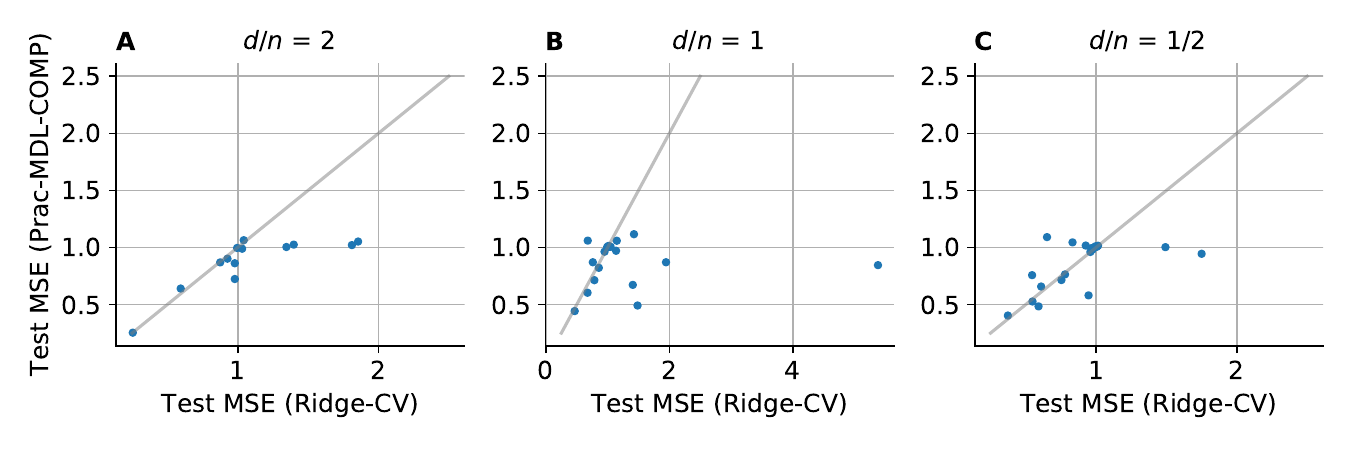}}
    \caption{Comparing test-error when selecting models using Prac-MDL-COMP
    versus using cross-validation on real datasets. \textbf{A}. When data
    is most limited, Prac-MDL-COMP based hyperparameter tuning outperforms Ridge-CV. \textbf{B, C}. As the amount of training samples increases, Prac-MDL-COMP performs comparably to Ridge-CV. Each point averaged over 3 random bootstrap samples.}
    \label{fig:pmlb}
\end{figure}

\cref{fig:pmlb}A compares the performance of Prac-MDL-COMP to
Ridge-CV.  We find that shows that in the limited data regime,
i.e. when $d/n$ is large, Prac-MDL-COMP tends to outperform.  As the
number of training samples is increased (i.e., $d/n$ decreases), the
advantage provided by selection via Prac-MDL-COMP decreases.  Further
details, and experiments on omitted datasets are provided in
\cref{sub:real_data_experiments_continued}; in particular, see
\cref{fig:pmlb_5fold,fig:pmlb_full,tab:datasets}.

\subsection{Experiments with fMRI data}
\label{sub:fmri}
We now focus on a challenging type of data that arises in
neuroscience. The dataset consists of neural responses of human
subjects, as recorded by functional magnetic-resonance imaging (fMRI),
as they are shown natural
movies~\citep{nishimoto2011reconstructing}. The training data consists
of 7,200 time points and the test data consists of 540 time points,
where at each timepoint a subject is watching a video clip.  The test
data is averaged over 10 repetitions of showing the same clip to the
same subject.  Following the previous work, we extract video features
using a Gabor transform, resulting in 1,280 features per
timepoint. From these features, we predict the response for each voxel
in the brain using ridge regression. To summarize, for this setting,
we have $d=1280, n_{\textrm{train}}=7200$ and $n_{\textrm{test}}=540$.
We restrict our analysis to 50 voxels with no missing data in the V1,
V2, and V4 regions of the brain, which are known to be easier to
predict. Before fitting, the features and responses are each
normalized to have mean zero and variance one.  In all fMRI
experiments, $\lambda$ takes on 40 values equally spaced on a log
scale between $10^{0}$ and $10^6$.

\begin{figure}[ht]
    \centering
    \includegraphics[width=0.9\textwidth]{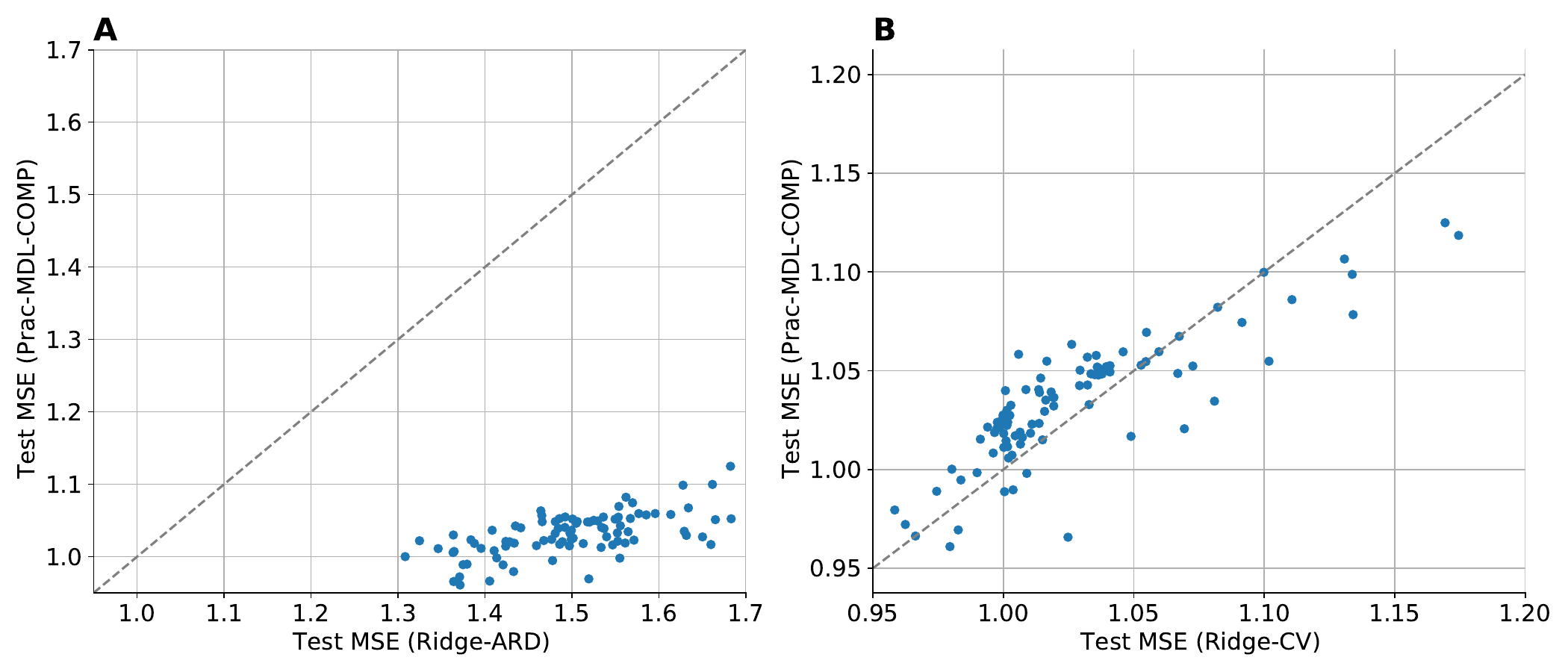}
    \caption{Prac-MDL-COMP succesfully selects models which predict fMRI responses well. \textbf{A}. Prac-MDL-COMP outperforms the Bayesian ARD baseline for every voxel (each point represents one voxel).
    Prac-MDL-COMP similarly outperforms a baseline using the BIC criterion (see \cref{fig:bic_fmri}).
    \textbf{B}. Prac-MDL-COMP is on par with cross-validation.
    }
    \label{fig:fmri_results}
\end{figure}

\cref{fig:fmri_results} shows our prediction results when using Prac-MDL-COMP for model selection to predict the fMRI responses. We also compare our approach to Automatic Relevance Determination (ARD),\footnote{Our choice to compare against ARD is also governed by the fact that ARD places a centered elliptic Gaussian distribution of the weights $w$; this means each coefficient $w_i$ can be drawn from a Gaussian distribution centered on zero with a unique covariance matrix. This leads to sparser coefficients $w$, a natural setting for the fMRI prediction problem we study.} also known as Sparse Bayesian Learning, a popular Bayesian approach which places an adaptive prior over the regression parameters~\citep{mackay1994bayesian} and BIC.\footnote{\label{fn:bic}In contrast to \pracmdltag's objective~\eqref{eq:mdl_comp_objective}, the BIC objective for tuning $\lambda$ is $ \min_{\lam} \frac{1}{n} \parenth{\frac{\enorm{\xmat
       \that_{\lam}-\vy}^2}{2 \noisevar} + \frac{\log n}{2} \sum_{i=1}^{\min\braces{n,d}} \frac{\evali}{\evali+\lam}}$.}
\cref{fig:fmri_results}A and \cref{fig:bic_fmri} respectively show that Prac-MDL-COMP consistently outperforms the Bayesian ARD baseline as well as BIC across voxels. Moreover, Prac-MDL-COMP is roughly on par with leave-one-out cross-validation (CV) for model selection in this data (\cref{fig:fmri_results}).  CV outperforms Prac-MDL-COMP for a majority of the voxels by a slight margin, but on the remaining voxels, Prac-MDL-COMP tends to outperform CV by a substantial margin. 

\begin{figure}[t!]
    \centering
    \includegraphics[width=0.9\textwidth]{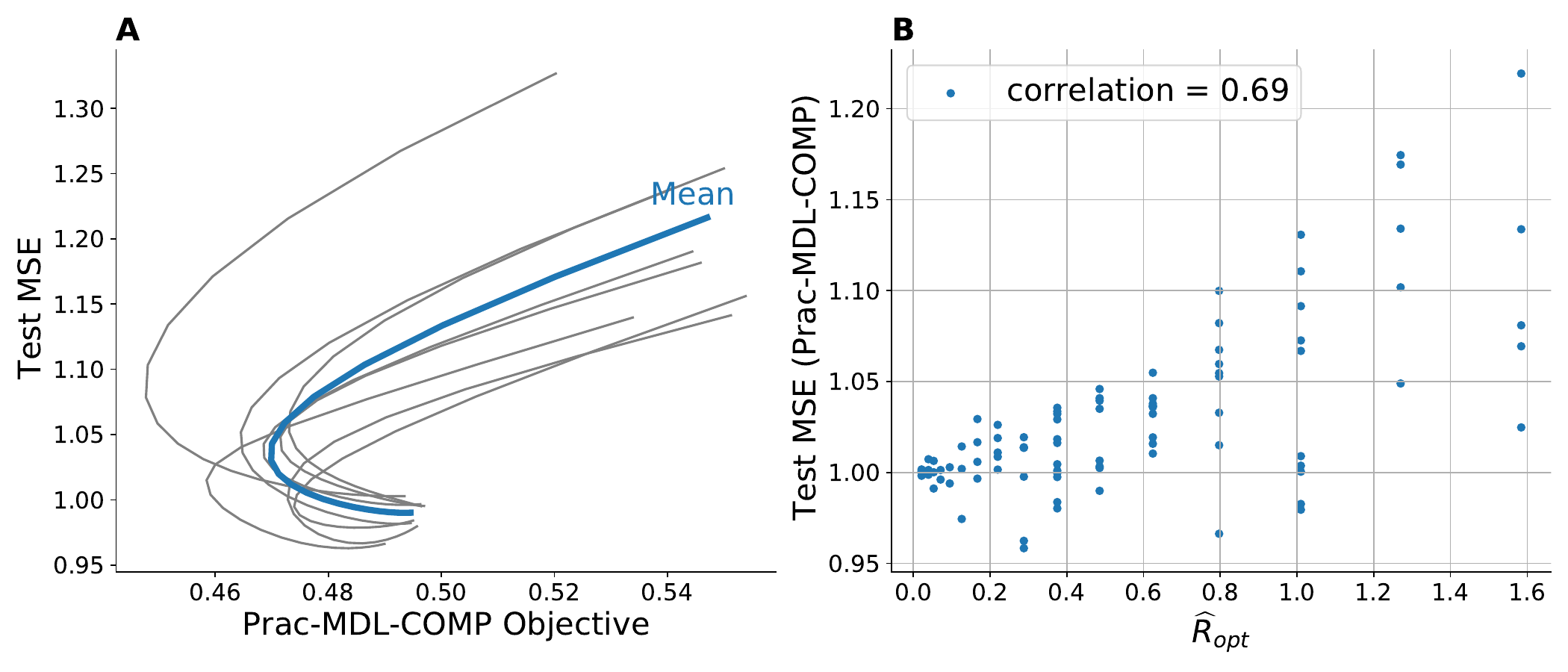}
    \caption{\textbf{A}. Test MSE often increases when Prac-MDL-COMP objective increases, showing that minimizing the objective informs good model selection for minimizing Test MSE. Each gray curve represents one randomly chosen voxel (only 10 out of 50 are shown for visualization). \textbf{B} Scatter plot (over the 50 different voxels) of test MSE versus the estimated $\widehat{\mc R}_{\textrm{opt}}$.}
    \label{fig:fmri_results_scatter_plot}
\end{figure}

\cref{fig:fmri_results_scatter_plot}A shows the relationship between the Prac-MDL-COMP objective and the test error. Test error tends to increase as the Prac-MDL-COMP objective increases, showing that minimizing the objective continues to inform good model selection for minimizing test error even for this challenging real dataset. In \cref{fig:fmri_results_scatter_plot}B, we provide a scatter plot of the test MSE versus $\widehat{\mc R}_{\textrm{opt}}$ computed from data. We notice a linear relationship between the two quantities, and observe a correlation of $0.69$. While, \cref{thm:lam_opt} guaranteed that the optimal redundancy $\ropt$ bounds the in-sample MSE, \cref{fig:fmri_results_scatter_plot}B suggests that its data-driven approximation also provides useful information about the (ordering of) test MSE.

\subsection{Experiments with neural tangent kernel on fMRI data}
\label{sub:fmri_ntk}
\begin{figure}[t]
    \centering
    \includegraphics[width=\textwidth]{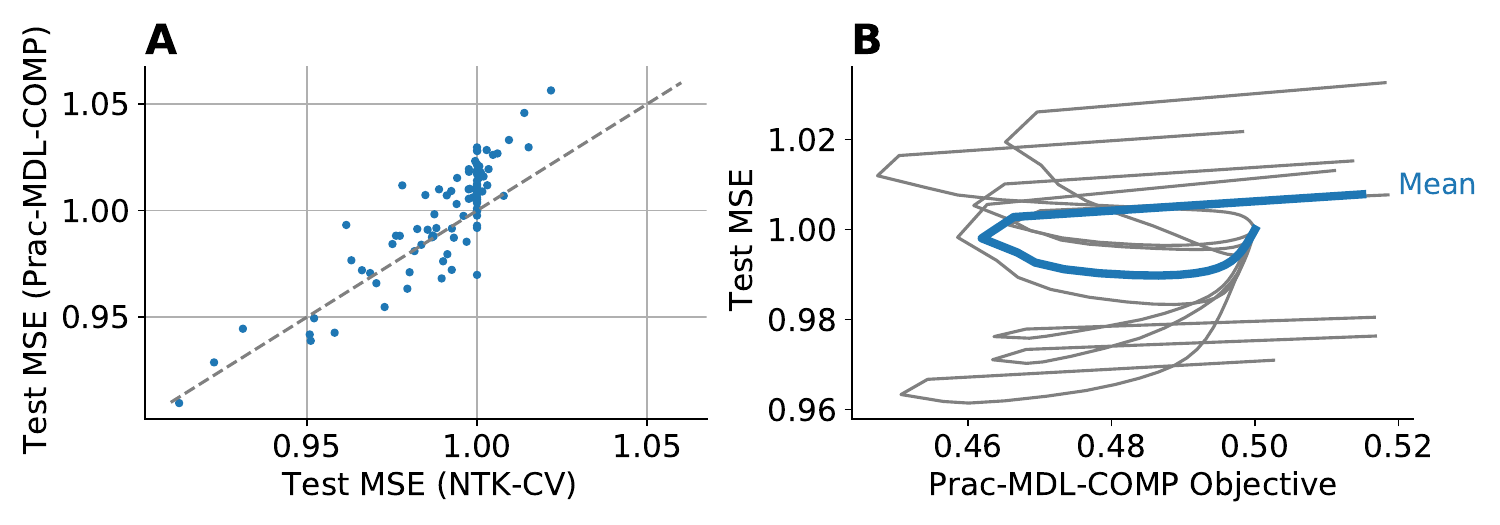}
    \caption{Prac-MDL-COMP succesfully selects models which predict
      fMRI responses well when using
      NTK-kernel. \textbf{A}. Prac-MDL-COMP is on par with
      cross-validation for every voxel (each point represents one
      voxel). \textbf{B}. Test MSE often increases when Prac-MDL-COMP
      objective increases, showing that minimizing the objective
      informs good model selection for minimizing Test MSE. Each gray
      curve represents one randomly chosen voxel (only 8 are shown for
      visualization).}
    \label{fig:fmri_ntk_results}
\end{figure}

\cref{fig:fmri_ntk_results} shows the results when repeating the
experiments in \cref{fig:fmri_results}, but now using kernel ridge
regression with the neural tangent kernel~\citep{jacot2018neural}
rather than linear ridge regression. It shows the prediction results
when using Prac-MDL-COMP (applied to kernel ridge regression, see
\cref{eq:kernel_prac_mdl_comp}) for model selection to predict the
fMRI responses. For the neural tangent kernel computation, we use the
neural-tangents library~\citep{neuraltangents2020} with its default
parameters (ReLU nonlinearity, two hidden linear layers with hidden
size of 512). \cref{fig:fmri_ntk_results}A shows that Prac-MDL-COMP is
roughly on par with leave-one-out cross-validation for model selection
in this data. \cref{fig:fmri_ntk_results}B shows the relationship
between the Prac-MDL-COMP objective and the test error, where we see
that the curves are flatter when compared to
\cref{fig:fmri_results_scatter_plot}A. Nonetheless, the test error
often decreases as the Prac-MDL-COMP objective increases, suggesting
that minimizing the objective is a good proxy for model selection for
minimizing test error even for this setting.

\section{Discussion}
\label{sec:discussion}
In this work, we revisited complexity
measures in the context of overparameterized models. We argued that there is a lack of theoretical justification for using parameter count as a complexity measure in overparameterized settings. We defined an MDL-based complexity measure MDL-COMP for linear and kernel methods, using codes induced by ridge estimators that can also deal with overparameterized settings. Our analytical results show that MDL-COMP depends on dimension $d$, sample size $n$, the covariate/kernel matrix, the true parameter/function properties, and the noise in the data. It does not grow linearly in $d$
for over-parameterized linear models, in fact, it often grows much more
slowly as $\log d$ for $d > n$. We prove that MDL-COMP informs the fixed design generalization error and provides good empirical results for the random design generalization error.  Numerical experiments show that a practical hyperparameter tuning scheme, inspired by MDL-COMP, provides generalization performance for various types of ridge estimators, often (although not always) outperforming cross-validation (CV) when the number of observations is limited.
Moreover, MDL-COMP based tuning offers computational savings over $K$-fold cross validation as it tunes the parameter using only a single-fold computation, thereby serving as a competitive alternative to CV in limited data regimes.

\subsection{Consequences for bias-variance tradeoff in overparameterized models}
Another consequence of our results is a better understanding of the recent mysteries around the bias-variance tradeoff principle with overparameterized models which we now elaborate. While the classical bias-variance tradeoff is known to exhibit a U-shaped curve in well-posed regimes, several recent works have exhibited a \emph{double-descent} curve that looks like the peaky curve of the OLS estimator from \cref{fig:vary_lambda}F (also see \cref{fig:test_mse_diff_design}(a)) in ill-posed regimes. (Notably we do not observe the double descent with good regularized estimators.)
Such a double-descent behavior for test error---with parameter count like measures on the $x$-axis as a proxy for complexity---has been observed and investigated in a series of recent works for linear regression~\citep{hastie2019surprises,belkin2019two,muthukumar2020harmless} as well as DNNs for classifications \citep{advani2017high,Nakkiran2019DeepDD} among other models~\citep{belkin2019reconciling}.\footnote{It is worth noting that this double-descent phenomenon was
documented as instability in early work on linear discriminant
analysis~\citep{skurichina1998bagging,skurichina2001bagging,skurichina2002bagging}. See \cite{loog2020brief} for further discussion on the history of double descent.
These non-classical test error curves have prompted several researchers to question the validity of the bias-variance tradeoff
principle, especially in overparameterized regimes~\citep{belkin2019reconciling, belkin2019two}. We highlight that double descent can also occur in non-overparameterized regimes when the covariate matrix is ill-conditioned (see \cref{fig:test_mse_diff_design}(c), and the discussion in this section).}

\paragraph{Investigating two possible causes of double descent:}
Our investigation into the complexity measures was partially motivated to better understand the double descent phenomenon. First, we note that the classical bias-variance tradeoff generally applies to \emph{a fixed training/test dataset in well-posed regimes, with a class of good estimators that are ordered by a suitable notion of complexity}. Thus, a non-classical test error curve, in principle, can arise for two reasons: (i) The choice of complexity measure is not suitable or well-justified, e.g., the parameter count in overparameterized regimes. (ii) The choice of estimators is not suitable due to ill-posedness of the data/model, e.g., OLS estimators with ill-conditioned covariate matrix, or in overparameterized regimes. 
The concurrent work on double descent uses parameter count as the complexity measure while plotting the test error of a range of ill-posed and well-posed estimators or models\footnote{Applied papers use fixed test set, but theory papers use varying training and test sets. Refer to the discussion on related work in the sequel.} (the shape of the OLS curve in \cref{fig:vary_lambda}F is representative of the double descent figures in the recent related work), and thus apriori it is not clear if the double descent phenomenon is a consequence of (i) or (ii). 

As a check for (i) (complexity), we can replace parameter count with MDL-COMP expressions (\cref{thm:complexity_expressions}, \cref{fig:mdl_gaussian}) on the $x$-axis of these test error plots (as it is valid even for overparameterized models). However, since MDL-COMP  remains monotone with dimension $d$, the qualitative conclusion about the double descent of the test error does not change.
Next, to investigate (ii) (estimators), we note from \cref{fig:vary_lambda}F that the tuned ridge estimators, either using cross-validation, or \pracmdltag, do exhibit the classical U-shaped curve with either choice of complexity measures, and obtain minimum test error at the true dimensionality of the data generating model. Furthermore, this conclusion is robust across ill-conditioned designs: \cref{fig:test_mse_diff_design} shows that tuned ridge estimators exhibit U-shaped test error curves even when the covariate matrix is ill-conditioned, and achieve superior test error than OLS. In fact, for our choice of covariate matrices, the OLS estimator exhibits double descent and  \emph{even multiple descent} in both low and high-dimensional settings, including non-overparameterized regimes.\footnote{Here low and high-dimensional setting respectively denotes whether or not the true generative model has more parameters than the training sample size.} Moreover, when using the best prediction error to do model selection (across the choice of number of features), the OLS estimator admits worse (sometimes significant worse) prediction error than its Ridge counterpart.

\paragraph{Related work on double descent:}
Recently, there has been a lot of research interest in probing the double descent phenomenon via different lenses.
On the one hand, several works establish sufficient conditions for the OLS estimator to achieve good generalization in an overparameterized setting ($d\gg n$), a phenomenon referred to as the \emph{benign overfitting}~\citep{bartlett2020benign,muthukumar2020harmless,hastie2019surprises,tsigler2020benign}. It is worth noting that the OLS estimator can continue to exhibit double descent even for these settings.
On the other hand, many works investigate the test error curves of ridge estimators for various settings: (a) When the dimensionality of the generative model $d$ varies along with the fitted model (keeping the fitted model correctly specified), \cite{hastie2019surprises} prove that optimally tuned ridge estimators exhibit a U-shape curve for linear models and isotropic features. In their set-up, one plots the test error curve for range of $d$ keeping the sample size fixed. 
(b) \cite{nakkiran2020optimal} make a similar conclusion for linear models with isotropic features but now for a fixed $d$ while varying the sample size $n$. We highlight that both (a) and (b) are subtly different than the practical set-up discussed in the previous paragraph where the training observations are fixed and only the fitted models vary. In contrast, both (a) and (b) vary the training observations simultaneously with the fitted estimator (and the fitted model always remains correctly specified). However, the qualitative conclusions drawn remain similar---tuned regularization restores the classical U-shape of test error curves.

\paragraph{Conclusions about the possible cause of double descent:}
Combining our findings along with the concurrent work, we can hypothesize that while parameter count is not a justified complexity measure, the double descent phenomenon is likely to be a consequence of a poor choice of estimators in ill-posed regimes. Poor estimators like OLS can exhibit double descent or even multiple descent depending on the covariate matrix, while regularized estimators with tuned hyper-parameters continue to exhibit U-shaped test error curves.

\subsection{Future directions}
We believe that our work takes a useful step towards questioning the fundamental components that underlie the principle of bias-variance tradeoff, and provides a proof of concept for the value of revisiting the complexity measure in overparameterized settings. 
Besides, several direct future directions arise from our work. Relating MDL-COMP with
out-of-sample guarantees under additional assumptions on the covariate
design, like those in~\citep{hsu2012random,dobriban2018high} for the linear model is an interesting open problem. Our results for kernel methods are in the classical low-dimensional regime, and it is known that several kernels in the high-dimensional regime behave very similar to the linear kernel~\citep{el2010information,el2010spectrum}. Understanding the consequences in such high-dimensional regime for kernel methods with the MDL lens is another interesting future direction. Next, we note that our measures are based on ridge estimators but they are often
not suitable for parameter estimation with sparse models in
high-dimensions, and thus deriving MDL-COMP with codes suitably
adapted for $\ell_1$-regularization would also be
interesting. Additionally, it remains to investigate suitable variants of our \mdltag\ or the more general one-part universal coding~\citep{grunwald2019minimum,grunwald2017tight} beyond linear and kernel methods, e.g., for deep neural
networks, as well as for classification tasks.

\acks{This work was partially supported by
National Science Foundation
  grant NSF-DMS-1613002, NSF-2015341, and the Center for Science of
  Information (CSoI), a US NSF Science and Technology Center, under
  grant agreement CCF-0939370, the NSF grant 2023505 on Collaborative Research: Foundations of Data Science Institute (FODSI), the NSF and the Simons Foundation for the Collaboration on the Theoretical Foundations of Deep Learning through awards DMS-2031883 and 814639, and an Amazon Research Award to BY, and
  Office of Naval Research grant DOD ONR-N00014-18-1-2640 and National
  Science Foundation grant NSF-DMS-2015454 to MJW.}

\appendix
\tableofcontents
\setcounter{table}{0}
\setcounter{figure}{0}
\renewcommand{\thefigure}{A\arabic{figure}}
\renewcommand{\thetable}{A\arabic{table}}

\section{Further discussion of \mdltag }
\label{sub:non_isotropic}

We start with additional details on the simulation set-up and sketch
related to the \mdltag\ scaling from
\cref{sub:mdl_complexity_for_linear_models} in
\cref{sub:mdl_plots_details}.  In \cref{sub:sketch_gaussian}, we
sketch the proof for the \mdltag's behavior as observed in
\cref{fig:mdl_gaussian}, and then provide an analytical bound for
$\ropt$ in \cref{sub:ropt_large_scale} under high-dimensional
asymptotics.  In \cref{sub:a_minimax_optimality_via_mdltag}, we argue
that the minimax optimality of MDL-COMP holds under a broad class of
noise distributions (beyond Gaussian) under the linear
model~\eqref{eq:linear_model_revisit}.  Finally, we collect some
additional background related to MDL in \cref{sub:further_background}.

\subsection{Simulation set-up} 
\label{sub:mdl_plots_details}

Here we elaborate the set-up associated with
\cref{fig:mdl_gaussian,fig:mdl_spike_design}. The true dimensionality
$d_\star$ denotes that the observations $\vy = \widetilde{\xmat}\tstar
+ \noise[]$ depend only on first $d_\star$ covariates of the full
matrix $ \xmat_{\textrm{full}}$, Given a sample size $n$ (fixed for a
given dataset, fixed in a given plot), the matrix
$\xmat_{\textrm{full}} \in \real^{n \times \dmax}$, and
$\widetilde{\xmat} \in \real^{n\times d_\star}$ where $\dmax$ denotes
the maximum number of covariates available for fitting the model (and
in our plots can be computed by multiplying the sample size with the
the maximum value of $d/n$ denoted on the $x$-axis). When we vary $d$,
we use $\xmat \in \real^{n \times d}$ (selecting first $d$ columns of
$\xmat_{\textrm{full}}$) for fitting the ridge model and computing the
\mdltag.

\paragraph{Defining $\v{\myvec}$, and clarifying footnote~\ref{footnote:defining_w}:}
In order to compute \mdltag~\eqref{eq:mdl_linear_explicit}, we need to
compute the eigenvalues $\evali$ of $\xtx$, where $\xmat \in \real^{n
  \times d}$ denotes the first $d$ columns of the full matrix
$\xmat_{\textrm{full}}$ that are used for fitting the ridge model, and
computing the corresponding encoding and \mdltag.  Moreover, we need
to compute the vector $\v\myvec$ defined equal to $\mat U \tp \tstar$
in equation~\eqref{eq:xmatrix}.  Note that $\mat U$ has size $d \times
d$ and $\tstar$ is $d_\star$ dimensional. So when $d<d_\star$, the
vector $\v\myvec$ is computed by restricting $\tstar$ to first $d$
dimensions (in \cref{eq:xmatrix}), i.e., using $\widetilde{\theta}_{\star} =
(\tstar)_{[1:d]}$ (the orthogonal projection of $\tstar$ on $\xtx$)
and define $\v\myvec = \mat U \tp \widetilde{\tstar}$.  On the other
hand, for $d>d_\star$, we simply extend the $\tstar$ by appending
$d-d_\star$ $0$'s, i.e., $\widetilde{\theta}_{\star} = \begin{bmatrix} \tstar
  \\ \mathbf 0 _{d-d_\star}
\end{bmatrix}$ and then set $\v\myvec = \mat U \tp \widetilde{\theta}_{\star}$.

\subsection{Proof sketch for random isotropic
  designs}
\label{sub:sketch_gaussian}

In this section, we provide a proof sketch to explain the behavior of
\mdltag/ when applied to random isotropic designs, as plotted in
\cref{fig:mdl_gaussian}.  Suppose that the design matrix $\xmat \in
\real^{n \times d}$ has entries drawn from iid from the Gaussian
distribution $\mc N(0, 1/n)$; the $1/n$-variance serves to ensure that
$\xmat$ has columns with expected squared norm equal to one.  When $n
\gg d$, standard random matrix theory guarantees that $\xtx \approx
\mat I_d$ and hence $\evali \approx 1$.  For $d \gg n$, one can apply
the same argument to the matrix $\xmat\xmat\tp$ to conclude that
$\xmat \xmat\tp \approx \frac dn \mat I_n$; thus, the matrix $\xtx$
has rank $n$ with its non-zero eigenvalues roughly equal to $d/n$. And, from above, we recall the definition $\v\myvec = \mat U\tp \widetilde{\theta}_{\star}$, where $\widetilde{\theta}_{\star}$ is either a restriction of $\tstar$ when the model is under-specified, or is obtained by appending zeros when the model is over-specified.

Since the matrix $\mat U$ consists of the eigenvectors of $\xtx$, it
has a uniform distribution over the space of all orthogonal matrices
in dimension $d$. Let $r^2 \defn \enorm{\tstar}^2$ and, note from above that $\enorm{\v\myvec}^2 = \enorm{\widetilde{\theta}_{\star}}^2$. When $d \geq d_\star$, we have $\enorm{\widetilde{\theta}_{\star}}^2 = r^2$, and when $d<d_{\star}$, and the coordinates of $\tstar$ are drawn iid, we have $\enorm{\widetilde{\theta}_{\star}}^2 \approx \frac{d}{d_{\star}} r^2$. Given the distribution of $\mat U$, we have that the entries of $w_i^2$ are approximately equal, and thus conclude that $ w_i^2 \approx \min\braces{1, \frac{d}{d_{\star}}} \frac{r^2}{d} = r^2 \cdot
\min\braces{\frac{1}{d}, \frac{1}{d_\star}}$.  Plugging in these
approximations, when $d_\star <n$, we find that
\begin{align}
    \mdltag\!=\! \frac1{2n}\sumn[\minnd] \log\parenth{\evali+ \frac{\noisevar}
    {\myvec_i^2}} 
    \approx \begin{cases}
     \displaystyle \frac d{2n} \log\parenth{1+ d_\star/r^2}
    &\text{ if }d \in [1, d_\star]\\[2mm]
    \displaystyle \frac d{2n} \log\parenth{1+d/r^2}
     &\text{ if } d \in [d_\star, n]\\[2mm]
    \displaystyle \frac12\log \brackets{d \parenth{\frac 1n+\frac{1}{r^2}}}
    &\text{ if } d \in [n, \infty)
    \end{cases}
    \label{eq:adj_scaling} 
\end{align}
\begin{align}
    \ropt\!=\!\frac1{2n}\sumn[\minnd] \log\parenth{1 +  \frac{\evali\myvec_i^2}{\noisevar}}
    \approx \begin{cases}
    \displaystyle \frac d{2n} \log\parenth{1+ \frac{r^2}{d_\star}}
    &\text{ if } d \in [1, d_\star]\\[2mm]
    \displaystyle\frac d{2n} \log\parenth{1+ \frac{r^2}{d}}
    &\text{ if }d \in [d_\star, n]\\[2mm]
    \displaystyle \frac12\log\parenth{1+ \frac{r^2}{n}} 
    &\text{ if }  d \in [n, \infty).
    \end{cases}
    \label{eq:adj_scaling_ropt}
\end{align}
For the case when $d_\star > n$, i.e., the true dimensionality is larger than the sample size, we have
\begin{align}
    \mdltag\!=\! \frac1{2n}\sumn[\minnd] \log\parenth{\evali+ \frac{\noisevar}
    {\myvec_i^2}}
    &\approx \begin{cases}
    \displaystyle \frac d{2n} \log\parenth{1+ \frac{d_\star}{r^2}}
    &\text{ if } d \in [1, n]\\[2mm]
    \displaystyle\frac12\log\parenth{\frac d{n} + \frac{d_\star}{r^2}}
    &\text{ if } d \in [n, d_\star]\\[2mm]
    \displaystyle  \frac12\log \brackets{d \parenth{\frac 1n+\frac{1}{r^2}}}
    &\text{ if } d \in [d_\star, \infty),
    \end{cases}
    \label{eq:adj_scaling_high_dim} \\
    \ropt= \frac1{2n}\sumn[\minnd] \log\parenth{1 +  \frac{\evali\myvec_i^2}{\noisevar}}
    &\approx \begin{cases}
    \displaystyle \frac d{2n} \log\parenth{1+ \frac{r^2}{d_\star}}
    &\text{ if } d \in [1, d_\star]\\[2mm]
    \displaystyle\frac d{2n} \log\parenth{1+ \frac{r^2}{d}}
    &\text{ if } d \in [d_\star, n]\\[2mm]
    \displaystyle \frac12\log\parenth{1+ \frac{r^2}{n}} 
    &\text{ if }  d \in [n, \infty).
    \end{cases}
    \label{eq:adj_scaling_ropt_high_dim}
\end{align}
In both of the cases covered by equations~\eqref{eq:adj_scaling} and
\eqref{eq:adj_scaling_high_dim}, we find that the MDL-COMP has scaling
$\mc O(d/n)$ for small $d$, and $\mc O (\log d)$ for $d \gg n$.
Overall, this argument along with results in \cref{fig:mdl_gaussian}
suggest that $d/n$ should not be treated as the default complexity for
$d > n$, and that \mdltag\ provides a scaling of order $\log d$ for $d
> n$.

\subsection{Expressions for $\ropt$ for large scale settings}
\label{sub:ropt_large_scale}

In recent work, a number of
authors~\citep{hastie2019surprises,belkin2019two} have studied the
generalization error of different estimators applied to linear models
under the high-dimensional asymptotic scaling $\dims, \obs \to \infty$
and $\dims/\obs \to \gamma$.  In a similar setting, we can derive an
expression for the optimal redundancy $\ropt$. 
\begin{theorem}
\label{thm:rmt_expressions}
Suppose that the covariate matrix $\mat X \in \real^{n \times d}$ has
i.i.d. entries drawn from a distribution with mean $0$, variance $1/n$
and fourth moments of order $1/n^2$, and that the parameter $\tstar$
is drawn randomly from a rotationally invariant distribution, with
$\Exs[\frac{\enorm{\tstar}^2}{\noisevar d}] = \snr$.  Then, we have
  \begin{align}
    \label{eq:mdl_complexity_rmt}
    \lim_{n, d\to \infty, \frac{d}{n}\to\gamma}\Exs_{\tstar}[\ropt(\ptrue, \qsetlin)]
    \leq \gamma \log(1+\snr - \delta)
    + \log(1+\gamma \cdot \snr - \delta)
    - \frac{\delta}{\snr},
  \end{align}
  almost surely, where $\delta := (\sqrt{\snr(1+\sqrt{\gamma})^2+1} - \sqrt{\snr(1-\sqrt{\gamma})^2+1})^2/4$.
\end{theorem}
See \cref{sub:proof_of_thm:rmt_expressions} for the proof.  We remark
that the inequality in the theorem is a consequence of Jensen's
inequality and can be removed whenever a good control over the
quantity $\Exs_{w_i^2}\log(1+w_i^2 \evali)$ is available, where $\evali$ denoting the eigenvalues of $\xtx$.
The first term on the RHS of \cref{eq:mdl_complexity_rmt} has a
scaling of order $d/n$ when the norm of $\tstar$ grows with $d$.
However, when $\enorm{\tstar}$ is held constant with $d$, the optimal
redundacy $\ropt$ does not grow and remains bounded by a constant. The
bound~\eqref{eq:mdl_complexity_rmt} has a similar scaling as that
noted earlier in
\cref{eq:adj_scaling_ropt,eq:adj_scaling_ropt_high_dim} from our proof
sketch (for $d \in [d_\star, n]$ or $d\in [n , \infty)$, assuming
  $d_\star$ fixed, $r^2$ growing with $d$).

\subsection{\mdltag\ over a wider range of noise distributions} 
\label{sub:a_minimax_optimality_via_mdltag}

\cref{thm:complexity_expressions} provides an explicit expression for
\mdltag\ assuming that the noise follows a Gaussian distribution.  In
this section, we show that the optimal code from \cref{eq:ropt_linear}
defining the \mdltag\ procedure~\eqref{eq:mdl_linear} also achieves a
minimax codelength over a wider range of noise distributions. Let
$\mc{P}$ denote the set of all distributions on $\real^n$, and define
the family
\begin{align}
  \label{eq:plin_defn}
  \plin = \braces{\P \in \mc{P}: \Exs[\noise[]] = 0,
    \textrm{Var}(\noise[])\preceq \noisevar \mat I_n }.
\end{align}
For the generative linear model~\eqref{eq:linear_model_revisit} with
noise distribution now allowed to belong to the family
$\plin$~\eqref{eq:plin_defn}, we have
\begin{align*}
    \Exs_{\noise[]}[\vy\vert \xmat] = \xmat \tstar, \qtext{and}
    \textrm{Var}_{\noise[]}(\vy \vert \xmat) \preceq \noisevar \mat
    I_n.
\end{align*}
Our next result shows that the code definining \mdltag\ also achieves
the minimax codelength over the noise distributions in $\plin$.
\begin{theorem}
\label{thm:minimax_codelength}
  The distribution $\qdist_{\lammatopt}$ that achieves the \mdltag\ in
  \cref{eq:mdl_linear} also achieves the minimax codelength in the
  class $\plin$, i.e.,
  \begin{align}
    \label{eq:minimax_codelength}
    \qdist_{\lammatopt} \in\arg\min_{\qdist\in\qsetlin}
    \max_{\P\in\plin}
    \Exs_{\noise[]\sim\P}\log\parenth{\frac{1}{\qden(\vy)}},
  \end{align}
  where $\qden$ denotes the density of the distribution $\qdist$.
\end{theorem}
\noindent See \cref{sub:proof_of_thm:minimax_codelength} for the proof
of this claim. \\

\subsection{MDL-COMP with multiplicative factor $1$ in \cref{eq:lamopt_claim}}
\label{sub:mult_factor_1}
Note that we change the multiplicative factor of the codelength for $\lammat$ in  \cref{eq:lamopt_claim} from $\frac12$ to $1$, and thereby the definition of \mdltag\ in \cref{eq:lamopt_claim} the expression~\eqref{eq:mdl_linear_explicit} for MDL-COMP in \cref{thm:complexity_expressions} becomes
\begin{align*}
  \mdltag(\ptrue, \qsetlin) & = \frac{1}{2n}\sumn[\minnd]
        \log\parenth{ \frac{\evali\noisevar}{\myvec_i^2} + \frac{\sigma^4}{\myvec_i^4}},
\end{align*}
where we use the following results from the proof of \cref{thm:complexity_expressions}:
\begin{align*}
\lambda_i^{\textrm{opt}} = \frac{\noisevar}{\myvec_i^2} \quad
\mbox{and} \quad \ropt = \frac{1}{2n} \sumn[\minnd] \log\parenth{1 +
  \frac{\evali}{\lambda_i^{\textrm{opt}}}},
\end{align*}
which then yields the following expressions in place of \cref{eq:adj_scaling} for the approximate scaling of \mdltag\ for random isotropic designs:
\begin{align*}
    \mdltag\!=\! \frac1{2n}\sumn[\minnd] \log\parenth{ \frac{\evali\noisevar}{\myvec_i^2} + \frac{\sigma^4}{\myvec_i^4}}
    \!\approx\! \begin{cases}
     \displaystyle \frac d{2n} \brackets{  \log\parenth{1+ d_\star/r^2} + \log( d_\star/r^2)}
    &\text{ if }d \in [1, d_\star]\\[2mm]
    \displaystyle \frac d{2n}  \brackets{  \log\parenth{1+ d/r^2} + \log( d/r^2)}
     &\text{ if } d \in [d_\star, n]\\[2mm]
    \displaystyle  \frac12 \log \brackets{d \parenth{\frac 1n+\frac{1}{r^2}}}
    &\text{ if } d \in [n, \infty),
    \end{cases}
\end{align*}
and we can bound all of these. Analogous expressions in place of \cref{eq:adj_scaling_high_dim} can be derived similarly. In all cases, the new expressions can be bounded by at most $2$ times of the previous expressions, so that the scaling of \mdltag\ remains unaffected.

\subsection{Relation between NML and maximum likelihood principle}
\label{sub:further_background}

As noted earlier, for computing the optimal redundancy in
\cref{eq:ropt_defn}, one can replace the set of codes
$\qset_{\tagridge}$ by a generic set of codes $\qset$ corresponding to
many classes of models. In contrast, for maximum likelhood estimation
(MLE), one maximizes the log likelihood of the data over a fixed model
class. Maximum likelihood is generally used when a model class is
fixed, and is known to break down when considering even nested classes
of parametric models~\citep{hansen2001model}. On the other hand, the
definition~\eqref{eq:ropt_defn} can be suitably adjusted even for the
case when there is no true generative model for $\vy$. At least in
general, the quantity $\qdist(\vy)$ need not denote the likelihood of
the observation $\vy$, and the distribution $\qdist$ may not even
correspond to a generative model. In such cases, the optimal choice of
$\qdist$ in \cref{eq:ropt_defn} is supposed to be optimal not just for
$\vy$ from a parametric family, but also for $\vy$ from one of many
parametric model classes of different dimensions.

However, when $\qset$ is a single parametric family, i.e., $\qset =
\braces{\pdist_{\theta}, \theta \in \Theta}$ where $\theta$ denotes
the unknown parameter of interest, the MDL principle does reduce to
the MLE principle. In more precise terms, the MLE can be seen as the
continuum limit of two-part MDL (see Chapter
15.4~\citep{grunwald2007minimum}). In this case, the optimal NML code
$\qdist^\star(\vy)$ is given by the logarithm of the likelihood
computed at MLE plus a term that depends on the normalization constant
of the maximum likelihood over all possible observations; this fact
underlies the nomenclature of normalized maximum likelihood, or NML
for short.

For the low-dimensional linear models (fixed $d$ and $n\to \infty)$,
while several MDL-based complexities, namely two-part codelength,
stochastic information complexity, and NML complexity are equivalent
to the first $\mc O(\log n)$ term---which in turn scales like
$\tfrac{1}{2} d \log n$. Moreover, the NML complexity can be deemed
optimal when accounting for the constant term, i.e., NML complexity
achieves the optimal $\mc O(1)$ term which involves Jeffrey's
prior~\citep{barron1998minimum}, asymptotically for low-dimensional
linear models. We refer readers to the
sources~\citep{rissanen1986stochastic,barron1998minimum,hansen2001model}
for further discussion on two-part coding, stochastic information
complexity, and toSections 5.6, 15.4 of the
book~\citep{grunwald2007minimum} for further discussion on the
distinctions between MDL and MLE.

\section{Further numerical experiments}
\label{sec:supp_experiments}

\setcounter{table}{0} \setcounter{figure}{0}
\renewcommand{\thefigure}{B\arabic{figure}}
\renewcommand{\thetable}{B\arabic{table}}

We now present additional experiments showing the usefulness of the
data-driven \pracmdltag~\eqref{eq:mdl_comp_objective}.  In particular,
we show that \pracmdltag\ informs out-of-sample MSE in Gaussian as
well as several misspecified linear models
(\cref{sub:misspecified_linear_models}), and then provide further
insight on the performance of real-data experiments (deferred from
\cref{sub:real_experiments} of the main paper) in
\cref{sub:real_data_experiments_continued}.

\begin{figure}[t!]
    \centering
    \includegraphics[width=0.3\textwidth]{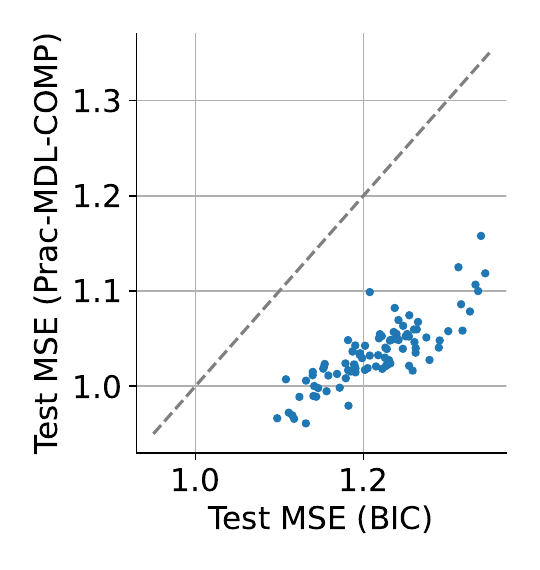}
    \caption{Prac-MDL-COMP successfully selects models which predict fMRI responses well compared to the BIC criterion (comparisons with cross-validation and Ridge-ARD shown in \cref{{fig:fmri_results}}).}
    \label{fig:bic_fmri}
\end{figure}

\begin{figure}[ht]
    \centering
    \makebox[\textwidth][c]{\includegraphics[width=\textwidth]{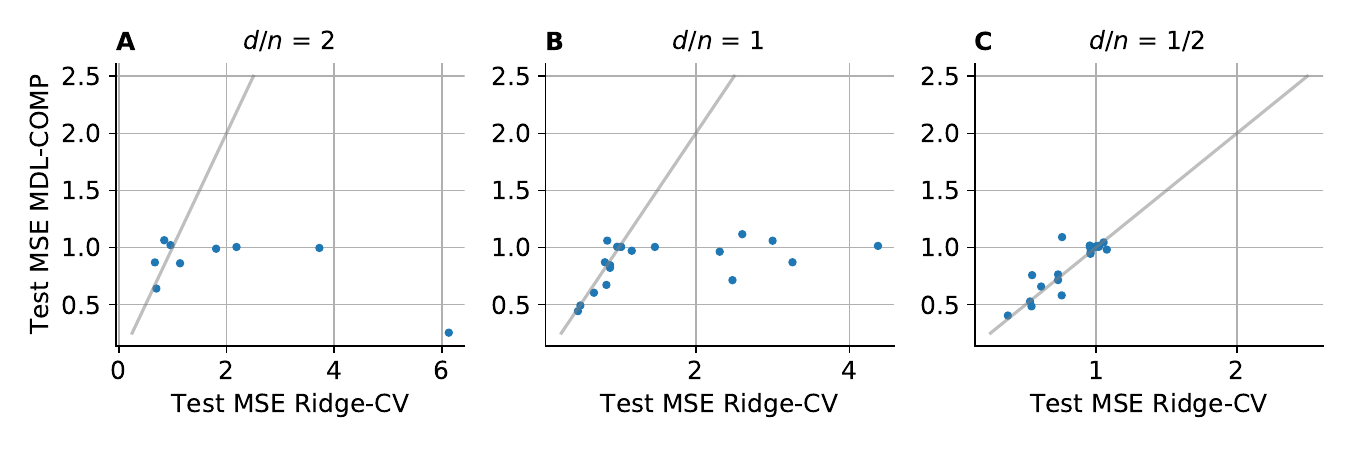}}
    \caption{Results for \fref{fig:pmlb} hold when using 5-fold CV rather than LOOCV.}
    \label{fig:pmlb_5fold}
\end{figure}

\subsection{Misspecified linear models} 
\label{sub:misspecified_linear_models}

\begin{figure}[ht]
    \centering \includegraphics[width=\textwidth]{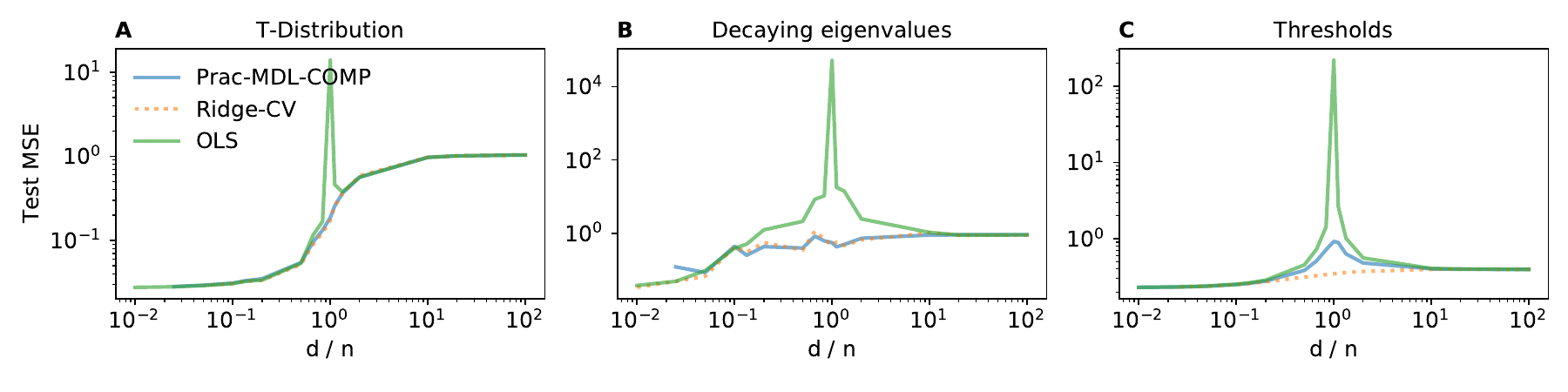}
    \caption{Model selection under misspecification. Under various
      misspecifications, \pracmdltag\ still manages to select models
      which generalize reasonably well. Different from other figures
      presented in the paper, here $d$ is fixed and the sample size
      $n$ is varied.}
    \label{fig:shifts}
\end{figure}

We specify three different types of model misspecification taken from
prior work \citep{yu2020veridical} and analyze the ability of MDL-COMP
to select models that generalize. \cref{fig:shifts} shows that under
these conditions, MDL-COMP still manages to pick a $\lambda$ which
generalizes fairly well. \textit{T-distribution} refers to errors
being distributed with a t-distribution with three degrees of freedom.
\textit{Decaying eigenvalues} refers to the eigenvalues of the
covariance matrix $\lambda_i$ decaying as $1/2^{i}$, inducing
structure in the covariance matrix.  \textit{Thresholds} refers to
calculating the outcome using indicator functions for $X>0$ in place
of $X$.  Here, Ridge-CV (orange dotted line) refers to model selection
using leave-one-out cross-validation and Prac-MDL-COMP (blue line)
refers to model selection for a ridge model based on optimizing
Prac-MDL-COMP.

\subsection{Real data experiments continued} %
\label{sub:real_data_experiments_continued}
We now provide further investigation.  First, the good performance of
\pracmdltag\ based linear model also holds ground against $5$-fold CV,
see \cref{fig:pmlb_5fold}.

\begin{figure}[ht!]
    \centering
    \makebox[\textwidth][c]{\includegraphics[width=1\textwidth]{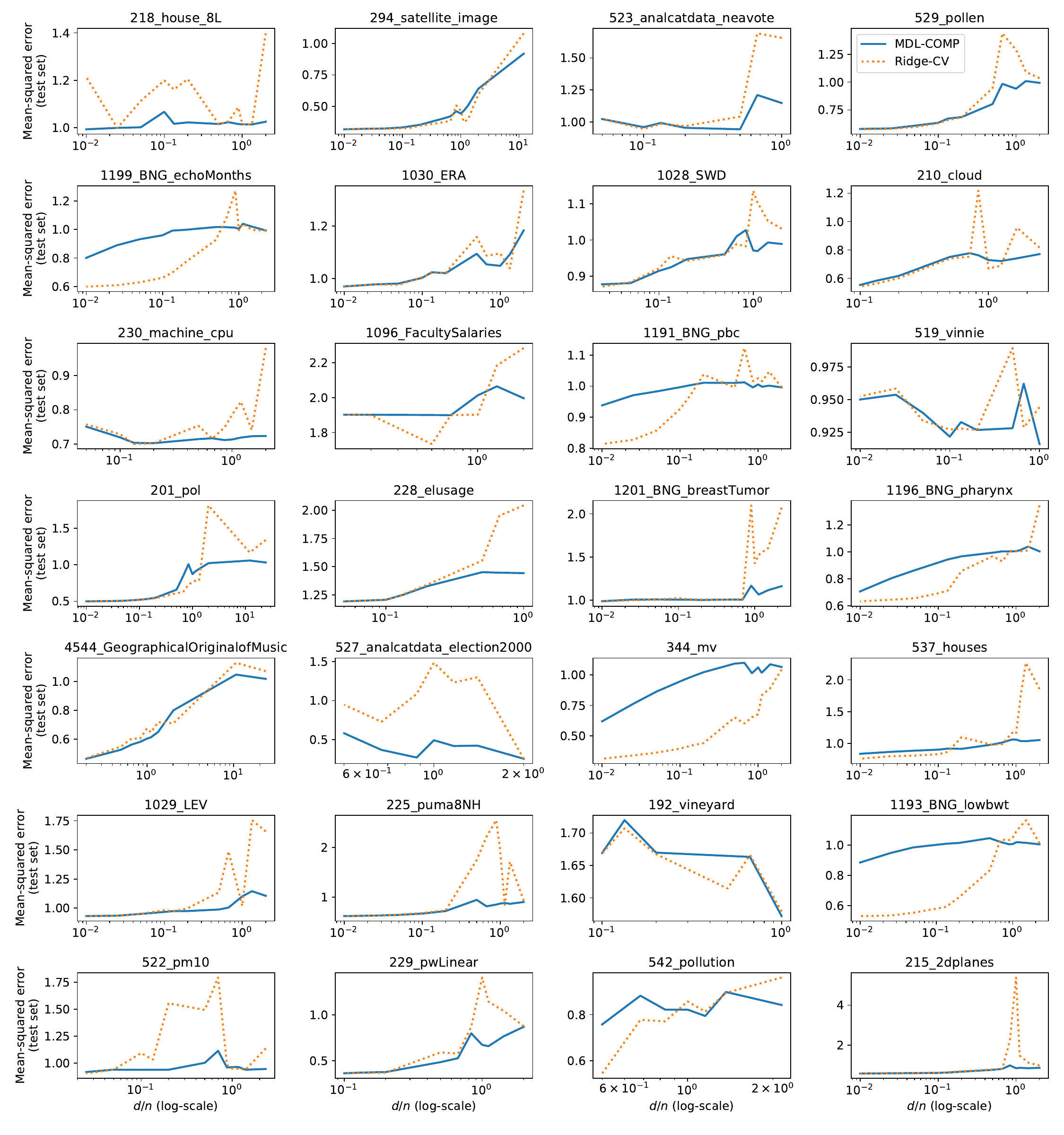}}
    \caption{Test MSE when varying the number of training points sampled from real datasets (lower is better). MDL-COMP performs well, particularly when the number of training points is small (right-hand side of each plot). Each point averaged over 3 random bootstrap samples.}
    \label{fig:pmlb_full}
\end{figure}

Here we show results for 28 real datasets in \cref{fig:pmlb_full}
where the plot titles correspond to dataset IDs in
OpenML~\citep{OpenML2013}.  In the limited data regime (when $d/n$ is
large, the right-hand side of each plot), MDL-COMP tends to outperform
Ridge-CV. As the number of training samples is increased, the gap
closes or cross-validation begins to outperform MDL-COMP.  These
observations provide evidence for promises of \pracmdltag\ based
regularization for limited data settings.

\begin{table}[ht!]
    \makebox[\textwidth][c]{
    \small
    \centering
    \begin{tabular}{lrrcc}
\toprule
         \shortstack{\bf Dataset name \\ \bf(OpenML ID)} &  \shortstack{\bf \#obs $n$} &  
         \shortstack{\bf \#feats  $d$ } &  \shortstack{\bf Ridge-CV\\ \bf  MSE} &  
         \shortstack{\bf Prac-MDL-COMP  \\ \bf MSE} \\
\midrule
                         1028\_SWD &                 1000 &                  11 &                                 1.17 &                                 \bf 0.97 \\
                         1029\_LEV &                 1000 &                   5 &                                      NaN* &                                 \bf 1.10 \\
                         1030\_ERA &                 1000 &                   5 &                                      NaN* &                                 \bf 1.05 \\
             1096\_FacultySalaries &                   50 &                   5 &                                      NaN* &                                 \bf 2.01 \\
                     1191\_BNG\_pbc &              1000000 &                  19 &                                 1.02 &                                 \bf 1.00 \\
                  1193\_BNG\_lowbwt &                31104 &                  10 &                                 \bf 0.98 &                                 1.01 \\
                 1196\_BNG\_pharynx &              1000000 &                  11 &                                 1.03 &                                 \bf 1.00 \\
              1199\_BNG\_echoMonths &                17496 &                  10 &                                 1.47 &                                 \bf 1.00 \\
             1201\_BNG\_breastTumor &               116640 &                  10 &                                 2.61 &                                 \bf 1.12 \\
                     192\_vineyard &                   52 &                   3 &                                      NaN* &                             \bf    1.57 \\
                          201\_pol &                15000 &                  49 &                                 \bf 0.82 &                                 0.87 \\
                        210\_cloud &                  108 &                   6 &                                 3.41 &                                 \bf 0.73 \\
                     215\_2dplanes &                40768 &                  11 &                                 0.88 &                                 \bf 0.85 \\
                     218\_house\_8L &                22784 &                   9 &                                 4.37 &                                 \bf 1.01 \\
                      225\_puma8NH &                 8192 &                   9 &                                 3.26 &                                \bf 0.87 \\
                      228\_elusage &                   55 &                   3 &                                      NaN* &                                 \bf 1.44 \\
                     229\_pwLinear &                  200 &                  11 &                                 0.83 &                                 \bf 0.67 \\
                  230\_machine\_cpu &                  209 &                   7 &                                 2.48 &                                 \bf 0.71 \\
              294\_satellite\_image &                 6435 &                  37 &                                 0.47 &                                 \bf 0.44 \\
                           344\_mv &                40768 &                  11 &                                 \bf 0.85 &                                 1.06 \\
 4544\_GeographicalOriginalofMusic &                 1059 &                 118 &                                 0.67 &                                 \bf 0.60 \\
                       519\_vinnie &                  380 &                   3 &                                      NaN* &                                 \bf 0.92 \\
                         522\_pm10 &                  500 &                   8 &                                 2.31 &                                 \bf 0.96 \\
          523\_analcatdata\_neavote &                  100 &                   3 &                                      NaN* &                                 \bf 1.15 \\
     527\_analcatdata\_election2000 &                   67 &                  15 &                                 0.50 &                                 \bf 0.49 \\
                       529\_pollen &                 3848 &                   5 &                                      NaN* &                                 \bf 0.94 \\
                       537\_houses &                20640 &                   9 &                                 3.00 &                                 \bf 1.06 \\
                    542\_pollution &                   60 &                  16 &                                 0.88 &                                 \bf 0.82 \\
\bottomrule
\end{tabular}
    }
    \caption{\pracmdltag\ vs Cross-validation for a range of datasets when the training data is limited. This table contains details about the datasets used in \cref{fig:pmlb,fig:pmlb_full,fig:pmlb_5fold}. The first column denotes the name of the datasets, and the second and third columns report the size of the datasets. In the last two columns, we report the performance for CV-tuned Ridge (LOOCV), and \pracmdltag\ tuned ridge estimator, trained on a subsampled dataset such that $d/n=1$. 
    The reported MSE is computed on a hold-out testing set (which consists of 25\% of the observations), and the better (of the two) MSE for each dataset is highlighted in bold.  We observe that \pracmdltag\ based tuning leads to superior performance compared to cross-validation for most of the datasets. *When too few points are available, cross-validation fails to numerically fit for low values of $\lambda$, returning a value of NaN.}
    \label{tab:datasets}
\end{table}

\section{Proofs} 
\label{sec:proofs}
This appendix serves to collect the proofs of all of our results. The
proofs for Theorems~\ref{thm:complexity_expressions}
through~\ref{thm:minimax_codelength} are provided in
\cref{sub:proof_of_thm:complexity_expressions} through
\cref{sub:proof_of_thm:minimax_codelength}, and
for Corollary~\ref{cor:mdl_kernel_smooth} in
\cref{sub:proof_of_cref_cor_mdl_kernel_smooth}.

\subsection{Proof of \cref{thm:complexity_expressions}} 
\label{sub:proof_of_thm:complexity_expressions}

Our proof is based on establishing that
\begin{align}
  \label{eq:ropt_claim}   
\lambda_i^{\textrm{opt}} = \frac{\noisevar}{\myvec_i^2} \quad
\mbox{and} \quad \ropt = \frac{1}{2n} \sumn[\minnd] \log\parenth{1 +
  \frac{\evali}{\lambda_i^{\textrm{opt}}}}.
\end{align}
The claimed expressions for $\ropt$ and \mdltag\ follow by using these
expressions and performing some algebra.

Let $\ptrue$ denote the distribution of the multivariate Gaussian
$\mc{N}(\xmat\tstar, \sigma^2\mb{I}_n)$, and let $\pden(\vy;\xmat,
\tstar)$ denote its density.  Note $\vy \sim \ptrue$ by assumption.\footnote{\label{proof:thm1_under_over_specified} In our our earlier discussion  with under-specified models in the simulations related to \cref{fig:mdl_gaussian}, and the discussion in \cref{sub:mdl_plots_details,sub:sketch_gaussian}, $\xmat$ denotes a subset of full features, in which case $\xmat\tstar$ does not capture the mean of the random vector $\vy$, and there is a bias. However, given the definition of our MDL-COMP, where we compare to the best possible encoding using $\xmat$ at hand, this bias term arises in both the numerator and denominator in \cref{eq:mdl_comp_proof_reatio}, and cancels out thereby not affecting the subsequent derivations.  On the other hand, with over-specified models, i.e., when $\xmat$ is a superset of features needed to correctly specify the mean of the random vector $\vy$, we can append zeros to the true parameter $\tstar$ as necessary, and continue to assume $\vy \sim \ptrue$.}
In order to simplify notation, we introduce the shorthand $\that =
\that_{\lammat}(\vy)$.

We have
\begin{align}
\label{eq:mdl_comp_proof_reatio}
  \kull{\pdist_{\tstar}}{\qdist_{\lammat}} & =\Exs_{\vy}\brackets{
    \log\parenth{\frac{ \frac{1}{(2\pi\noisevar)^{n/2}}
        \exp\parenth{-\frac{1}{2\noisevar}\enorm{\vy-\xmat\tstar}^2 }}
      {\frac{1}{\normconst(2\pi\noisevar)^{n/2}}
        \exp\parenth{-\frac{1}{2\noisevar}\enorm{\vy-\xmat\that}^2
          -\frac{1}{2\noisevar} \that\tp\lammat\that}} } } \\
          &= \;
  \sum_{j=1}^3 T_j, \notag
\end{align}
where
\begin{align}
\label{eq:thm_1_step_1}  
T_1  \defn -\Exs_{\vy} \brackets{\frac{1}{2\noisevar} \enorm{\vy
    - \xmat \tstar}^2}, \quad
T_2  \defn \Exs \brackets {\frac{1}{2 \noisevar}\enorm{\vy -
    \xmat\that}^2 + \frac{1}{2\noisevar} \that \tp \lammat \that},
\quad 
\end{align}
and $T_3 \defn \log \normconst$.  By inspection, we have $T_1 = -n/2$.
Dividing both sides by $\obs$ yields
\begin{align}
  \label{eq:exp_1}  
  \ropt = \min_{\lammat \in \mc{M}} \braces{\Exs\brackets{
      \frac{\enorm{\vy -\xmat \that}^2 + \that\tp \lammat \that
      }{2n\noisevar} } + \frac{1}{2n}\sumn[d]
    \log\parenth{1+\frac{\evali}{\lambda_i}} } - \frac{1}{2}
\end{align}
Next we claim that
\begin{subequations}
\begin{align}
    T_2 & = \frac{(n-\minnd)}{2} + \frac12\sumn[\minnd]
    \frac{(\evali\myvec_i^2/\noisevar +
      1)\lambda_i}{\lambda_i+\evali}, \qtext{and} \label{eq:t2}\\ 
      T_3 = \log \normconst
    &=
    \frac12\sumn[\minnd]\log\parenth{\frac{\evali+\lambda_i}{\lambda_i}}\label{eq:t3}
\end{align}
\end{subequations}
Assuming these claims as given at the moment, let us now complete the
proof.  We have
\begin{subequations}
    \label{eq:thm_1_kull}  
  \begin{align}    
    \frac{1}{n} \kull{\ptrue}{\qdist_{\lammat}} = \frac{1}{n}
    \sum_{j=1}^3 T_j & = -\frac{\minnd}{2n} + \frac{1}{2n}
    \sumn[\minnd] f_i(\lambda_i), \qquad \mbox{where} \\
f_i(\lambda_i) & \defn \parenth{ \frac{ (\evali\myvec_i^2/\noisevar +
    1)\lambda_i}{\lambda_i+\evali} +
  \log\parenth{\frac{\evali+\lambda_i}{\lambda_i}} }.
\end{align}
\end{subequations}
Finally, in order to compute $\ropt$ defined in
equation~\eqref{eq:exp_1}, we need to minimize the
KL-divergence~\eqref{eq:thm_1_kull} as a function of the vector
$\lambda = (\lambda_1, \ldots, \lambda_{\minnd})$.  Note that the
objective on the RHS of \cref{eq:thm_1_kull} is separable across the
components of $\lambda$, so we need only solve a univariate problem
for each $\lambda_i$.  Taking derivatives to find stationary points,
we have
\begin{align}
    \label{eq:lam_opt_mdl}  
f_i'(\lambda_i) = 0 \quad \Longleftrightarrow \quad
-\frac{(\evali\myvec_i^2/\noisevar + 1)}{(1+\evali/\lambda_i)^2} +
\frac{1}{1+\evali/\lambda_i} = 0 \quad\Longleftrightarrow\quad
\lambda_i^{\textrm{opt}} = \frac{\noisevar}{\myvec_i^2}.
\end{align}
Moreover, checking on the boundary of feasible values of $\lambda_i =
[0, \infty)$, we have $f_i(\lambda_i)\to \infty$ as $\lambda_i\to
  0^+$, and $f_i(\lambda_i) \to 1+\evali \myvec_i^2/\noisevar$ as
  $\lambda_i\to \infty$. Noting that $1 +\log a \leq a$ for all $a
  \geq 1$, we conclude that $f_i$ achieves its minimum at
  $\lambda_i^{\textrm{opt}}$, and we have
  $f_i(\lambda_i^{\textrm{opt}}) = 1+ \log
  (1+\evali\myvec_i^2/\noisevar)$. Substituting this value into the
  expression~\eqref{eq:lam_opt_mdl} yields
\begin{align}
\ropt & = -\frac{\minnd}{2n}+ \frac1{2n}\sumn[\minnd] \parenth{1+
  \log\parenth{1+\frac{\evali\myvec_i^2} {\noisevar}}}
\notag \\
& =\frac{1}{2n} \sumn[\minnd] \log\parenth{1+\frac{\evali\myvec_i^2}
  {\noisevar}}. \label{eq:ropt_expression}
\end{align}

We now turn to proving our earlier claims~\eqref{eq:t2} and
\eqref{eq:t3}.  We prove them separately. We start with the
low-dimensional case, i.e., when $d<n$. Since the proof for the
high-dimensional case is similar, we only outline the mains steps in
Appendix~\ref{ssub:proof_high_dim}.

\subsubsection{Proof of claims~\eqref{eq:t2} and \eqref{eq:t3} for the case $d<n$}
\label{ssub:proof_low_dim}

Let the singular value decomposition of the matrix $\xmat$ be given by
\begin{align}
\label{eq:svd}
    \mat X = \mat V \sqrt{\mat D} \mat U\tp.
\end{align}
Here the matrix $\mat D\in \real^{d\times d}$ is a diagonal matrix,
with $i$-th diagonal entry denoting the $i$-th squared singular value
of the matrix $\xmat$. Moreover, the matrices $\mat V \in \real^{n
  \times d}$ and $\mat U \in \real^ {d\times d}$, respectively,contain
the left and right singular vectors of the matrix $\mat X$,
respectively. (I.e., the $i$-th column denotes the singular vector
corresponding the $i$-th singular value.)  Note that we have the
relations $\mat U\mat U\tp = \mat U\tp \mat U = \mat V\tp\mat V = \mat
I_d$, and moreover, the matrix $\mat V \mat V\tp$ is a projection
matrix of rank $d$.  Finally, let $\templam = \diag(\lambda_1, \ldots,
\lambda_d)$ be such that $\lammat = \mat U \templam \mat U\tp $.  With
this notation in place, we claim that
\begin{subequations}
\label{eq:linear_algebra}
\begin{align}
  \label{eq:i_minus_xlam}  
    (\mat I_n - \xmat (\xtx + \lammat) \inv \xmat\tp)^2 &= \mat V
  \templam^2(\mat D + \templam)^{-2} \mat V \tp + (\mat I_n-\mat V
  \mat V\tp) \\
    \label{eq:x_lam}   
\xmat(\xtx+\lammat)\inv \lammat (\xtx+\lammat)\inv \xmat\tp &= \mat V
\mat D\templam (\mat D+\templam)^{-2} \mat V\tp, \quad \mbox{and}
\\
\label{eq:w_expand}
\xmat(\xtx+\lammat)\inv \xtx\tstar &= \mat V \sqrt{\mat D} (\mat D +
\templam)\inv \mat D \mat U\tp \tstar\notag \\ &= \mat V \sqrt{\mat D}
(\mat D + \templam)\inv \mat D \v{\myvec}
\end{align}
\end{subequations}
See Appendix~\ref{sub:few_linear_algebra_simplifications} for the
proofs of these claims, which involve elementary linear algebra.

\subsubsection{Proof of the expression~\eqref{eq:t2} for term $T_2$} 

We have
\begin{align}
\frac{1}{2\noisevar} & \enorm{\vy - \xmat \that}^2 + \frac{1}{2
  \noisevar} \that \tp \lammat \that \notag \\
& = \frac{1}{2 \noisevar} \enorm{\parenth{\mat I_n - \xmat (\xtx +
    \lammat)\inv \xmat\tp}\vy)}^2 + \frac{1}{2 \noisevar} \enorm{\sqrt
  \lammat (\xtx + \lammat)\inv \xmat\tp \vy }^2 \notag \\
& \stackrel{(i)}{=} \frac{1}{2 \noisevar} \vy \tp \parenth{\mat V
  \templam^2 (\mat D + \templam)^{-2} \mat V \tp} \vy + \frac{1}{2
  \noisevar} \vy \tp (\mat I_n-\mat V \mat V \tp) \vy + \frac{1}{2
  \noisevar} \vy \tp \parenth{\mat V \templam \mat D (\mat D +
  \templam)^{-2} \mat V \tp} \vy \notag \\
& = \frac{1}{2\noisevar} \vy \tp \parenth{\mat V (\templam^2+\templam
  \mat D) (\mat D + \templam)^ {-2} \mat V\tp} \vy +
\frac{1}{2\noisevar} \vy \tp \mat (\mat I_n-\mat V \mat V\tp) \vy
\notag \\
\label{eq:thm_1_step_2}
& =\frac{1}{2 \noisevar} \vy \tp \parenth{\mat V \templam (\mat D +
  \templam)^{-1} \mat V \tp} \vy + \frac{1}{2\noisevar} \vy \tp (\mat
I_n - \mat V \mat V \tp ) \vy,
\end{align}
where step~(i) follows from \cref{eq:i_minus_xlam,eq:x_lam}. The
latter steps make use of the fact that diagonal matrices commute.

Next we note that $\vy \sim \mc{N}(\xmat\tstar, \sigma^2\mb{I}_n)$,
and thus $\Exs[\vy\tp \mat A \vy ] = \tstar\tp\xmat\tp \mat A \xmat
\tstar + \sigma^2 \trace(\mat A)$.  Using
\cref{eq:thm_1_step_1,eq:thm_1_step_2}, we find that
\begin{align}
T_2 & = \Exs\brackets{\frac{1}{2\noisevar}\enorm{\vy-\xmat\that}^2 +
  \frac{1} {2\noisevar} \that\tp\lammat\that} \notag \\
    & = \frac{1}{2\noisevar} \tstar\tp\xmat\tp \parenth{\mat V
  \templam (\mat D + \templam)^ {-1} \mat V\tp}\xmat \tstar +
\frac{1}{2\noisevar} [\noisevar\trace(\mat V \templam (\mat D +
  \templam)^ {-1} \mat V\tp)] \notag \\
    & \qquad\qquad\qquad +\frac{1}{2\noisevar} \tstar\tp\xmat\tp (\mat
I-\mat V\mat V\tp)\xmat \tstar + \frac{1}{2\noisevar}
[\noisevar\trace(\mat I-\mat V\mat V\tp)] \notag \\
    & \stackrel{(i)}{=} \frac{1}{2\noisevar} \tstar\tp\xmat\tp
\parenth{\mat V \templam (\mat D + \templam)^ {-1} \mat V\tp}\xmat
\tstar + \frac{1}{2\noisevar} [\noisevar\trace(\mat V \templam (\mat D
  + \templam)^ {-1} \mat V\tp)]\label{eq:thm_1_step_3a} \\
    & \qquad\qquad\qquad + 0 +\frac{1}{2}\cdot (n-d) \notag \\
    & \stackrel{(ii)}{=} \frac1{2\noisevar}\parenth{\v{w}\tp \mat D
  \lammat (\mat D + \lammat)\inv \v{w} + \noisevar \trace [\templam
    (\mat D +\templam)\inv]} +\frac{1}{2}\cdot (n-d) \notag \\
\label{eq:thm_1_step_3}    
& = \frac{(n-d)}{2} +\frac{1}{2}\sumn[d]
\parenth{\frac{\evali\myvec_i^2 } {\noisevar} + 1} \frac{\lambda_i}
        {\lambda_i+\evali}
\end{align}
where step~(i) follows from the facts that the matrix $(\mat I_n -
\mat V\mat V\tp)$ is a projection matrix of rank $n-d$, and is
orthogonal to the matrix $\mat X$, i.e., $(\mat I_n - \mat V\mat
V\tp)\mat X = 0$.  Step~(ii) follows from a similar computation as
that done to obtain \cref{eq:thm_1_step_2}, along with
claim~\eqref{eq:w_expand}, and the following identity for the matrix
trace
\begin{align*}
    \trace(\mat V \templam (\mat D + \templam)^{-1} \mat V\tp) =
    \trace(\templam (\mat D + \templam)^{-1} \mat V\tp\mat V ) =
    \trace(\templam (\mat D + \templam)^{-1}\mat I_d).
\end{align*}

\subsubsection{Proof of claim~\eqref{eq:t3} (term $T_3$):}

From the normalization of a multivariate Gaussian density, we have
\begin{align}
\label{eq:gauss_norm}
    \frac{1}{(2\pi\noisevar)^{n/2}} \int_{\vy\in\real^n} \exp\parenth{-
    \frac{\vy\tp \mat A \vy}{2\noisevar}} d\vy
    = \sqrt{\det(\mat A\inv)},
\end{align}
valid for any positive definite matrix $\mat A \in \real^ {n \times
  n}$.

Putting together the definition~\eqref{eq:ridge_dist} of $\normconst$
and equation~\eqref{eq:thm_1_step_2}, we find that
\begin{align}
    \normconst = \frac{1}{(2\pi\noisevar)^{n/2}} \int_{\vy\in\real^n} \exp\parenth{-
    \frac{\vy\tp \mat A_{\lammat} \vy}{2\noisevar}} d\vy
    \qtext{where}
    \mat A_{\lammat} &= \parenth{\mat V \templam (\mat D + \templam)^{-1}
    \mat V\tp}
    +  (\mat I_n-\mat V \mat V\tp)  \notag\\
    &= \mat I_n - \mat V \mat D (\mat D + \templam)^{-1}\mat V\tp.
    \label{eq:a_lammat}
\end{align}
The eigenvalues of the $n \times n$ matrix $\mat A_{\lammat}$
are given by
$\braces{\frac{\lambda_1}{\lambda_1+\evali[1]},
\frac{\lambda_2}{\lambda_2+\evali[2]}, \ldots, \frac{\lambda_d}{\lambda_d+\evali
[d]},
1, 1, \ldots, 1}$, where the multiplicity of the eigenvalue $1$ is $n-d$.
Finally, applying \cref{eq:gauss_norm}, we find that
\begin{align}
    T_3 = \log \normconst = 
    \log\sqrt{\det(\mat A_{\lammat}\inv)}
    = 
    \frac12\sumn[d]\log\parenth{\frac{\evali+\lambda_i}{\lambda_i}}
    \label{eq:thm_1_step_4}
\end{align}
and the claim follows.

\subsubsection{Proof of claims~\eqref{eq:t2} and \eqref{eq:t3} for the case $d>n$}
\label{ssub:proof_high_dim}
In this case, the dimensions of the matrices in the singular value decomposition~\eqref{eq:svd} changes.
The argument for the proof remains similar with suitable adaptations due
to the change in the size of the matrices. As a result, we only outline the main steps.

We write
\begin{align}
     \label{eq:svd_new}
     \mat X = \mat V \begin{bmatrix}
         \sqrt{\widetilde{\mat  D}} & \mat 0 
     \end{bmatrix} \mat U\tp 
     \quad\Longrightarrow \quad
     \xtx = \mat U \underbrace{\begin{bmatrix}
         \widetilde{\mat D} & \mat 0 \\ \mat 0 & \mat 0
     \end{bmatrix}}_{\mat D} \mat U\tp
 \end{align}
 where $\mat V \in \real^{n\times n}$, $\widetilde{\mat D} \in \real^{n\times
 n}$ and $\mat U \in \real^{d\times d}$.
 Note that the non-zero entries of the matrix $\mat D$ are precisely the
 ones denoted by $\widetilde{\mat D}$.

Let $\templam_n = \diag(\lambda_1, \ldots, \lambda_n)$ denote the $n \times
n$ principal minor of the matrix $\templam$, where $\lammat = \mat U \templam
\mat U\tp$.
With these notations, we find that the claims~\eqref{eq:linear_algebra}
are replaced by
\begin{subequations}
\label{eq:linear_algebra_new}
\begin{align}
    (\mat I_n - \xmat (\xtx+\lammat)\inv \xmat\tp)^2
    &= \mat V \templam_n^2(\widetilde{\mat D} + \templam_n)^{-2} \mat V\tp
        \label{eq:i_minus_xlam_high}\\
    \xmat(\xtx+\lammat)\inv \lammat (\xtx+\lammat)\inv \xmat\tp
    &= \mat V \widetilde{\mat D}\templam_n (\widetilde{\mat D}+\templam_n)^
    {-2}  \mat V\tp 
    \label{eq:x_lam_high}\\
    \xmat(\xtx+\lammat)\inv \xtx\tstar &=
    \mat V \begin{bmatrix}
         \sqrt{\widetilde{\mat  D}} & \mat 0 
     \end{bmatrix} (\mat D + \templam)\inv \mat D
    \underbrace{\mat U\tp \tstar}_{=:\v{\myvec}}\notag
    \\
    &=  \mat V \sqrt{\widetilde{\mat D}} (\widetilde{\mat D} + \templam_n)\inv
    \widetilde{\mat D} \v{\myvec}_{1:n}
    \label{eq:w_expand_high}
\end{align}
\end{subequations}
where $\v{\myvec}_{1:n}$ denotes the first $n$ entries of the vector $\v{\myvec}$.
The proof of these claims can be derived in a similar manner to that of
claims~\eqref{eq:linear_algebra} (see Appendix~\ref{sub:few_linear_algebra_simplifications}).

Applying \cref{eq:i_minus_xlam_high,eq:x_lam_high}, we find that the equation~\eqref{eq:thm_1_step_2}
is modified to
\begin{align*}
    \frac{1}{2\noisevar}&\enorm{\vy-\xmat\that}^2 +\frac{1}{2\noisevar} \that\tp\lammat\that
    = \frac{1}{2\noisevar}\vy\tp \parenth{\mat V \templam_n (\widetilde{\mat
    D} +
    \templam_n)^{-1} \mat V\tp} \vy 
\end{align*}
which along with \cref{eq:w_expand_high} implies that the equation~\eqref{eq:thm_1_step_3}
is now replaced by
\begin{align*}
    T_2 &= \Exs\brackets{\frac{1}{2\noisevar}\enorm{\vy-\xmat\that}^2 +
    \frac{1}
    {2\noisevar} \that\tp\lammat\that}
    = \frac{1}{2}\sumn[n] \parenth{\frac{\evali\myvec_i^2 }
    {\noisevar} + 1} \frac{\lambda_i}{\lambda_i+\evali}.
\end{align*}
Thus, we obtain the claimed expression~\eqref{eq:t2} for
$T_2$ when $d>n$.

Next, to prove~\eqref{eq:t3} for this case, we find that the matrix $\mat
A_{\lammat}$ (defined in equation~\eqref{eq:a_lammat}) for this case gets
modified to
\begin{align*}
    \mat A_{\lammat} = \mat I_n -  \mat V \mat D (\mat D + \templam)^{-1}\mat V\tp.
\end{align*}
Since $\mat V\mat V\tp = \mat I_n$, we find that the eigenvalues of the
$n\times n$ matrix $\mat A_{\lammat}$ are given by $\braces{\frac{\lambda_1}{\evali [1]+\lambda_1},\ldots,\frac{\lambda_n}{\evali
[n]+\lambda_n}}$. Therefore, we conclude that
\begin{align*}
    T_3 = \log \normconst = 
    \log\sqrt{\det(\mat A_{\lammat}\inv)}
    = 
    \frac12\sumn[n]\log\parenth{\frac{\evali+\lambda_i}{\lambda_i}}.
\end{align*}

\subsubsection{Proof of claims~\eqref{eq:i_minus_xlam} and ~\eqref{eq:i_minus_xlam_high}} %
\label{sub:few_linear_algebra_simplifications}
For completeness, we discuss the proofs of the linear-algebra claims~\eqref{eq:linear_algebra}
and \eqref{eq:linear_algebra_new}.
Here we establish the claim~\eqref{eq:i_minus_xlam} (for $n>d$) and \eqref{eq:i_minus_xlam_high}
(for $n<d$). The other claims in the equations~\eqref{eq:linear_algebra}
and \eqref{eq:linear_algebra_new} can be derived
in a similar fashion.
Note that for both cases $n>d$ and $n<d$, our notations~\eqref{eq:svd} and
\eqref{eq:svd_new} are set-up such that 
\begin{align*}
    (\xtx+\lammat)\inv = \mat U (\mat D + \templam)\inv \mat U\tp,
\end{align*}
where the inverse is well-defined since $\templam = \diag(\lambda_1, \ldots,
\lambda_d)$ is assumed to be a positive definite matrix.
We use the fact that diagonal matrices commute, several times in our arguments.

\paragraph{Proof of claim~\eqref{eq:i_minus_xlam}:}
Using the decomposition~\eqref{eq:svd} for $n>d$ and noting that $\mat U\tp
\mat U = \mat I_d$, we find that
\begin{align}
    (\mat I_n - \xmat (\xtx+\lammat)\inv \xmat\tp)^2
    &=  (\mat I_n -  \mat V \sqrt{\mat D} \mat U\tp \mat U (\mat D + \templam)\inv
        \mat U\tp \mat U \sqrt{\mat D} \mat V\tp)^2 \notag\\
    &= (\mat I_n-\mat V\mat V\tp +\mat V \mat V\tp 
        -  \mat V  \sqrt{\mat D}(\mat D + \templam)\inv
         \sqrt{\mat D} \mat V\tp)^2 \notag\\
    &= (\mat I_n-\mat V\mat V\tp +\mat V (\mat I_d -\sqrt{\mat D}(\mat D
    + \templam)\inv
         \sqrt{\mat D}) \mat V\tp)^2 \notag\\
    &= (\mat I_n-\mat V\mat V\tp +\mat V \templam(\mat D + \templam)\inv
    \mat V\tp )^2 \notag\\
    &= \mat I_n-\mat V\mat V\tp +\mat V \templam^2(\mat D + \templam)^{-2}
    \mat V\tp,
    \notag
\end{align}
where the last step follows from the following facts (a) $\mat I-\mat V\mat
V\tp$ is a projection matrix,  (b) $(\mat I-\mat V\mat V\tp) \mat V = \mat
0$, and (c) $\mat V\tp\mat V = \mat I_d$.

\paragraph{Proof of claim~\eqref{eq:i_minus_xlam_high}:}
Note that for the decomposition~\eqref{eq:svd_new}, we have $\mat V\mat
V\tp = \mat V\tp \mat V = \mat I_n$ and $\mat U \mat U\tp = \mat U\tp \mat
U = \mat I_d$. Using these facts along with the notation $\templam_n
=\diag(\lambda_1, \ldots, \lambda_n)$ for $d>n$, we find that
\begin{align}
    (\mat I_n - \xmat (\xtx+\lammat)\inv \xmat\tp)^2
    &=  (\mat I_n -   \mat V \begin{bmatrix}
         \sqrt{\widetilde{\mat  D}} & \mat 0 
     \end{bmatrix} \mat U\tp  \mat U (\mat
    D + \templam)\inv
        \mat U\tp \mat U \begin{bmatrix}
            \sqrt{\widetilde{\mat D}} \\
            \mat 0
        \end{bmatrix} \mat V\tp )^2\notag\\
    &=  (\mat I_n -   \mat V \begin{bmatrix}
         \sqrt{\widetilde{\mat  D}} & \mat 0 
     \end{bmatrix} (\mat D + \templam)\inv
        \begin{bmatrix}
            \sqrt{\widetilde{\mat D}} \\
            \mat 0
        \end{bmatrix} \mat V\tp )^2\notag\\
    &=(\mat I_n - \mat V \sqrt{\widetilde{\mat D}} (\widetilde{\mat D}+\templam_n)\inv
    \sqrt{\widetilde{\mat D}}
    \mat V\tp)^2 \notag\\
    &=(\mat V \mat V\tp -  \mat V \widetilde{\mat D} (\widetilde{\mat
        D} + \templam_n)\inv \mat V\tp)^2 \notag \\
    &=(\mat V(\mat I_n-  \widetilde{\mat D} (\widetilde{\mat
        D} + \templam_n)\inv) \mat V\tp)^2 \notag\\
    &=\mat V \templam_n^2 (\widetilde{\mat
        D} + \templam_n)^{-2} \mat V\tp,
    \notag
\end{align}
which yields the claim.
\subsection{Proof of \cref{thm:lam_opt}}
\label{sub:proof_of_thm:lam_opt}
For $\that = \that_{\lammat}(\vy)$, we claim that
\begin{align}
\label{eq:in_sample_mse_exp}
    \frac{1}{n}\Exs\brackets{\enorm{\xmat\that -\xmat\tstar }^2}
    = \frac{1}{n}\sumn[\minnd] \underbrace{\frac{\lambda_i^2\evali\myvec_i^2
    + \noisevar
    \evali^2}{(\evali+\lambda_i)^2}}_{=:h_i(\lambda_i)}
\end{align}
Let us assume the claim as given at the moment, and prove \cref{eq:opt_lam_insample_mse}.
Recalling the optimal choice $\lambda_{i, {\textrm{opt}}}$
of the regularization parameter for \mdltag\ from equation~\eqref{eq:lam_opt_mdl},
and noting that the in-sample MSE
is separable in each term $\lambda_i$, it remains
to show that 
\begin{align}
\label{eq:g_opt}
    \argmin_{\lambda_i \in[0, \infty)} h_i(\lambda_i) = \lambda_{i, {\textrm{opt}}}
    = \frac{\noisevar}{\myvec_i^2}
\end{align}
To this end, we note that
\begin{align*}
    h_i'(\lambda_i) = 0 
    \quad\Longleftrightarrow\quad
    2\frac{\evali^2\myvec_i^2\lambda_i}{(\evali+\lambda_i)^3}
    -2\frac{\evali^2\noisevar}{(\evali+\lambda_i)^3}
     = 0
    \quad\Longleftrightarrow\quad
    \widetilde{\lambda}_{i, \textrm{opt}} = \frac{\noisevar}{\myvec_i^2}.
\end{align*}
On the boundary of feasible values of $\lambda_i = [0,
\infty)$, we have $h_i(0) = \noisevar$, and $h_i(\lambda_i) \to \evali\myvec_i^2$
as $\lambda_i\to
\infty$. Noting that 
\begin{align*}
    h_i(\widetilde{\lambda}_i^{\textrm{opt}})  = \frac{\evali\myvec_i^2\noisevar}
    {\evali\myvec_i^2+\noisevar}
    \leq \min\braces{\evali\myvec_i^2, \noisevar},
\end{align*}
yields the claim~\eqref{eq:g_opt}. 

Next we establish the bound~\eqref{eq:ropt_bounds_insample_mse} on the
fixed design prediction error. Using $\lambda_i$ as a convenient
shorthand for $\lambda_{i, {\textrm{opt}}}$, we substitute it into the
RHS of \cref{eq:in_sample_mse_exp}, thereby finding that
\begin{align*}
  \frac{1}{n} \Exs\brackets{\enorm{\xmat \that_{\lammatopt} - \xmat
      \tstar }^2} & = \frac{1}{n} \sumn[\minnd] \frac{ (
    \frac{\noisevar}{\myvec_i^2})^2 \evali \myvec_i^2 + \noisevar
    \evali^2}{(\evali + (\frac{\noisevar}{\myvec_i^2} ) )^2} \\
  & = \frac{\noisevar}{n} \sumn[\minnd] \frac{1}{1 +
    \frac{\noisevar}{\evali \myvec_i^2}}.
\end{align*}
We now make note of the elementary inequality $x \leq -\log(1- x)$
valid for $x \in [0, 1)$.  Applying this inequality with $x = (1 +
  \frac{\noisevar}{\evali \myvec_i^2})^{-1}$, we have $-\log(1-x) =
  \log (1 + \tfrac{\evali \myvec_i^2}{\noisevar})$, and hence
\begin{align*}
 \frac{1}{n} \Exs \brackets{\enorm{\xmat \that_{\lammatopt} - \xmat
     \tstar}^2 } & \leq \frac{\noisevar}{n} \sumn[\minnd] \log
 \parenth{ 1 + \frac{\evali\myvec_i^2}{\noisevar}} =
 \frac{\noisevar}{n} \ropt,
\end{align*}
  where we have used $\ropt$ as a shorthand for the optimal redundancy.

\paragraph{Proof of \cref{eq:in_sample_mse_exp}:}
There are two cases to be considered, namely $d > n$ and $d < n$.
Both cases can be handled together by using the linear-algebraic
claims~\eqref{eq:linear_algebra} and \eqref{eq:linear_algebra_new} as
needed.  Following steps similar to those in the proof of
Theorem~\ref{thm:complexity_expressions}, we find that
\begin{align*}
  \Exs \brackets{\enorm{\xmat\that -\xmat\tstar }^2} &= \Exs
  \brackets{\enorm{\xmat(\xtx+\lammat)\inv \xmat\tp \vy -\xmat\tstar
    }^2} \\ &= \enorm{\xmat(\xtx+\lammat)\inv \xtx\tstar -\xmat\tstar
  }^2 + \Exs\brackets{\enorm{\xmat(\xtx+\lammat)\inv
      \xmat\tp\noise[]}^2} \\
& \stackrel{(i)}{=} \enorm{\mat V \sqrt{\mat D} ((\mat D+\templam)\inv
    \mat D -\mat I)\v{\myvec} }^2 + \Exs\brackets{\enorm{\mat V
      \sqrt{\mat D}(\mat D + \templam)\inv \sqrt{\mat D} \mat V\tp
      \noise[]}^2} \\
  & = \enorm{\sqrt{\mat D} ((\mat D+\templam)\inv \templam)\v{\myvec}
  }^2 + \noisevar\trace\parenth{\sqrt{\mat D} (\mat D + \templam)\inv
    \mat D (\mat D+\templam)\inv \sqrt{\mat D}} \\
  & = \sumn[\minnd] \parenth{\frac{ \evali \lambda_i^2
      \myvec_i^2}{(\evali + \lambda_i)^2} + \frac{\noisevar
      \evali^2}{(\evali + \lambda_i)^2}},
\end{align*}
and we are done.

\subsection{Proof of Theorem~\ref{thm:mdl_kernel}} %
\label{sub:proof_of_theorem_thm:mdl_kernel}

Let us introduce the shorthand $\ystar = (\fstar(x_1), \ldots,
\fstar(x_n))\tp$. Given data $\vy$ generated according to the
model~\eqref{eq:kernel_Generative_model}, we have
$\vy\sim\mc{N}(\ystar, \sigma^2\mb{I}_n)$.  Let $\pden(\vy;\ystar)$
denote the corresponding density of $\vy$. With this notation, we can
write
\begin{align*}
  \kull{\pdist_{\fstar}}{\qdist_{\lam}} &=
  \Exs_{\vy}\brackets{\log{\frac{\pden(\vy;\ystar)}
      {\qden_{\lam}(\vy)} } } \\
& = \underbrace{-\Exs_{\vy}\brackets{\frac{1}{2\noisevar} \enorm{\vy -
        \ystar}^2}}_{=:T_1} + \underbrace{\Exs_{\vy}
    \brackets{\frac{1}{2 \noisevar}\enorm{\vy - \kmat \that}^2 +
      \frac{\lam}{2 \noisevar} \that \tp \kmat \that}}_{=:T_2} +
  \underbrace{\log \normconstk}_{=:T_3}.
\end{align*}
Note that the term $T_1 = - n/2$ by definition of our noisy
observation model. Turning to the term $T_2$, we substitute the
expression~\eqref{eq:kernel_ridge2} for $\that$, thereby finding that
\begin{align*}
\enorm{\vy-\kmat\that}^2 + \lam\enorm{\that}^2 &= \vy\tp \lam^2
(\kridge)^ {-2} \vy + \lam \vy\tp (\kridge)\inv \kmat (\kridge)\inv
\vy \\ &= \lam\vy\tp (\kridge)\inv \vy
\end{align*}
and thus
\begin{align*}
  T_2 =\frac{1}{2\noisevar} \Exs[\enorm{\vy-\kmat\that}^2 + \lam\enorm{\that}^2]
  = \frac{\lam}{2\noisevar} \ystar\tp (\kridge)\inv \ystar  + \frac12\sum_
  {i=1}^
  {n} \frac{\lam}{\evali+\lam}.
\end{align*}
Moreover, using an argument similar to that used in \cref{eq:thm_1_step_4},
we can write
\begin{align}
    T_3 = \log \normconstk = 
    \log\sqrt{\det(\frac{1}{\lam} (\kridge))}
    = 
    \frac12\sumn[n]\log\parenth{\frac{\evali+\lambda_i}{\lambda_i}}.
    \notag
\end{align}
Using the eigenvalue decomposition $\kmat = \mat U \mat D \mat U\tp$ where
$\mat D = \diag(\evali[1], \ldots, \evali[n]) $,
and setting $\alpha := \mat U\tp \v{y^\star}$, we thus find that
\begin{align}
\label{eq:mdl_kridge}
    \frac{1}{n}\kull{\mb{P}_{\fstar}}{\qdist_{\lam}}
    = \frac{1}{2n}\sumn \frac{\lam}{\lam+\evali}\parenth{\frac{\alpha_i^2}{\sigma^2}+1}
    + \frac{1}{2n}\sumn \log\parenth{\frac{\evali}{\lam}+1}-\frac{1}{2}.
\end{align}

The remaining proof make use of arguments similar to those in
\citet[Lemma 7]{raskutti2014early}.  Recall the
decomposition~\eqref{eq:mercer_kernel} for the kernel $\kernel$.  For
any function $\fstar \in \rkhs$, we can write
\begin{align*}
\fstar(x) = \sum_{k=1}^{\infty} \sqrt{\mu_k} a_k \phi_k(x)
\qtext{with} a_k = \frac{1}{\sqrt{\mu_k}} \int f(z) \phi_k(z)d\nu(z)
\qtext{and} \enorm{\fstar}^2_{\rkhs} = \sum_{k=1}^\infty a_k^2,
\end{align*}
where $\nu$ denotes the marginal distribution of the covariates.
Define the linear operators $\Psi_{\xmat}:\ell^2(\mb{N}) \to \real^n$
as $\Psi_{\xmat}[j, i] = \phi_i(x_j)$ for $i \in \mb{N}$ and $j \in
[n]$ (the matrix $\Psi_{\xmat}$ has $n$ rows and infinite columns).  and the diagonal linear operator $\mathfrak{D}:\ell^2(\mb{N})
\to \ell^2 (\mb{N})$ as $\mathfrak{D}[i, i]=\mu_i$ and
$\mathfrak{D}[i, j]=0$ if $i \neq j$ for $i, j \in \mb{N}$.  Given
this representation, we can rewrite the vector $\ystar$ and the kernel
matrix $\kmat$ as follows
\begin{align*}
  \ystar =\Psi_{\xmat} \mathfrak{D}^{\frac12} a 
  \qtext{and}
  \kmat = \Psi_{\xmat} \mathfrak{D} \Psi_{\xmat}\tp 
\end{align*}
Combining the latter representation with $\kmat = \mat U \mat D \mat U\tp$,
we find that there exists an operator $\Gamma: \ell^2(\mb{N})\to \real^n$
with adjoint $\Gamma^*:\real^n \to \ell^{2}(\mb{N})$ and $\Gamma\Gamma^*=
\mat I_{n}$ such that 
\begin{align*}
  \Psi_{\xmat} \mathfrak{D}^{\frac12} = \mat U \mat D^{\frac12} \Gamma
  \implies \alpha = \mat U\tp \ystar = \mat U\tp \Psi_{\xmat} \mathfrak{D}^{\frac12} a
  = \mat U\tp \mat U \mat D^{\frac12} \underbrace{\Gamma a}_{=:\beta}
  = \mat D^{\frac12}\beta.
\end{align*}
Note that $\beta \in \real^n$ and we have $\alpha_i^2
= \evali \beta_i^2$. Furthermore,
\begin{align}
\label{eq:beta_norm}
    \sumn\beta_i^2 = \enorm{\Gamma a}_2^2\leq \enorm{a}_2^2 = \enorm{\fstar}_{\rkhs}^2,
\end{align}
since $\Gamma$ is a unitary operator.  Thus, we can
rewrite~\eqref{eq:mdl_kridge} as
\begin{align}
\frac{1}{n} \kull{\mb{P}_{\fstar}}{\qdist_{\lam}} & = \frac{1}{2n}
\sumn \frac{\lam}{\lam + \evali}
\parenth{\frac{\evali\beta_i^2}{\sigma^2}+1} + \frac{1}{2n} \sumn
\log\parenth{\frac{\evali}{\lam}+1}-\frac{1}{2} \notag \\
& = \frac{1}{2n} \lam \sumn
\frac{\evali}{\lam+\evali}\frac{\beta_i^2}{\sigma^2} + \frac{1}{2n}
\sumn \log\parenth{\frac{\evali}{\lam}+1} + \frac{1}{2n} \sumn
\frac{\lam}{\lam+\evali}- \frac{1}{2} \notag \\
& \sless{(i)} {\frac{\lam}{2n} \parenth{\frac{\sumn
      \beta_{i}^2}{\sigma^2}} + \frac{1}{2n}\sumn \log\parenth{
    \frac{\evali}{\lam}+1}} \notag\\
\label{eq:final_bound_ropt_kernel}
& \sless{(ii)} {\frac{\lam}{2n} \frac{
    \enorm{\fstar}^2_{\rkhs}}{\sigma^2} + \frac{1}{2n} \sumn \log
  \parenth{ \frac{\evali}{\lam} + 1}},
\end{align}
where step~(i) follows from the fact that $\max\braces{\evali,
  \lam}/(\lam+\evali)\leq 1$, and step~(ii) from the
bound~\eqref{eq:beta_norm}. The proof is now complete.
 
\paragraph{Remark:}
Let us discuss some scenarios in which the inequalities in (i) and
(ii) are tight after optimizing the
bound~\eqref{eq:final_bound_ropt_kernel} over $\lam$.  Call the
optimal choice $\lam_{\mrm{opt}}$, and let $i^\star$ denote the index
such that $\lam_{\mrm{opt}} \approx \evali[i^\star]$.  The inequality
in step~(i) is relatively tight when: (a)
$\sum_{i=1}^{n}\lam/(\lam+\evali) \approx n$; and (b)
$\enorm{\beta}_2^2 \approx \sum_{j\leq i^\star} \beta_j^2$.  Condition
(a) holds if the eigenvalue $\braces{\evali}$ have suitablly rapid
decay.  whereas condition (b) holds when $i^\star < n$ and the
$\{\beta_j\}$ sequence is suitably decaying.

In order for inequality~(ii) to be relatively tight, we need the
inequality~\eqref{eq:beta_norm} to be relatively tight.  This property
holds when we can obtain a good approximation of the
sum~\eqref{eq:mercer_kernel} by summing only its first $n$ terms; in
this case, $\Gamma$ and $\Gamma^\star$ are simply $n \times n$ unitary
matrices. This property holds when either the eigenvalues $\{\mu_k\}$
decay quickly, or the kernel $\kernel$ has a finite rank strictly
smaller than $n$, in which case $\mu_k=0$ for $k \geq n$. Note that
the different decay conditions covered in
Corollary~\ref{cor:mdl_kernel_smooth} are all sufficient for these
kernel conditions to hold.

\subsection{Proof of \cref{thm:rmt_expressions}} 
\label{sub:proof_of_thm:rmt_expressions}

Our result involves the function
\begin{align}
\label{eq:g_fun}
g(a, b) \defn \frac{1}{4} \parenth{\sqrt{\snr (1 + \sqrt{\gamma} )^2 +
    1} - \sqrt{\snr(1 - \sqrt{\gamma})^2 + 1}}^2, \qtext{for} a, b >
0.
\end{align}
The proof of this theorem makes use of the analytical expression of
$\mdltag$ from Theorem~\ref{thm:complexity_expressions} and few random
matrix theory results.  Let $\mb{F}_d: a \mapsto
\frac{1}{d}\sumn[d]\mathbf{1}(a\leq\evali)$ denote the empirical
distribution of the eigenvalues of the matrix $\xtx$.

First, we note that since $\tstar$ is assumed to be drawn from
rotationally invariant distribution, we can rewrite the MDL-complexity
expression as
\begin{align*}
  \Exs_{\tstar}[\ropt] = \Exs_{\v{\myvec}}\brackets{\frac{1}{n}
    \sumn[\minnd] \log \parenth{1 + \frac{\evali\myvec_i^2}
      {\noisevar}}} & \stackrel{(i)}{\leq} \brackets{\frac{1}{n}
    \sumn[\minnd] \log \parenth{1 + \frac{\evali
        \Exs_{\v{\myvec}}[\myvec_i^2]} {\noisevar}}} \\
& \stackrel{(ii)}{=} \frac{1}{n} \sumn[d] \log\parenth{1 + \evali
    \cdot \snr} \\
& = \frac{d}{n} \parenth{\frac{1}{d} \sumn[d] \log \parenth{1+\evali
      \cdot \snr}} \\
& = \frac{d}{n} \cdot \int_{Z=0}^\infty \log(1+Z\cdot \snr) d \mb{F}_d
  (Z),
\end{align*}
where step~(i) follows from Jensen's inequality, i.e.,
$\Exs[\log(1+W)] \leq \log (1+\Exs[W])$, and step~(ii) follows from
the fact that $\tstar$ is drawn from rotationally-invariant
distribution and hence $\v{\myvec} = \mat U\tp \tstar \stackrel{d}{=}
\tstar$, and that $\Exs[\myvec_i^2] = \Exs[\enorm{\tstar}^2]/d$.

Next, we recall a standard result from random matrix theory.  Under
the assumptions of Theorem~\ref{thm:rmt_expressions}, as $d, n\to
\infty$ with $d/n\to \gamma$, the empirical distribution $\mb{F}_d$ of
the eigenvalues of the matrix $\xtx$ converges weakly, almost surely
to a nonrandom limiting distribution $\mb{MP}_\gamma$ with density
\begin{align}
\label{eq:mp_law}
 m_{\gamma}(a) = \parenth{1-\frac{1}{\gamma}}_{+} \delta(a) +
 \frac{\sqrt{(a-b_1)_+(b_2-a)_+}}{2\pi\gamma a}, \qtext{where $b_1 =
   (1-\sqrt\gamma)^2$ and $b_2 = (1+\sqrt\gamma)^2$.}
\end{align}
This distribution is also known as the
Mar$\breve{\textrm{c}}$enko-Pastur law with ratio index $\gamma$. (For
background, see the
papers~\citep{marvcenko1967distribution,silverstein1995strong}, or
Theorem 2.35 in the book~\citep{tulino2004random})

Next, we claim that as $d, n\to \infty$ with $d/n\to \gamma$, we have
\begin{align}
\label{eq:rmt_claim}
    \ropt 
    = \frac{d}{n} \cdot \int_{0}^\infty \log(1+Z\cdot \snr) d\mb{F}_d(Z)
    \longrightarrow  
    \gamma\cdot \Exs_{Z\sim \mb{MP}_\gamma}\brackets{\log(1+\snr\cdot Z)},
\end{align}
almost surely.  Assuming this claim as given at the moment, let us now
complete the proof.  In general, the analytical expressions for
expectations under the distribution $\mb{MP}_{\gamma}$ are difficult
to compute. However, the expectation $\Exs_{Z\sim \mb{MP}_\gamma}
\brackets{\log(1+\gamma Z)}$ is known as the Shannon transform and
there exists a closed form for it (see Example
2.14~\citep{tulino2004random})\footnote{The transform in the
reference~\citep{tulino2004random} uses the $\log_2$ (base-$2$
logarithm) notation. The results stated here have been adapted for
$\log_e$ (base-$e$, natural logarithm) in a straightforward manner.}:
\begin{align}
    \Exs_{Z\sim \mb{MP}_\gamma}\brackets{\log(1+\snr\cdot Z)}
    =: \mc{V}_Z(\snr)
    = \log\parenth{1+\snr-\delta} + \frac{1}{\gamma}\log\parenth{1+\gamma\cdot\snr
    -\delta} - \frac{\delta}{\snr\cdot \gamma},
    \label{eq:shannon_transform}
\end{align}
where $\delta = g(\snr, \gamma)$ and the function $g$ was defined in equation~\eqref{eq:g_fun}.
Putting the \cref{eq:rmt_claim,eq:shannon_transform} together yields the
expression~\eqref{eq:mdl_complexity_rmt} stated in Theorem~\ref{thm:rmt_expressions}.

\paragraph{Proof of claim~\eqref{eq:rmt_claim}:}
This part makes use of the Portmanteau theorem and some known results
from~\cite{bai2008limit}.  Recalling the constant $b_2 \defn (1+\sqrt
\gamma)^2$ from equation~\eqref{eq:mp_law}, for $a \in [0, \infty)$,
  consider the function $f(a) \log (1+\snr\cdot a) \mathbf{1} (a\leq
  2b_2) $ As $n, d\to \infty$ with $d/n\to\gamma$, we have
\begin{align*}
 \int_{0}^\infty f(Z) d\mb{F}_d(Z) \stackrel{(i)}{=} \int_{0}^{2b_2}
 f(Z) d\mb{F}_d(Z) \stackrel{(ii)}{\longrightarrow} \int_{0}^{2b_2}
 f(Z) d\mb{MP}_\gamma(Z) &\stackrel{(iii)}{=} \int_{0}^{\infty} f(Z)
 d\mb{MP}_\gamma(Z) \\ &= \Exs_{Z\sim\mb{MP}_\gamma}[1+\log(\snr\cdot
   Z)],
\end{align*}
where steps~(i) and (iii) follow from the definition of $f$, i.e.,
$f(Z)=0$ for any $Z>2b+2$.  Moreover, step~(ii) follows from applying
the Portmanteau theorem, which states that the weak almost sure
convergence~\eqref{eq:mp_law} implies the convergence in expectation
for any bounded function, and $f$ is a bounded function.  Finally, we
argue that for large enough $n$, we have
\begin{align*}
    \int_{0}^\infty \log(1+Z\cdot \snr) d\mb{F}_d(Z)
    \stackrel{(iv)}{=} \int_{0}^{2b_2} \log(1+Z\cdot \snr)
    d\mb{F}_d(Z) \stackrel{(v)}{=} \int_{0}^\infty f(Z) d\mb{F}_d(Z)
\end{align*}
In order to establish step~(iv), we invoke a result from the
work~\citep{bai2008limit} (Corollary 1.8). It states that for large
enough $n$, the largest eigenvalue of the matrix $\xtx$ is bounded
above by $2(1+\sqrt\gamma)^2$ almost surely, and thus, we can restrict
the limits of the integral on the LHS to the interval $[0, 2b_2]$.
Step~(v) follows trivially from the definition of the function $f$.
Putting the pieces together yields the claim~\eqref{eq:rmt_claim}.

\subsection{Proof of \cref{thm:minimax_codelength}} 
\label{sub:proof_of_thm:minimax_codelength}

The proof of this theorem makes use of the algebra used in previous
proofs, and hence we simply illustrate the main steps.  We have
\begin{align*}
\E \brackets{\log \parenth {\frac{1}{\qden_{\lammat}(\vy)}}} & = \Exs
\brackets{\frac{1}{2\noisevar}\enorm{\vy - \xmat \that}^2 +
  \frac{1}{2\noisevar} \that \tp \lammat \that} + \log \normconst,
\end{align*}
where \mbox{$\log \normconst = \frac{1}{2} \sumn[\minnd]
  \log\parenth{\frac{\evali+\lambda_i}{\lambda_i}}$.}  Repeating
arguments similar to those in \cref{eq:thm_1_step_3a,eq:thm_1_step_3},
we find that
\begin{align*}
\notag & \Exs\brackets{\frac{1}{2 \noisevar} \enorm{ \vy - \xmat
    \that}^2 + \frac{1}{2 \noisevar} \that \tp \lammat\that} \\
& = \frac{1}{2\noisevar} \tstar \tp \xmat \tp \parenth{\mat V \templam
  (\mat D + \templam)^ {-1} \mat V\tp} \xmat \tstar +
\frac{1}{2\noisevar}\trace \brackets{\textrm{Var}(\vy\vert\xmat) \mat
  V \templam (\mat D + \templam)^{-1} \mat V\tp } \notag \\
& \qquad\qquad\qquad + \frac{1}{2\noisevar} \trace
\brackets{\textrm{Var}(\vy \vert \xmat)(\mat I - \mat V \mat V\tp)}
\notag \\
& \stackrel{(i)}{\leq} \frac1{2\noisevar}\parenth{\v{w} \tp \mat D
  \lammat (\mat D + \lammat) \inv \v{w}} + \frac{1}{2 \noisevar}\trace
\brackets{\noisevar \mat I_n {\textrm{Var}(\vy \vert \xmat)} \mat V
  \templam (\mat D + \templam)^{-1} \mat V \tp } \notag \\
& \qquad\qquad\qquad + \frac{1}{2 \noisevar} \trace
\brackets{\noisevar \mat I_n (\mat I - \mat V \mat V \tp)} \notag \\
& =\frac{(n -\min \braces{n, d})}{2} +\frac{1}{2}\sumn[d]
\parenth{\frac{\evali \myvec_i^2 } {\noisevar} + 1} \frac{\lambda_i}
        {\lambda_i + \evali}
\end{align*}
where step~(i) follows from the following linear algbera fact: For
matrices $\mat A, \mat B, \mat C \succeq 0$, if $\mat A \succeq \mat
B$, then $\trace[\mat{AC}] \geq \trace[\mat{BC}]$. Moreover, noting
that in step~(i), we have equality when $\text{Var}(\vy\vert\xmat) =
\noisevar \mat I_n$. Putting the pieces together, we conclude that
\begin{align*}
  \max_{\P \in \plin} \E \brackets{\log
    \parenth{\frac{1}{\qden_{\lammat}(\vy)}}} = \frac{(n - \min
    \braces{n, d})}{2} + \frac{1}{2}\sumn[d] \parenth{\frac{\evali
      \myvec_i^2 } {\noisevar} + 1} \frac{\lambda_i} {\lambda_i +
    \evali} + \frac{1}{2} \sumn[\minnd] \log \parenth{\frac{\evali +
      \lambda_i}{\lambda_i}}.
\end{align*}
As function of $\braces{\lambda_i}$ in comparison to the function
\cref{eq:thm_1_kull}, this objective is a simply shifted by a
constant.  Consequently, it admits the same minimizer
\begin{align*}
  \arg \min_{\qdist_{\lammat} \in \qset_{\tagridge}} \max_{\P \in
    \plin} \E \brackets{\log \parenth{\frac{1}{\qden_{\lammat}(\vy)}}}
  = \qdist_{\lammatopt},
\end{align*}
as claimed.

\subsection{Proof of Corollary~\ref{cor:mdl_kernel_smooth}} 
\label{sub:proof_of_cref_cor_mdl_kernel_smooth}

We prove the bounds for polynomial and exponential decay of
eigenvalues separately.

\subsubsection{Proof with polynomial decay of eigenvalues}

When $\kernel(x, x)=1$, we have $\trace(\kmat) = \sum_{i=1}^n\evali =
n$. Thus, with $\evali \precsim i^{-2\alpha}$, we can conclude that
$\evali \leq c_\alpha n i^{-2\alpha} $ where
$c_{\alpha}:=\sum_{j=1}^nj^{-2\alpha} \leq 1/(2\alpha-1)$, where the additional factor of $n$ arises due to the constraint that $\trace(\kmat) = \sum_{i=1}^n\evali =
n$.

Since the MDL-COMP is an increasing function in $\evali$, it suffices
to consider the case when $\evali = n c_\alpha i^{-2\alpha}$. Setting
$\alpha = \omega/d$ covers the case of $\evali \precsim
i^{-2\omega/d}$, and setting $\alpha = \frac{1}{2} (d+a)$ covers the
case of $\evali \precsim i^{-(d+a)}$. With the assumptions $\omega >
d/2$ and $d + a >1$, we find that for both cases $2 \alpha - 1 >0$ and
hence $c_{\alpha}$ is finite.

Let us write $\lambda$ as $nM^{-2\alpha}$, and given the rapid decay
of $\evali$, we can use the bounds
\begin{align*}
  \sum_{i=1}^{\ceil{M}} \log(\evali/\lambda+1) \leq
  \sum_{i=1}^{\ceil{M}} \log((M/i)^{2\alpha}+1) \stackrel{(i)}{\leq}
  c_1 \cdot \alpha M \log M
  \\
\sum_{i=\ceil{M}+1}^n \log(\evali/\lambda+1) \leq
\sum_{i=\ceil{M}+1}^{n} \log((M/i)^{2\alpha}+1) \leq
\sum_{i=\ceil{M}+1}^{n} (M/i)^{2\alpha} \leq c_\alpha.
\end{align*}
where step~(i) uses the fact that $\int_{1}^{x} \log z dz = x \log x -
x \leq x \log x$ for $x > 1$, and step~(ii) uses the fact that
$\sum_{i=\ceil{M}+1}^{n}(M/i)^{2\alpha} \leq
\sum_{j=1}^{\infty}j^{-2\alpha} = c_{\alpha}$.  Thus, we can write the
RHS in \cref{eq:mdl_kernel} as
\begin{align*}
  \frac{\lam}{2n} \frac{\enorm{\fstar}^2_{\rkhs}}{\sigma^2} +
  \frac{1}{2n} \sumn \log\parenth{\frac{\evali}{\lam}+1} \leq
  \frac{1}{2n} \brackets{n M^{-2\alpha}\SNR^2 + c \alpha M\log M +
    c_{\alpha}}
\end{align*}
In order to minimize the RHS above with respect to $M$, we can equate
the two terms inside the brackets, and obtain
\begin{align*}
  \frac{n \SNR^2}{\alpha} \asymp M^{1+2\alpha}\log M
  \quad\Longleftrightarrow\quad M \asymp \left(\frac{{n
      \SNR^2}/{\alpha}}{\log({n
      \SNR^2}/{\alpha})}\right)^{1/(1+2\alpha)}
\end{align*}
which yields
\begin{align*}
  \mdltag \asymp M^{-2\alpha}\SNR^2 & = \left(\frac{{n
      \SNR^2}/{\alpha}}{ \log({n
      \SNR^2}/{\alpha})}\right)^{-2\alpha/(1+2\alpha)} \SNR^2 \\
  & = c'_{\alpha} \left(\frac{ \log({n
      \SNR^2})}{n}\right)^{2\alpha/(1+2\alpha)} \SNR^{2/(1+2\alpha)}
  \\
  & = C_{\alpha, \SNR} \left( \frac{\log
    (n\SNR^2)}{n}\right)^{2\alpha/(1+2\alpha)},
\end{align*}
where
\begin{align}
\label{eq:constant_ker_smooth}
    C_{\alpha, \SNR} = \alpha^{2\alpha/(1+{2\alpha})} \SNR^{2/(1+2\alpha)}.
\end{align}
Setting $\alpha = \omega/d$ and $\frac12(d+a)$ respectively recovers
the first two bounds stated in the
display~\eqref{eq:mdl_kernel_smooth}.

\subsubsection{Proof with exponential decay of eigenvalues}

Now, we do a reparametrization of $\lambda$ as $n\exp(-M/d)$, and
given the rapid decay of $\evali$, we use the following bounds:
\begin{align*}
  \sum_{i=1}^{\ceil{M}}\log(\evali/\lambda+1) \leq
  \sum_{i=1}^{\ceil{M}}\log(e^{(M-i)/d}+1) \stackrel{(i)}{\leq}
  \frac{c}{d} \cdot \sum_{i=1}^{\ceil{M}} (M-i) \leq c'
  \frac{M^2}{d} \\
\sum_{i={\ceil{M}}+1}^n\log(\evali/\lambda+1) =
\sum_{i={\ceil{M}}+1}^{n}\log(e^{(M-i)/d}+1) \leq
\sum_{i={\ceil{M}}+1}^{n} e^{(M-i)/d} \leq c_d''.
\end{align*}
where we use the fact that $\log(1+ e^{x}) \leq c x$ for large $x$,
and $\log(1+e^x) \leq e^x$ for all $x$. Thus, we can write the RHS in
\cref{eq:mdl_kernel} as
\begin{align*}
  \frac{\lam}{2n} \frac{\enorm{\fstar}^2_{\rkhs}}{\sigma^2} +
  \frac{1}{2n}\sumn \log\parenth{\frac{\evali}{\lam}+1} \leq
  \frac{1}{2n}\brackets{ne^{-M/d}\SNR^2 + c' \frac{M^2}{d} + c_{d}''}
\end{align*}
In order to minimize the RHS of the above display with respect to $M$,
we can equate the two terms inside the brackets, and obtain
\begin{align*}
  d n \SNR^2 \asymp e^{M/d} M^2 \quad\Longleftrightarrow\quad M \asymp
  d \log\parenth{\frac{nd\SNR^2}{d\log(nd\SNR^2))}}.
\end{align*}
Thus, we have
\begin{align*}
  \mdltag \asymp \frac{M^2}{dn} &= \frac{d\log^2(nd \SNR^2)}{n},
\end{align*}
as claimed.

\section{Bias-variance tradeoff: Role of estimator and design matrix}
\label{sub:different_design}

\setcounter{table}{0} \setcounter{figure}{0}
\renewcommand{\thefigure}{F\arabic{figure}}
\renewcommand{\thetable}{F\arabic{table}}

In this appendix, we show that the bias-variance tradeoff for OLS
heavily depends on the design matrix.  More precisely, depending on
the structure of the design matrix, it is possible to observe
double-descent or multiple descent, where the peaks can occur at
values of $d/n$ not necessarily equal to $1$ in the test MSE when
varying $d$.  While OLS can exhibit a wide range of behavior, we also
show that the test MSE for CV-tuned ridge regression exhibits
the familiar U-shaped behavior for all the choices of design
considered here.

In the experimental results shown here, we generate data from a linear
Gaussian model of the form
\begin{align}
\vy = \underbrace{\xmat \tstar}_{\vy_\star} + \noise[],
\end{align}
where $\noise[] \in \real^n \sim \mc N(0, \sigma \mat I_n)$ with
$\sigma = 0.1$.  In all cases, we keep the training sample size fixed
at $n=200$, and the maximum size of covariates is fixed at $2000$.
With these fixed choices, we consider several different choices of the
design matrix $\xmat$, along with two choices for the unknown
regression vector $\tstar$.

\paragraph{Choice for $\xmat$:}
We consider two possible random ensembles for the design matrix
$\xmat$.  Let $\mat A \in \real^{200 \times 2000}$ denote a matrix
whose entries are drawn iid from $\mc N(0, 1)$.  Let $\mat B \in
\real^{2000 \times 2000}$ denote a diagonal matrix such that $\mat
B_{ii} = \vert\cos i\vert$ for $i$ even, and $0$ otherwise. Then the
two choices for the design matrix $\xmat$ are (I) \emph{Gaussian
design} with $\xmat = \mat A$, and (II) \emph{Cosine design} with
$\xmat = \mat A \mat B$.

\paragraph{Choice for $\tstar$:}
In parallel, we consider two possible random ensembles for the unknown
regression vector $\tstar$.  In all cases, the response $\vy_\star$
depends on only on $d_\star$ covariates in total.  In setting (A)
where $\mathbf {d_\star < n}$, we choose $\tstar$ to have non-zero
entries for the index $\braces{11, 12, \ldots, 60}$, i.e., the true
dimensionality of the dataset is $d_\star =60$, which is less than the
sample size $n=200$.  In setting (B) where $\mathbf {d_\star > n}$, we
choose $\tstar$ to have non-zero entries for all indices in the set
$\braces{1, 2, \ldots, 400}$.  Consequently, the number of free
parameters $d_\star=400$ is much larger than the sample size
$n=200$. In both settings, the non-zero entries of $\tstar$ are drawn
iid from $\mc N(0, 1)$, and then normalized such that $\enorm{\tstar}=
1$. \\

\vspace*{0.2in}

Taking all possible combinations of random ensembles for $\xmat$ and
$\tstar$ yields four distinct experimental settings: IA, IB, IIA, IIB.
Given a particular setting---for instance, setting IA with Gaussian
design and $d_\star = 60 < n$ (IA)---we generate a dataset of $n$
observations and then fit different estimators with a varying number
of covariates (denoted by $d$); i.e., the response variable $\vy$
remains fixed and we only vary the dimensionality of the design matrix
for fitting OLS or Ridge model.  We then compute the test MSE
(computed on an independent draw of $\vy_\star^{\textrm{test}}$ of
size $n_{\textrm{test}}=1000$).  We redraw the noise in the
observation $50$ times ($\vy_\star^{\textrm{train}} \text{ and }
\vy_\star^{\textrm{test}}$ remain fixed), and plot the average of test
MSE over these runs. For a given $d$, the bias and variance for a
given estimator (OLS or Ridge) are computed as follows:
\begin{align*}
    \textrm{Bias}(d) &= \frac{1}{n_{\textrm{test}}}
    \enorm{\vy_\star^{\textrm{test}}-\overline{\vy^{\textrm{est}}}}^2,
    \qtext{and}\\ \textrm{Variance}(d) &=\frac{1}{50} \sum_{r=1}^{50}
    \frac{1}{n_{\textrm{test}}}
    \enorm{\vy^{\textrm{est}}_r-\overline{\vy^{\textrm{est}}}}^2,
\end{align*}
where $r$ denotes the index of the experiment (redraw of noise
$\noise[]$), $\vy^{\textrm{est}}_r$ denotes the estimate for the
response for the test dataset for the $r$-th experiment, and
$\overline{\vy^ {\textrm{est}}}$ denotes the average of the
predictions made by the estimator across all $50$ experiments.

\begin{figure}[ht]
    \centering
    \begin{tabular}{cc}
         \includegraphics[width=0.36\linewidth]{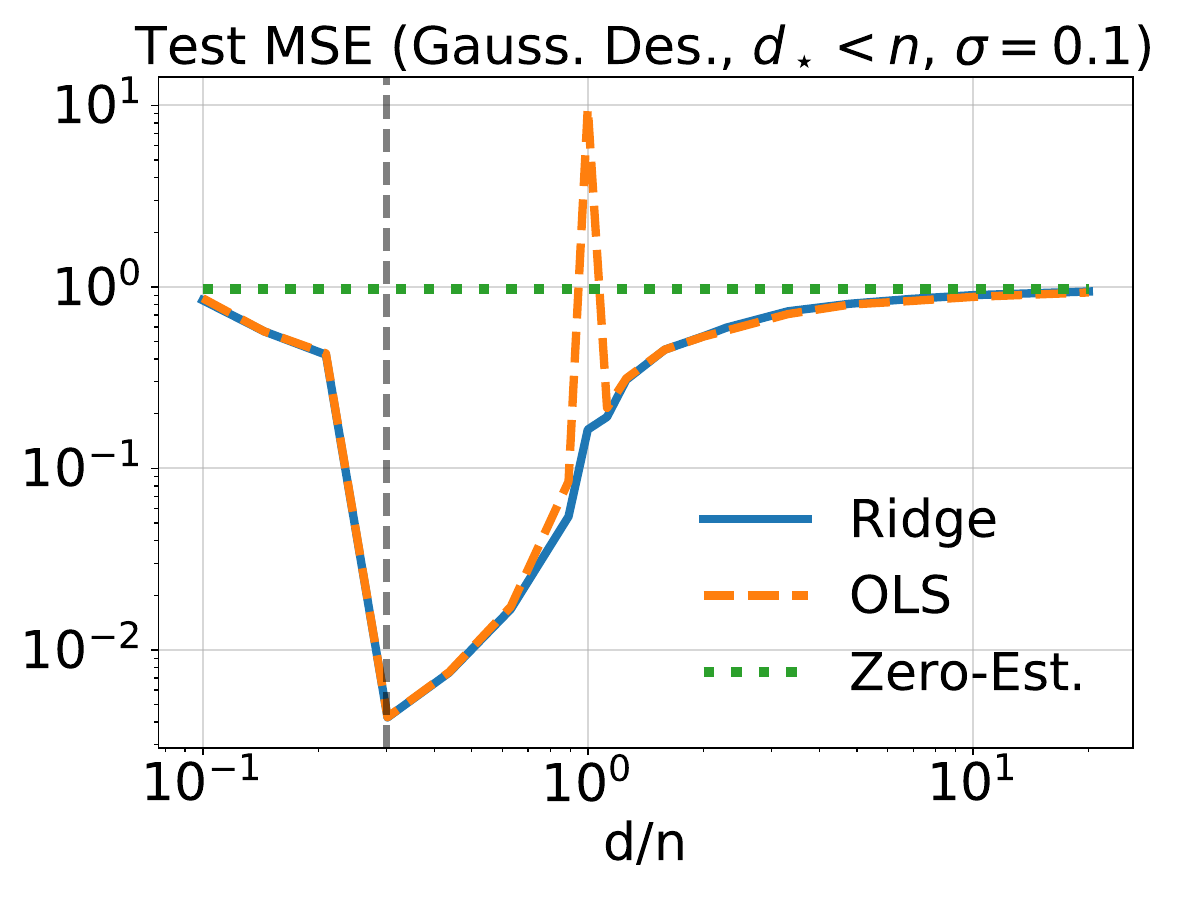}
         &
         \includegraphics[width=0.36\linewidth]{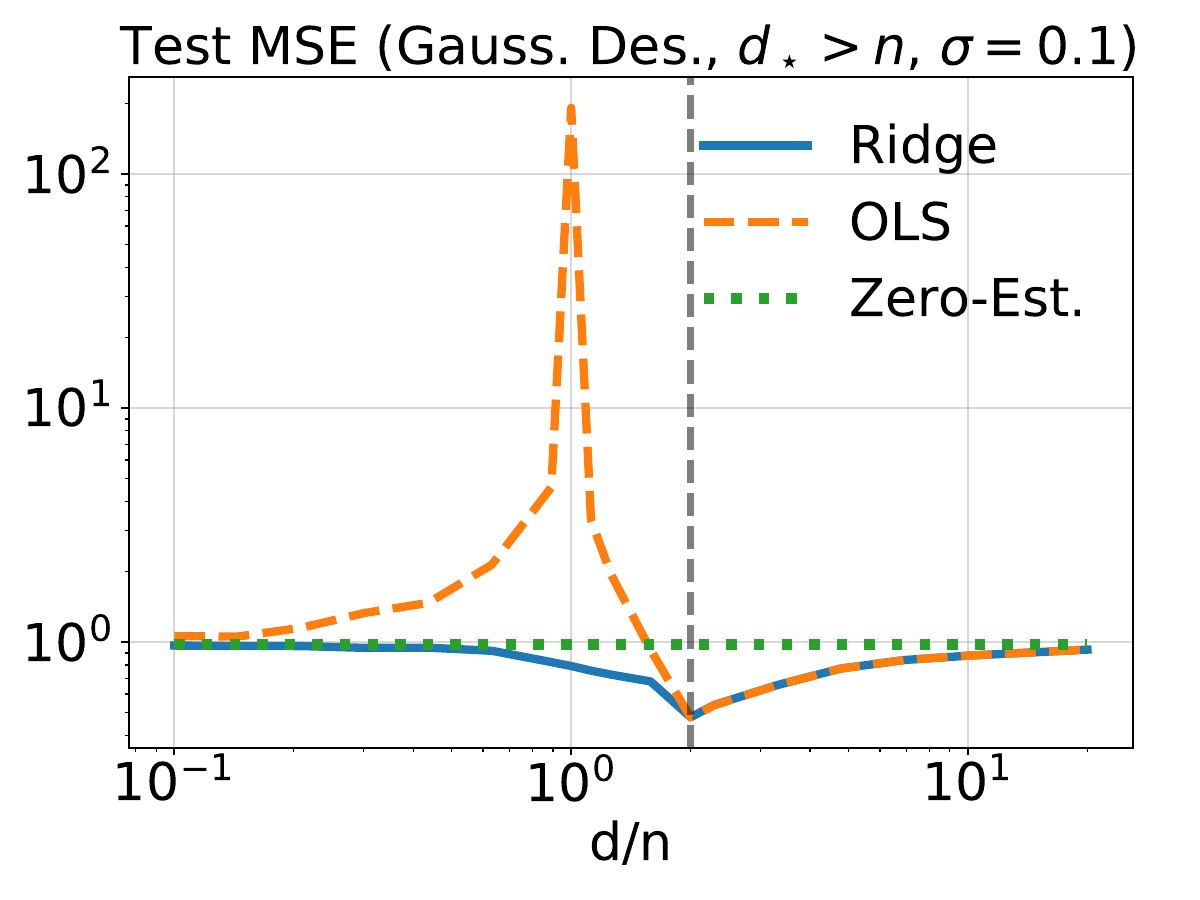}
         \\ (a) & (b)
         \\ \includegraphics[width=0.36\linewidth]{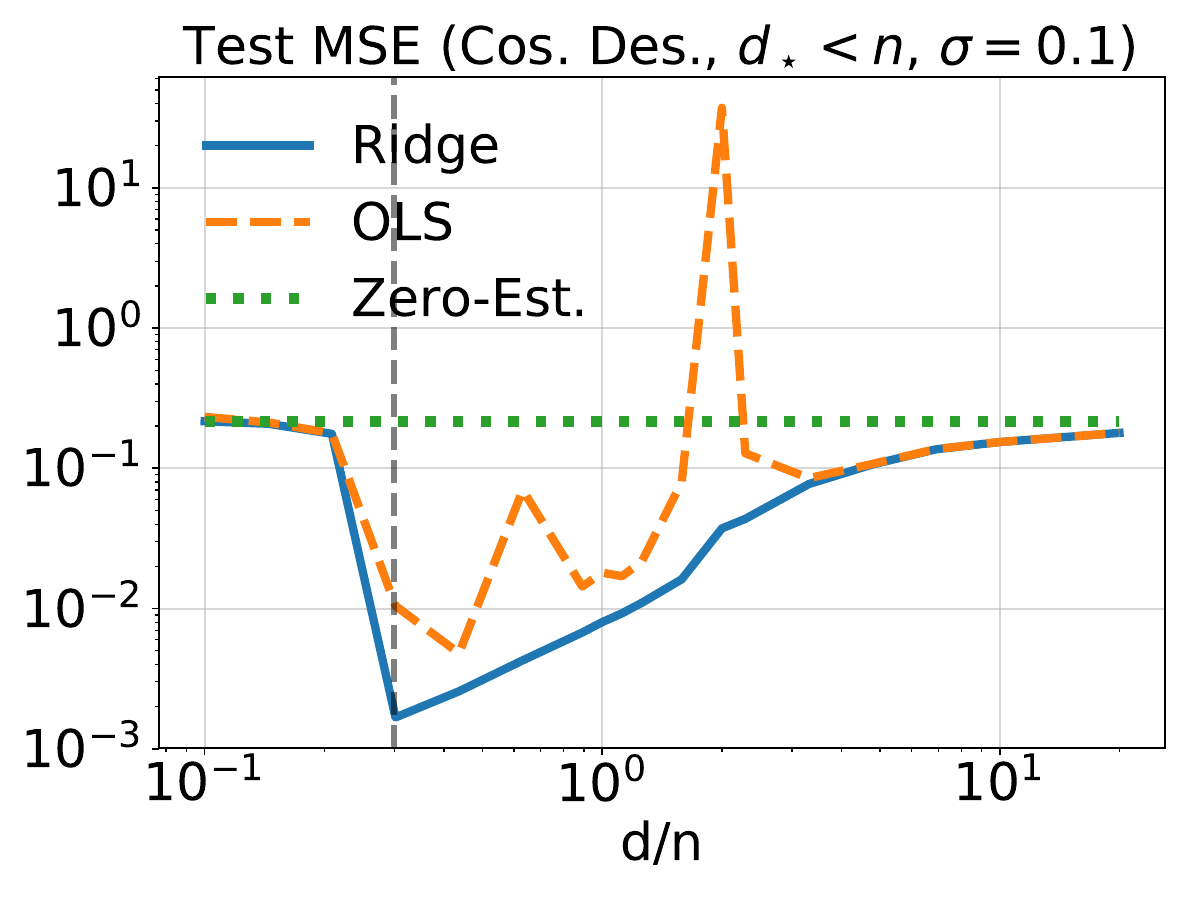}
         &
         \includegraphics[width=0.36\linewidth]{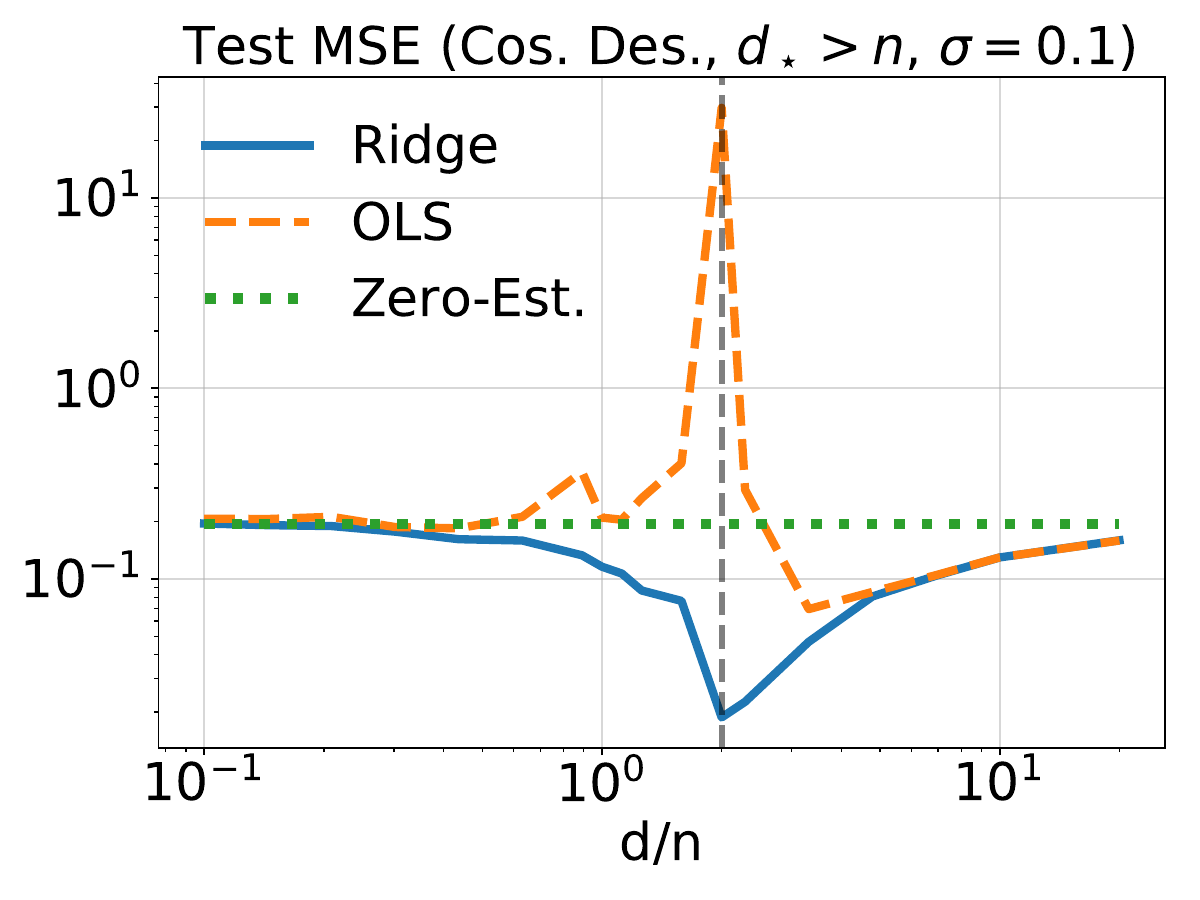}
         \\ (c) & (d)
    \end{tabular}
    \caption{Plots of test MSE for OLS and CV-tuned Ridge for the four
      designs discussed in \cref{sub:different_design}, versus the
      dimensionality of the covariates in the fitted model varies. The
      training sample size is fixed at $n=200$.
        } \vspace{2mm}
  \label{fig:test_mse_diff_design}
\end{figure}

\begin{figure}[ht]
    \centering
    \begin{tabular}{cc}
         \includegraphics[width=0.36\linewidth]{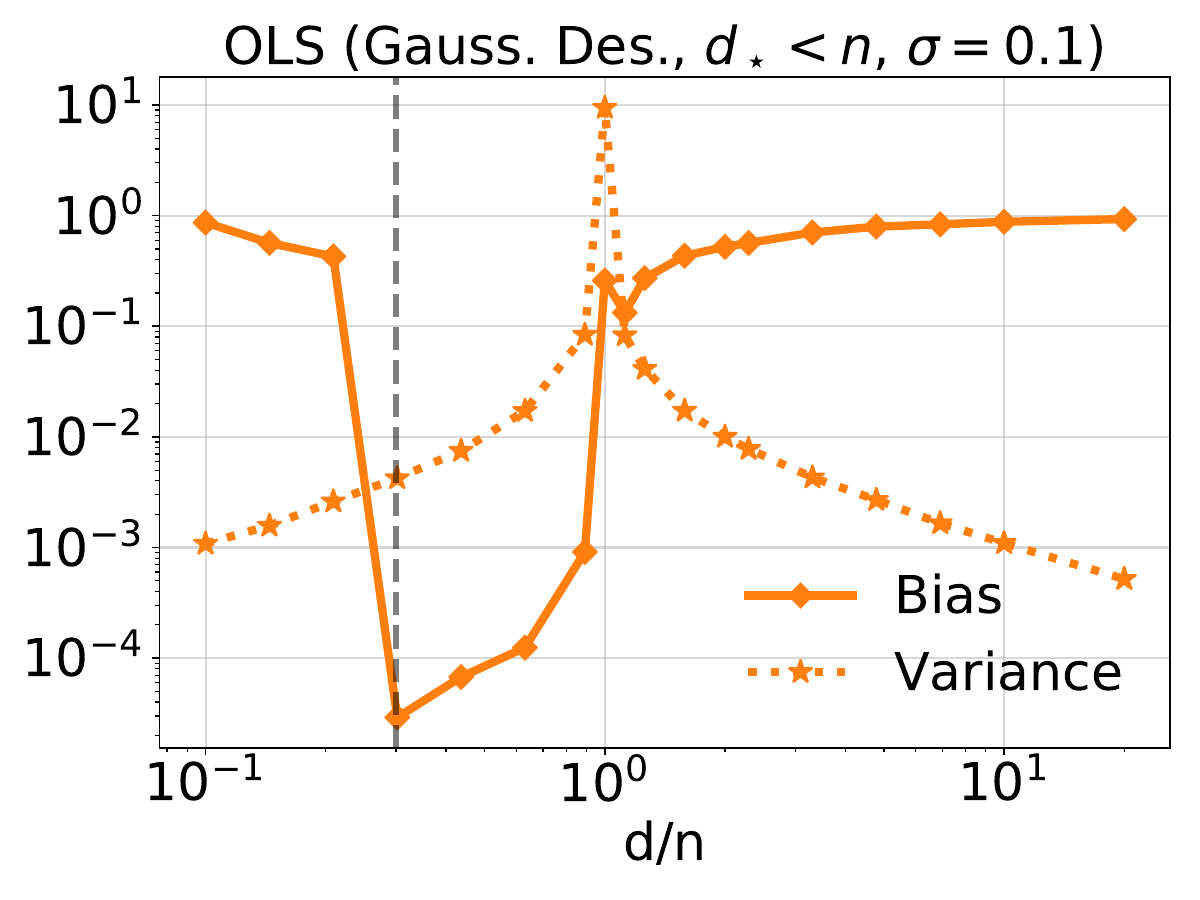}
         &
         \includegraphics[width=0.36\linewidth]{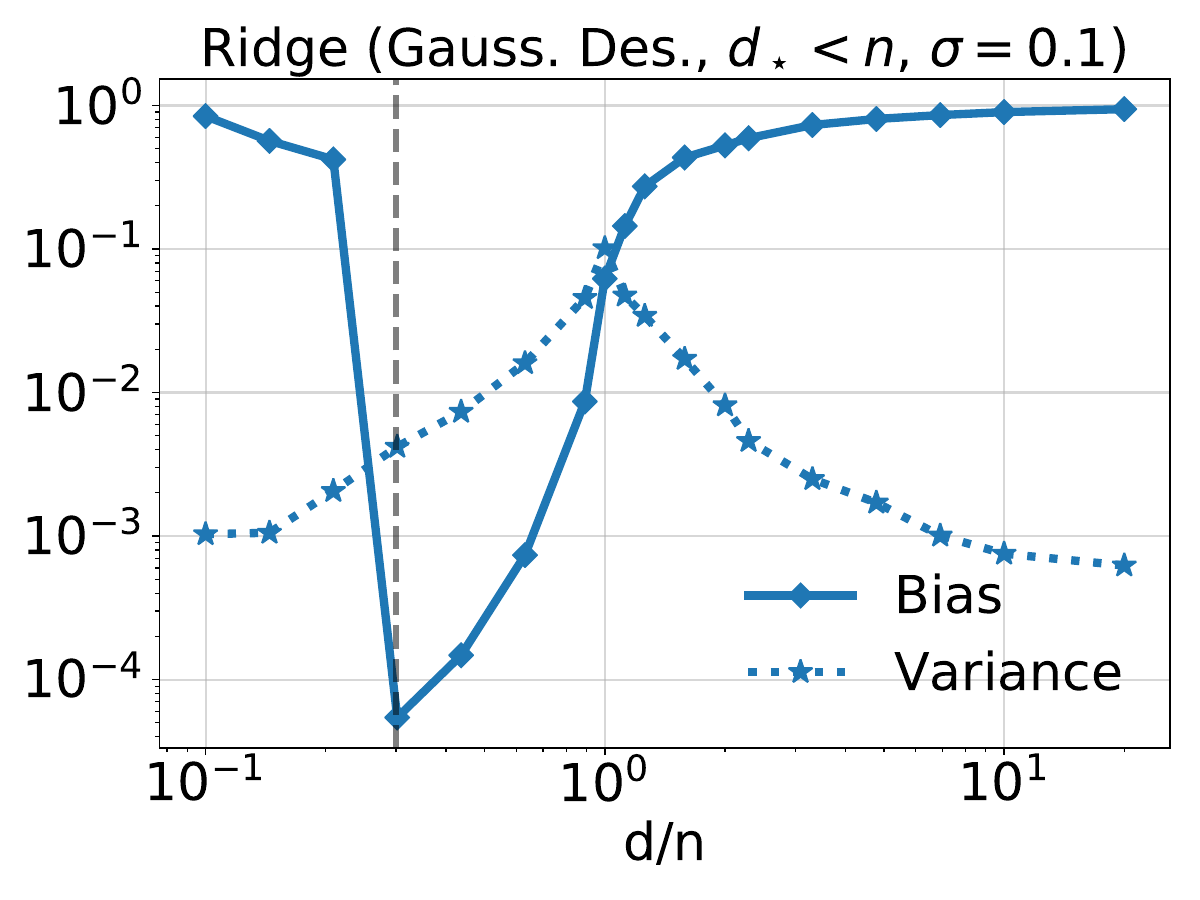}
         \\ (a) & (b)
         \\ \includegraphics[width=0.36\linewidth]{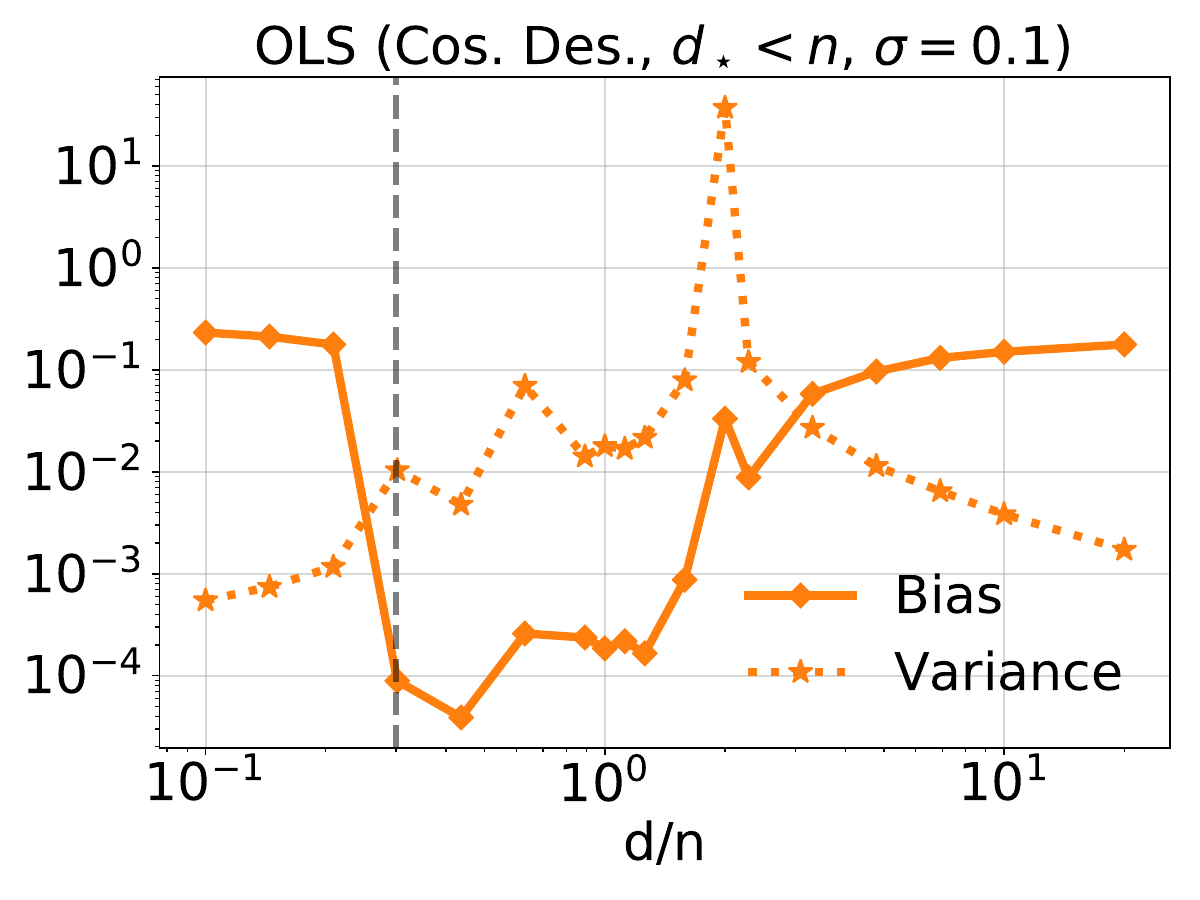}
         &
         \includegraphics[width=0.36\linewidth]{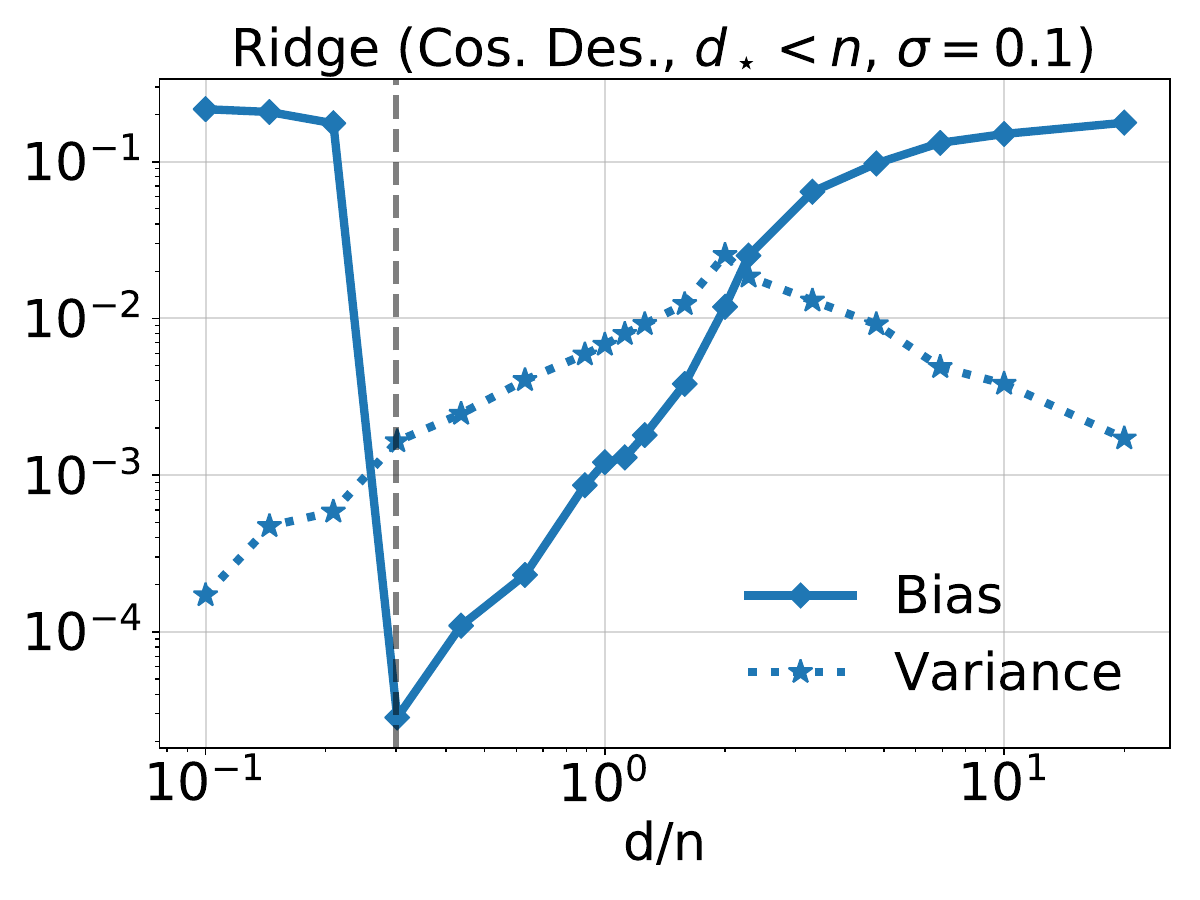}
         \\ (c) & (d)
    \end{tabular}
    \caption{Bias-variance tradeoff of OLS and CV-tuned Ridge
      estimators for various designs discussed in
      \cref{sub:different_design}. Panels (a) and (b) show the results
      for design IA corresponding to the
      \cref{fig:test_mse_diff_design}(a), and panels (c) and (d) show
      it for design IIA corresponding to the
      \cref{fig:test_mse_diff_design}(c).  }
  \label{fig:bias_var_diff_design}
\end{figure}

The four panels in \cref{fig:test_mse_diff_design} show, for each of
these four settings, plots of the radio $d/n$ versus test MSE for the
OLS and CV-tuned ridge estimators.  \cref{fig:bias_var_diff_design}
shows the underlying bias-variance tradeoff curves in settings IA and
IIA.

From these figures, we can draw the following conclusions.  On one
hand, the test MSE of the CV-tuned ridge estimator exhibits the
classical U-shaped curve for all the settings.  On the other hand, the
behavior of OLS is heavily dependent on the covariate design along
with the choice of $d_\star$. For $d<d_\star$, both OLS and Ridge show
the classical tradeoff between bias (deccreasing in $d$) and variance
(increasing in $d$). For $d>d_\star$, the bias of the tuned ridge
increases monotonically with $d$, but the variance increases until a
design-dependent threshold, after which it then decreases. In almost
all cases, the bias of OLS monotonically increases as $d$ increases
above $d_\star$; however, the variance term can show multiple peaks
depending on the design, and these peaks frequently show up in the
test MSE as well.

{\small \bibliography{refs.bib} }

\begin{thebibliography}{98}
\providecommand{\natexlab}[1]{#1}
\providecommand{\url}[1]{\texttt{#1}}
\expandafter\ifx\csname urlstyle\endcsname\relax
  \providecommand{\doi}[1]{doi: #1}\else
  \providecommand{\doi}{doi: \begingroup \urlstyle{rm}\Url}\fi

\bibitem[Advani and Saxe(2017)]{advani2017high}
Madhu~S Advani and Andrew~M Saxe.
\newblock High-dimensional dynamics of generalization error in neural networks.
\newblock \emph{arXiv preprint arXiv:1710.03667}, 2017.

\bibitem[Akaike(1974)]{akaike1974new}
Hirotugu Akaike.
\newblock A new look at the statistical model identification.
\newblock In \emph{Selected Papers of Hirotugu Akaike}, pages 215--222. Springer, 1974.

\bibitem[Allen-Zhu et~al.(2018)Allen-Zhu, Li, and Song]{allen2018convergence}
Zeyuan Allen-Zhu, Yuanzhi Li, and Zhao Song.
\newblock A convergence theory for deep learning via over-parameterization.
\newblock \emph{arXiv preprint arXiv:1811.03962}, 2018.

\bibitem[Anthony and Bartlett(2009)]{anthony2009neural}
Martin Anthony and Peter~L Bartlett.
\newblock \emph{Neural network learning: Theoretical foundations}.
\newblock Cambridge University Press, 2009.

\bibitem[Arora et~al.(2018)Arora, Ge, Neyshabur, and Zhang]{arora2018stronger}
Sanjeev Arora, Rong Ge, Behnam Neyshabur, and Yi~Zhang.
\newblock Stronger generalization bounds for deep nets via a compression approach.
\newblock \emph{arXiv preprint arXiv:1802.05296}, 2018.

\bibitem[Arora et~al.(2019)Arora, Cohen, Hu, and Luo]{arora2019implicit}
Sanjeev Arora, Nadav Cohen, Wei Hu, and Yuping Luo.
\newblock Implicit regularization in deep matrix factorization.
\newblock In \emph{Advances in Neural Information Processing Systems}, pages 7411--7422, 2019.

\bibitem[Bai and Yin(2008)]{bai2008limit}
Zhi-Dong Bai and Yong-Qua Yin.
\newblock Limit of the smallest eigenvalue of a large dimensional sample covariance matrix.
\newblock In \emph{Advances In Statistics}, pages 108--127. World Scientific, 2008.

\bibitem[Barron et~al.(1998)Barron, Rissanen, and Yu]{barron1998minimum}
Andrew Barron, Jorma Rissanen, and Bin Yu.
\newblock The minimum description length principle in coding and modeling.
\newblock \emph{IEEE Transactions on Information Theory}, 44\penalty0 (6):\penalty0 2743--2760, 1998.

\bibitem[Bartlett and Mendelson(2002)]{bartlett2002rademacher}
Peter~L Bartlett and Shahar Mendelson.
\newblock {Rademacher and Gaussian complexities: Risk bounds and structural results}.
\newblock \emph{The Journal of Machine Learning Research}, 3\penalty0 (Nov):\penalty0 463--482, 2002.

\bibitem[Bartlett et~al.(2017)Bartlett, Foster, and Telgarsky]{bartlett2017spectrally}
Peter~L Bartlett, Dylan~J Foster, and Matus~J Telgarsky.
\newblock Spectrally-normalized margin bounds for neural networks.
\newblock In \emph{Advances in Neural Information Processing Systems}, pages 6240--6249, 2017.

\bibitem[Bartlett et~al.(2020)Bartlett, Long, Lugosi, and Tsigler]{bartlett2020benign}
Peter~L Bartlett, Philip~M Long, G{\'a}bor Lugosi, and Alexander Tsigler.
\newblock Benign overfitting in linear regression.
\newblock \emph{Proceedings of the National Academy of Sciences}, 117\penalty0 (48):\penalty0 30063--30070, 2020.

\bibitem[Belkin et~al.(2018)Belkin, Ma, and Mandal]{belkin2018understand}
Mikhail Belkin, Siyuan Ma, and Soumik Mandal.
\newblock To understand deep learning we need to understand kernel learning.
\newblock \emph{arXiv preprint arXiv:1802.01396}, 2018.

\bibitem[Belkin et~al.(2019{\natexlab{a}})Belkin, Hsu, Ma, and Mandal]{belkin2019reconciling}
Mikhail Belkin, Daniel Hsu, Siyuan Ma, and Soumik Mandal.
\newblock Reconciling modern machine-learning practice and the classical bias--variance trade-off.
\newblock \emph{Proceedings of the National Academy of Sciences}, 116\penalty0 (32):\penalty0 15849--15854, 2019{\natexlab{a}}.

\bibitem[Belkin et~al.(2019{\natexlab{b}})Belkin, Hsu, and Xu]{belkin2019two}
Mikhail Belkin, Daniel Hsu, and Ji~Xu.
\newblock Two models of double descent for weak features.
\newblock \emph{arXiv preprint arXiv:1903.07571}, 2019{\natexlab{b}}.

\bibitem[Bernau et~al.(2014)Bernau, Riester, Boulesteix, Parmigiani, Huttenhower, Waldron, and Trippa]{bernau2014cross}
Christoph Bernau, Markus Riester, Anne-Laure Boulesteix, Giovanni Parmigiani, Curtis Huttenhower, Levi Waldron, and Lorenzo Trippa.
\newblock Cross-study validation for the assessment of prediction algorithms.
\newblock \emph{Bioinformatics}, 30\penalty0 (12):\penalty0 i105--i112, 2014.

\bibitem[Bietti and Bach(2021)]{bietti2021deep}
Alberto Bietti and Francis Bach.
\newblock Deep equals shallow for {ReLU} networks in kernel regimes.
\newblock In \emph{International Conference on Learning Representations}, 2021.
\newblock URL \url{https://openreview.net/forum?id=aDjoksTpXOP}.

\bibitem[Bietti and Mairal(2019)]{bietti2019inductive}
Alberto Bietti and Julien Mairal.
\newblock On the inductive bias of neural tangent kernels.
\newblock \emph{arXiv preprint arXiv:1905.12173}, 2019.

\bibitem[Blier and Ollivier(2018)]{blier2018description}
L{\'e}onard Blier and Yann Ollivier.
\newblock The description length of deep learning models.
\newblock In \emph{Advances in Neural Information Processing Systems}, pages 2216--2226, 2018.

\bibitem[B{\"u}hlmann and Yu(2003)]{buhlmann2003boosting}
Peter B{\"u}hlmann and Bin Yu.
\newblock Boosting with the {L2} loss: Regression and classification.
\newblock \emph{Journal of the American Statistical Association}, 98\penalty0 (462):\penalty0 324--339, 2003.

\bibitem[Buja et~al.(1989)Buja, Hastie, and Tibshirani]{buja1989linear}
Andreas Buja, Trevor Hastie, and Robert Tibshirani.
\newblock Linear smoothers and additive models.
\newblock \emph{The Annals of Statistics}, 17\penalty0 (2):\penalty0 453--510, 1989.

\bibitem[Dicker(2016)]{dicker2016ridge}
Lee~H Dicker.
\newblock Ridge regression and asymptotic minimax estimation over spheres of growing dimension.
\newblock \emph{Bernoulli}, 22\penalty0 (1):\penalty0 1--37, 2016.

\bibitem[Dobriban and Wager(2018)]{dobriban2018high}
Edgar Dobriban and Stefan Wager.
\newblock High-dimensional asymptotics of prediction: Ridge regression and classification.
\newblock \emph{The Annals of Statistics}, 46\penalty0 (1):\penalty0 247--279, 2018.

\bibitem[Du et~al.(2018)Du, Zhai, Poczos, and Singh]{du2018gradient}
Simon~S Du, Xiyu Zhai, Barnabas Poczos, and Aarti Singh.
\newblock Gradient descent provably optimizes over-parameterized neural networks.
\newblock \emph{arXiv preprint arXiv:1810.02054}, 2018.

\bibitem[Efron(1986)]{efron1986biased}
Bradley Efron.
\newblock How biased is the apparent error rate of a prediction rule?
\newblock \emph{Journal of the American Statistical Association}, 81\penalty0 (394):\penalty0 461--470, 1986.

\bibitem[Efron(2004)]{efron2004estimation}
Bradley Efron.
\newblock The estimation of prediction error: covariance penalties and cross-validation.
\newblock \emph{Journal of the American Statistical Association}, 99\penalty0 (467):\penalty0 619--632, 2004.

\bibitem[El~Karoui(2010{\natexlab{a}})]{el2010information}
Noureddine El~Karoui.
\newblock On information plus noise kernel random matrices.
\newblock \emph{The Annals of Statistics}, 38\penalty0 (5):\penalty0 3191--3216, 2010{\natexlab{a}}.

\bibitem[El~Karoui(2010{\natexlab{b}})]{el2010spectrum}
Noureddine El~Karoui.
\newblock The spectrum of kernel random matrices.
\newblock \emph{The Annals of Statistics}, 38\penalty0 (1):\penalty0 1--50, 2010{\natexlab{b}}.

\bibitem[Foster and Stine(2004)]{foster2004contribution}
Dean~P Foster and Robert~A Stine.
\newblock The contribution of parameters to stochastic complexity.
\newblock \emph{Advances in Minimum Description Length Theory and Applications}, pages 195--213, 2004.

\bibitem[Golowich et~al.(2017)Golowich, Rakhlin, and Shamir]{golowich2017size}
Noah Golowich, Alexander Rakhlin, and Ohad Shamir.
\newblock Size-independent sample complexity of neural networks.
\newblock \emph{arXiv preprint arXiv:1712.06541}, 2017.

\bibitem[Gr{\"u}nwald and Roos(2019)]{grunwald2019minimum}
Peter Gr{\"u}nwald and Teemu Roos.
\newblock Minimum description length revisited.
\newblock \emph{International journal of mathematics for industry}, 11\penalty0 (01):\penalty0 1930001, 2019.

\bibitem[Gr{\"u}nwald(2007)]{grunwald2007minimum}
Peter~D Gr{\"u}nwald.
\newblock \emph{The minimum description length principle}.
\newblock MIT press, 2007.

\bibitem[Gr{\"u}nwald and Mehta(2017)]{grunwald2017tightarxiv}
Peter~D Gr{\"u}nwald and Nishant~A Mehta.
\newblock {A tight excess risk bound via a unified PAC-Bayesian-Rademacher-Shtarkov-MDL complexity}.
\newblock \emph{arXiv preprint arXiv:1710.07732}, 2017.

\bibitem[Gr{\"u}nwald and Mehta(2019)]{grunwald2017tight}
Peter~D Gr{\"u}nwald and Nishant~A Mehta.
\newblock A tight excess risk bound via a unified {PAC-Bayesian--Rademacher--Shtarkov--MDL complexity}.
\newblock In \emph{Algorithmic Learning Theory}, pages 433--465. PMLR, 2019.

\bibitem[Hansen and Yu(2001)]{hansen2001model}
Mark~H Hansen and Bin Yu.
\newblock Model selection and the principle of minimum description length.
\newblock \emph{Journal of the American Statistical Association}, 96\penalty0 (454):\penalty0 746--774, 2001.

\bibitem[Hastie et~al.(2005)Hastie, Tibshirani, Friedman, and Franklin]{hastie2005elements}
Trevor Hastie, Robert Tibshirani, Jerome Friedman, and James Franklin.
\newblock The elements of statistical learning: data mining, inference and prediction.
\newblock \emph{The Mathematical Intelligencer}, 27\penalty0 (2):\penalty0 83--85, 2005.

\bibitem[Hastie et~al.(2019)Hastie, Montanari, Rosset, and Tibshirani]{hastie2019surprises}
Trevor Hastie, Andrea Montanari, Saharon Rosset, and Ryan~J Tibshirani.
\newblock Surprises in high-dimensional ridgeless least squares interpolation.
\newblock \emph{arXiv preprint arXiv:1903.08560}, 2019.

\bibitem[Hastie(2017)]{hastie2017generalized}
Trevor~J Hastie.
\newblock Generalized additive models.
\newblock In \emph{Statistical models in S}, pages 249--307. Routledge, 2017.

\bibitem[Hayou et~al.(2019)Hayou, Doucet, and Rousseau]{hayou2019meanfield}
Soufiane Hayou, Arnaud Doucet, and Judith Rousseau.
\newblock Mean-field behaviour of neural tangent kernel for deep neural networks.
\newblock \emph{arXiv preprint arXiv:1905.13654}, 2019.

\bibitem[Hinton and Van~Camp(1993)]{hinton1993keeping}
Geoffrey Hinton and Drew Van~Camp.
\newblock Keeping neural networks simple by minimizing the description length of the weights.
\newblock In \emph{Proceedings of the 6th Annual ACM Conference on Computational Learning Theory}. Citeseer, 1993.

\bibitem[Hsu et~al.(2012)Hsu, Kakade, and Zhang]{hsu2012random}
Daniel Hsu, Sham~M Kakade, and Tong Zhang.
\newblock Random design analysis of ridge regression.
\newblock In \emph{Conference on learning theory}, pages 9--1. JMLR Workshop and Conference Proceedings, 2012.

\bibitem[Jacot et~al.(2018)Jacot, Gabriel, and Hongler]{jacot2018neural}
Arthur Jacot, Franck Gabriel, and Cl{\'e}ment Hongler.
\newblock Neural tangent kernel: Convergence and generalization in neural networks.
\newblock In \emph{Advances in Neural Information Processing Systems}, pages 8571--8580, 2018.

\bibitem[Janson et~al.(2015)Janson, Fithian, and Hastie]{janson2015effective}
Lucas Janson, William Fithian, and Trevor~J Hastie.
\newblock Effective degrees of freedom: a flawed metaphor.
\newblock \emph{Biometrika}, 102\penalty0 (2):\penalty0 479--485, 2015.

\bibitem[Kaufman and Rosset(2014)]{kaufman2014does}
S~Kaufman and S~Rosset.
\newblock When does more regularization imply fewer degrees of freedom? sufficient conditions and counterexamples.
\newblock \emph{Biometrika}, 101\penalty0 (4):\penalty0 771--784, 2014.

\bibitem[Kolmogorov(1963)]{kolmogorov1963tables}
Andrei~N Kolmogorov.
\newblock On tables of random numbers.
\newblock \emph{Sankhy{\=a}: The Indian Journal of Statistics, Series A}, pages 369--376, 1963.

\bibitem[Kolmogorov(1968)]{kolmogorov1968three}
Andrei~Nikolaevich Kolmogorov.
\newblock Three approaches to the quantitative definition of information.
\newblock \emph{International Journal of Computer Mathematics}, 2\penalty0 (1-4):\penalty0 157--168, 1968.

\bibitem[Lee(2001)]{lee2001intrductionto}
Thomas~CM Lee.
\newblock An intrductionto coding theory iafid the two-part minimumljescription length’.
\newblock \emph{International Statistical Review}, 69\penalty0 (2):\penalty0 169--133, 2001.

\bibitem[Li et~al.(2018{\natexlab{a}})Li, Farkhoor, Liu, and Yosinski]{li2018measuring}
Chunyuan Li, Heerad Farkhoor, Rosanne Liu, and Jason Yosinski.
\newblock Measuring the intrinsic dimension of objective landscapes.
\newblock \emph{arXiv preprint arXiv:1804.08838}, 2018{\natexlab{a}}.

\bibitem[Li and Vit{\'a}nyi(2008)]{li2008introduction}
Ming Li and Paul Vit{\'a}nyi.
\newblock \emph{An introduction to Kolmogorov complexity and its applications}, volume~3.
\newblock Springer, 2008.

\bibitem[Li et~al.(2018{\natexlab{b}})Li, Lu, Wang, Haupt, and Zhao]{li2018tighter}
Xingguo Li, Junwei Lu, Zhaoran Wang, Jarvis Haupt, and Tuo Zhao.
\newblock On tighter generalization bound for deep neural networks: {CNNs, resnets, and beyond}.
\newblock \emph{arXiv preprint arXiv:1806.05159}, 2018{\natexlab{b}}.

\bibitem[Liu et~al.(2020)Liu, Zheng, and Feng]{liu2020estimation}
X~Liu, S~Zheng, and X~Feng.
\newblock Estimation of error variance via ridge regression.
\newblock \emph{Biometrika}, 107\penalty0 (2):\penalty0 481--488, 2020.

\bibitem[Loog et~al.(2020)Loog, Viering, Mey, Krijthe, and Tax]{loog2020brief}
Marco Loog, Tom Viering, Alexander Mey, Jesse~H Krijthe, and David~MJ Tax.
\newblock A brief prehistory of double descent.
\newblock \emph{Proceedings of the National Academy of Sciences}, 117\penalty0 (20):\penalty0 10625--10626, 2020.

\bibitem[MacKay(1994)]{mackay1994bayesian}
David~JC MacKay.
\newblock Bayesian nonlinear modeling for the prediction competition.
\newblock \emph{ASHRAE Transactions}, 100\penalty0 (2):\penalty0 1053--1062, 1994.

\bibitem[Mallows(1973)]{mallows1973some}
Colin~L Mallows.
\newblock Some comments on ${C}_p$.
\newblock \emph{Technometrics}, 15\penalty0 (4):\penalty0 661--675, 1973.

\bibitem[Marcenko and Pastur(1967)]{marvcenko1967distribution}
Vladimir~A Marcenko and Leonid~Andreevich Pastur.
\newblock Distribution of eigenvalues for some sets of random matrices.
\newblock \emph{Mathematics of the USSR-Sbornik}, 1\penalty0 (4):\penalty0 457, 1967.

\bibitem[Mercer(1909)]{mercer1909functions}
J~Mercer.
\newblock Functions of positive and negative type and their connection with the theory of integral equations.
\newblock \emph{Philosophical Transactions of The Royal Society}, pages 4--415, 1909.

\bibitem[Meyer and Woodroofe(2000)]{meyer2000degrees}
Mary Meyer and Michael Woodroofe.
\newblock On the degrees of freedom in shape-restricted regression.
\newblock \emph{The Annals of Statistics}, 28\penalty0 (4):\penalty0 1083--1104, 2000.

\bibitem[Miyaguchi and Yamanishi(2018)]{miyaguchi2018high}
Kohei Miyaguchi and Kenji Yamanishi.
\newblock High-dimensional penalty selection via minimum description length principle.
\newblock \emph{Machine Learning}, 107\penalty0 (8-10):\penalty0 1283--1302, 2018.

\bibitem[Muthukumar et~al.(2020)Muthukumar, Vodrahalli, Subramanian, and Sahai]{muthukumar2020harmless}
Vidya Muthukumar, Kailas Vodrahalli, Vignesh Subramanian, and Anant Sahai.
\newblock Harmless interpolation of noisy data in regression.
\newblock \emph{IEEE Journal on Selected Areas in Information Theory}, 2020.

\bibitem[Nakkiran et~al.(2019)Nakkiran, Kaplun, Bansal, Yang, Barak, and Sutskever]{Nakkiran2019DeepDD}
Preetum Nakkiran, Gal Kaplun, Yamini Bansal, Tristan Yang, Boaz Barak, and Ilya Sutskever.
\newblock Deep double descent: Where bigger models and more data hurt.
\newblock \emph{arXiv preprint arXiv:1912.02292}, 2019.

\bibitem[Nakkiran et~al.(2020)Nakkiran, Venkat, Kakade, and Ma]{nakkiran2020optimal}
Preetum Nakkiran, Prayaag Venkat, Sham Kakade, and Tengyu Ma.
\newblock Optimal regularization can mitigate double descent.
\newblock \emph{arXiv preprint arXiv:2003.01897}, 2020.

\bibitem[Neyshabur et~al.(2014)Neyshabur, Tomioka, and Srebro]{neyshabur2014search}
Behnam Neyshabur, Ryota Tomioka, and Nathan Srebro.
\newblock In search of the real inductive bias: On the role of implicit regularization in deep learning.
\newblock \emph{arXiv preprint arXiv:1412.6614}, 2014.

\bibitem[Neyshabur et~al.(2015)Neyshabur, Tomioka, and Srebro]{neyshabur2015norm}
Behnam Neyshabur, Ryota Tomioka, and Nathan Srebro.
\newblock Norm-based capacity control in neural networks.
\newblock In \emph{Conference on Learning Theory}, pages 1376--1401, 2015.

\bibitem[Neyshabur et~al.(2017)Neyshabur, Bhojanapalli, and Srebro]{neyshabur2017pac}
Behnam Neyshabur, Srinadh Bhojanapalli, and Nathan Srebro.
\newblock {A PAC-Bayesian approach to spectrally-normalized margin bounds for neural networks}.
\newblock \emph{arXiv preprint arXiv:1707.09564}, 2017.

\bibitem[Neyshabur et~al.(2019)Neyshabur, Li, Bhojanapalli, LeCun, and Srebro]{neyshabur2018role}
Behnam Neyshabur, Zhiyuan Li, Srinadh Bhojanapalli, Yann LeCun, and Nathan Srebro.
\newblock The role of over-parametrization in generalization of neural networks.
\newblock \emph{International Conference on Learning Representations}, page To appear, 2019.

\bibitem[Nishimoto et~al.(2011)Nishimoto, Vu, Naselaris, Benjamini, Yu, and Gallant]{nishimoto2011reconstructing}
Shinji Nishimoto, An~T Vu, Thomas Naselaris, Yuval Benjamini, Bin Yu, and Jack~L Gallant.
\newblock Reconstructing visual experiences from brain activity evoked by natural movies.
\newblock \emph{Current Biology}, 21\penalty0 (19):\penalty0 1641--1646, 2011.

\bibitem[Novak et~al.(2020)Novak, Xiao, Hron, Lee, Alemi, Sohl-Dickstein, and Schoenholz]{neuraltangents2020}
Roman Novak, Lechao Xiao, Jiri Hron, Jaehoon Lee, Alexander~A. Alemi, Jascha Sohl-Dickstein, and Samuel~S. Schoenholz.
\newblock Neural tangents: Fast and easy infinite neural networks in python.
\newblock In \emph{International Conference on Learning Representations}, 2020.
\newblock URL \url{https://github.com/google/neural-tangents}.

\bibitem[Olson et~al.(2017)Olson, La~Cava, Orzechowski, Urbanowicz, and Moore]{olson2017pmlb}
Randal~S Olson, William La~Cava, Patryk Orzechowski, Ryan~J Urbanowicz, and Jason~H Moore.
\newblock {PMLB: A large benchmark suite for machine learning evaluation and comparison}.
\newblock \emph{BioData mining}, 10\penalty0 (1):\penalty0 36, 2017.

\bibitem[Pedregosa et~al.(2011)Pedregosa, Varoquaux, Gramfort, Michel, Thirion, Grisel, Blondel, Prettenhofer, Weiss, Dubourg, et~al.]{pedregosa2011scikit}
Fabian Pedregosa, Ga{\"e}l Varoquaux, Alexandre Gramfort, Vincent Michel, Bertrand Thirion, Olivier Grisel, Mathieu Blondel, Peter Prettenhofer, Ron Weiss, Vincent Dubourg, et~al.
\newblock Scikit-learn: Machine learning in python.
\newblock \emph{The Journal of Machine Learning Research}, 12:\penalty0 2825--2830, 2011.

\bibitem[Raskutti et~al.(2011)Raskutti, Wainwright, and Yu]{raskutti2011minimax}
Garvesh Raskutti, Martin~J Wainwright, and Bin Yu.
\newblock Minimax rates of estimation for high-dimensional linear regression over $\ell_q $-balls.
\newblock \emph{IEEE Transactions on Information Theory}, 57\penalty0 (10):\penalty0 6976--6994, 2011.

\bibitem[Raskutti et~al.(2014)Raskutti, Wainwright, and Yu]{raskutti2014early}
Garvesh Raskutti, Martin~J Wainwright, and Bin Yu.
\newblock Early stopping and non-parametric regression: An optimal data-dependent stopping rule.
\newblock \emph{The Journal of Machine Learning Research}, 15\penalty0 (1):\penalty0 335--366, 2014.

\bibitem[Rissanen(1983)]{rissanen1983universal}
Jorma Rissanen.
\newblock A universal prior for integers and estimation by minimum description length.
\newblock \emph{The Annals of statistics}, 11\penalty0 (2):\penalty0 416--431, 1983.

\bibitem[Rissanen(1986)]{rissanen1986stochastic}
Jorma Rissanen.
\newblock Stochastic complexity and modeling.
\newblock \emph{The Annals of Statistics}, pages 1080--1100, 1986.

\bibitem[Santin and Schaback(2016)]{santin2016approximation}
Gabriele Santin and Robert Schaback.
\newblock Approximation of eigenfunctions in kernel-based spaces.
\newblock \emph{Advances in Computational Mathematics}, 42\penalty0 (4):\penalty0 973--993, 2016.

\bibitem[Schmidhuber(1997)]{schmidhuber1997discovering}
J{\"u}rgen Schmidhuber.
\newblock Discovering neural nets with low {K}olmogorov complexity and high generalization capability.
\newblock \emph{Neural Networks}, 10\penalty0 (5):\penalty0 857--873, 1997.

\bibitem[Schwarz(1978)]{schwarz1978estimating}
Gideon Schwarz.
\newblock Estimating the dimension of a model.
\newblock \emph{The Annals of Statistics}, 6\penalty0 (2):\penalty0 461--464, 1978.

\bibitem[Shen and Ye(2002)]{shen2002adaptive}
Xiaotong Shen and Jianming Ye.
\newblock Adaptive model selection.
\newblock \emph{Journal of the American Statistical Association}, 97\penalty0 (457):\penalty0 210--221, 2002.

\bibitem[Shtar'kov(1987)]{shtar1987universal}
Yurii~Mikhailovich Shtar'kov.
\newblock Universal sequential coding of single messages.
\newblock \emph{Problemy Peredachi Informatsii}, 23\penalty0 (3):\penalty0 3--17, 1987.

\bibitem[Silverstein(1995)]{silverstein1995strong}
Jack~W Silverstein.
\newblock Strong convergence of the empirical distribution of eigenvalues of large dimensional random matrices.
\newblock \emph{Journal of Multivariate Analysis}, 55\penalty0 (2):\penalty0 331--339, 1995.

\bibitem[Simonoff(2013)]{simonoff2013analyzing}
Jeffrey~S Simonoff.
\newblock \emph{Analyzing categorical data}.
\newblock Springer Science \& Business Media, 2013.

\bibitem[Skurichina and Duin(1998)]{skurichina1998bagging}
Marina Skurichina and Robert~PW Duin.
\newblock Bagging for linear classifiers.
\newblock \emph{Pattern Recognition}, 31\penalty0 (7):\penalty0 909--930, 1998.

\bibitem[Skurichina and Duin(2001)]{skurichina2001bagging}
Marina Skurichina and Robert~PW Duin.
\newblock Bagging and the random subspace method for redundant feature spaces.
\newblock In \emph{International Workshop on Multiple Classifier Systems}, pages 1--10. Springer, 2001.

\bibitem[Skurichina and Duin(2002)]{skurichina2002bagging}
Marina Skurichina and Robert~PW Duin.
\newblock Bagging, boosting and the random subspace method for linear classifiers.
\newblock \emph{Pattern Analysis \& Applications}, 5\penalty0 (2):\penalty0 121--135, 2002.

\bibitem[Stone(1982)]{stone1982optimal}
Charles~J Stone.
\newblock Optimal global rates of convergence for nonparametric regression.
\newblock \emph{The Annals of Statistics}, pages 1040--1053, 1982.

\bibitem[Tanaka et~al.(2005)Tanaka, Iwamoto, and Uehara]{tanaka2005discovery}
Yoshiki Tanaka, Kazuhisa Iwamoto, and Kuniaki Uehara.
\newblock Discovery of time-series motif from multi-dimensional data based on {MDL} principle.
\newblock \emph{Machine Learning}, 58\penalty0 (2-3):\penalty0 269--300, 2005.

\bibitem[Tibshirani(2015)]{tibshirani2015degrees}
Ryan~J Tibshirani.
\newblock Degrees of freedom and model search.
\newblock \emph{Statistica Sinica}, pages 1265--1296, 2015.

\bibitem[Tibshirani and Taylor(2012)]{tibshirani2012degrees}
Ryan~J Tibshirani and Jonathan Taylor.
\newblock Degrees of freedom in lasso problems.
\newblock \emph{The Annals of Statistics}, 40\penalty0 (2):\penalty0 1198--1232, 2012.

\bibitem[Tsigler and Bartlett(2020)]{tsigler2020benign}
Alexander Tsigler and Peter~L Bartlett.
\newblock Benign overfitting in ridge regression.
\newblock \emph{arXiv preprint arXiv:2009.14286}, 2020.

\bibitem[Tulino and Verd{\'u}(2004)]{tulino2004random}
Antonia~M Tulino and Sergio Verd{\'u}.
\newblock Random matrix theory and wireless communications.
\newblock \emph{Foundations and Trends{\textregistered} in Communications and Information Theory}, 1\penalty0 (1):\penalty0 1--182, 2004.

\bibitem[Van De~Geer(2006)]{van2006empirical}
Sara Van De~Geer.
\newblock \emph{Empirical Processes in M-estimation}.
\newblock Cambridge University Press, 2006.

\bibitem[Vanschoren et~al.(2013)Vanschoren, van Rijn, Bischl, and Torgo]{OpenML2013}
Joaquin Vanschoren, Jan~N. van Rijn, Bernd Bischl, and Luis Torgo.
\newblock Openml: Networked science in machine learning.
\newblock \emph{SIGKDD Explorations}, 15\penalty0 (2):\penalty0 49--60, 2013.
\newblock \doi{10.1145/2641190.2641198}.
\newblock URL \url{http://doi.acm.org/10.1145/2641190.2641198}.

\bibitem[Vapnik and Chervonenkis(2015)]{vapnik2015uniform}
Vladimir~N Vapnik and A~Ya Chervonenkis.
\newblock On the uniform convergence of relative frequencies of events to their probabilities.
\newblock In \emph{Measures of Complexity}, pages 11--30. Springer, 2015.

\bibitem[{Virtanen} et~al.(2020){Virtanen}, {Gommers}, {Oliphant}, {Haberland}, {Reddy}, et~al.]{2020SciPy-NMeth}
Pauli {Virtanen}, Ralf {Gommers}, Travis~E. {Oliphant}, Matt {Haberland}, Tyler {Reddy}, and SciPy 1.~0 others.
\newblock {SciPy} 1.0: Fundamental algorithms for scientific computing in python.
\newblock \emph{Nature Methods}, 2020.
\newblock \doi{https://doi.org/10.1038/s41592-019-0686-2}.

\bibitem[Wainwright(2019)]{wainwright2019high}
Martin~J Wainwright.
\newblock \emph{High-dimensional statistics: A non-asymptotic viewpoint}, volume~48.
\newblock Cambridge University Press, 2019.

\bibitem[Yu and Kumbier(2020)]{yu2020veridical}
Bin Yu and Karl Kumbier.
\newblock Veridical data science.
\newblock \emph{Proceedings of the National Academy of Sciences}, 117\penalty0 (8):\penalty0 3920--3929, 2020.

\bibitem[Zhang et~al.(2012)Zhang, Shen, and Mumford]{zhang2012generalized}
Bo~Zhang, Xiaotong Shen, and Sunni~L Mumford.
\newblock Generalized degrees of freedom and adaptive model selection in linear mixed-effects models.
\newblock \emph{Computational Statistics \& Data Analysis}, 56\penalty0 (3):\penalty0 574--586, 2012.

\bibitem[Zhang(2006)]{zhang2006}
Tong Zhang.
\newblock From $\varepsilon$-entropy to {KL}-entropy: Analysis of minimum information complexity density estimation.
\newblock \emph{The Annals of Statistics}, 34\penalty0 (5):\penalty0 2180--2210, 2006.

\bibitem[Zhang et~al.(2015)Zhang, Duchi, and Wainwright]{zhang2015divide}
Yuchen Zhang, John Duchi, and Martin Wainwright.
\newblock Divide and conquer kernel ridge regression: A distributed algorithm with minimax optimal rates.
\newblock \emph{The Journal of Machine Learning Research}, 16\penalty0 (1):\penalty0 3299--3340, 2015.

\bibitem[Zou et~al.(2007)Zou, Hastie, and Tibshirani]{zou2007degrees}
Hui Zou, Trevor Hastie, and Robert Tibshirani.
\newblock On the “degrees of freedom” of the lasso.
\newblock \emph{The Annals of Statistics}, 35\penalty0 (5):\penalty0 2173--2192, 2007.

\end{thebibliography}

\end{document}